\documentclass[a4paper,11pt]{article}

\usepackage{soul,xcolor}
\usepackage{cancel}

\soulregister\cite7
\soulregister\ref7
\soulregister\pageref7

\ifdefined\forReviewers

\DeclareRobustCommand{\oureqref}[1]{}
\soulregister{\oureqref}{1}
\newcommand{\NOWE}[1]{\hl{#1}}

\newcommand{\KASUJ}[1]{\st{#1}}
\newcommand{\KASUJROVW}[1]{\cancel{#1}}
\else
\newcommand{\oureqref}{\eqref}
\newcommand{\NOWE}[1]{{#1}}

\newcommand{\KASUJ}[1]{}
\newcommand{\KASUJROVW}[1]{\nonumber}
\fi

\advance\hoffset by -0.5truein
\usepackage[utf8]{inputenc}
\usepackage{graphicx}
\usepackage{epstopdf}
\usepackage{diagbox}
\usepackage{tikz}
\usepackage{booktabs}

\usepackage{amsmath}
\usepackage{amssymb}
\usepackage{inputenc}
\usepackage{xcolor}
\usepackage{url}
\usepackage{pdflscape}

\usepackage{amssymb,latexsym}

\usepackage{enumerate}

 \usepackage[T1]{fontenc}

\usepackage{subcaption}

\newfont{\mm}{eufm10 scaled 1200}
\newcommand{\forReviewers}{1}
\newcommand{\Bem}[1]{}
\begin{document}


\vspace{0.7cm}


\begin{center}
    \Large\bf Sensitivity Analysis of the Transformation \\ Using a Parameterized CEC 2017 Benchmark
\end{center}

\vspace{0.55cm}

\centerline{\large Grzegorz Sroka } \vspace{4mm}
\centerline{\footnotesize\it Department of Analysis Nonlinear, Rzesz\'ow University of Technology, Powsta\'nc\'ow Warszawy 12,}
\centerline{\footnotesize\it  35-959 Rzesz\'ow, Poland}
\centerline{\footnotesize\it  e-mail: gsroka@prz.edu.pl}
\vspace{0.4cm}
\centerline{\large Sławomir T. Wierzchoń} \vspace{4mm}
\centerline{\footnotesize\it   Institute of Computer Science, Polish Academy of Sciences, }
\centerline{\footnotesize\it   ul. Jana Kazimierza 5, 01-248 Warsaw, Poland}

\vspace{0,35cm}

{\footnotesize  {\bf Abstract.}  \it  \ 
\Bem{The standard CEC 2017 benchmark applies bias, shift, and rotation simultaneously, thereby confounding their individual effects on algorithmic behavior.}
\NOWE{In the standard CEC 2017 benchmark test, bias, shift, and rotation are applied simultaneously, making it difficult to distinguish the individual effects of these operations on the algorithm's behavior.}
This study introduces a parameterized implementation that controls these transformations independently while preserving the original benchmark functions and transformation data. The framework is used to diagnose the hybrid Marine Predators Algorithm (hMPA), whose predicted-candidate mechanism relies on numerical objective values and coordinate-wise reconstruction. Deep Statistical Comparison (DSC) and extended DSC (eDSC) are adapted from algorithm-level comparison to configuration-level diagnosis. To our knowledge, this enables the first exhaustive DSC and eDSC analysis of all eight bias–shift–rotation configurations of a parameterized CEC~2017 benchmark. 
The study covers all 56 three-configuration subsets and comparisons with the untransformed control. Experiments involve 29 functions \Bem{dimensions} with dimensions $dim \in \{10, 30,  50, 100\}$, 30 independent runs, and a fixed 
budget of $10000\cdot dim$ objective-function evaluations.

\NOWE{DSC and eDSC reveal statistically significant differences among configurations at every tested dimension. However, configuration rankings vary across benchmark functions, precluding a consistent ordering.} 
Most individual effect-size confidence intervals included zero. Of those that excluded zero, MPA-favouring comparisons were more frequent at $dim = 10$, whereas hMPA-favouring comparisons were more frequent at $dim = 50$ and $dim = 100$. Shift generally worsens 
objective-value outcomes. Isolated rotation causes no systematic deterioration, whereas isolated bias produces little change in final-solution distributions. Configurations 
combining shift and rotation differ most consistently from the control. Convergence plots further show function- and dimension-dependent separation, plateaus, and late improvement. 

\Bem{The numerical findings are limited to hMPA, the adopted parameter settings, 30 runs, the fixed evaluation budget, and the analyzed CEC~2017 functions. Within this scope, the framework provides a reproducible diagnostic protocol 
that can be extended to other continuous optimizers and adapted to related CEC \Bem{protocols}\NOWE{benchmarks}, including CEC~2024 and CEC~2025.
}
\NOWE{Although the numerical findings are limited to hMPA, they can be easily extended to other continuous optimizers and adapted to related CEC \KASUJ{protocols}\NOWE{benchmarks}, including CEC~2024 and CEC~2025.}
}

\vspace{0,3cm}


\vskip0.15cm

\section{Introduction}

Benchmarking is most informative when it explains why algorithm performance changes, rather than only reporting final rankings. This requires controlled variation of relevant problem properties and analysis beyond a single terminal objective value. The issue is particularly important for transformed test functions, where objective offset, optimum relocation, and variable dependencies may affect different elements of an optimization method. When these transformations are activated jointly, their individual roles remain confounded. 

The CEC~2017 benchmark (Competitions on Evolutionary Computation 2017), defined by Awad et al. in \cite{NHA2016}, is a standard suite for single-objective real-parameter optimization. Its conventional form activates bias, shift, and rotation simultaneously. Building on the binary transformation logic introduced for CEC~2021 by Mohamed et al. in \cite{AWM2023}, this study develops a parameterization layer for CEC~2017. The original function definitions and transformation data were retained, \Bem{while the three components are controlled independently}\NOWE{with three of their components controlled independently}. The complete configuration reproduces the standard benchmark. The remaining configurations provide diagnostic variants for separating transformation-specific responses.

The diagnostic case study concerns the hybrid Marine Predators Algorithm (hMPA,  with $h$ denoting the hybrid variant, i.e., the $h$-operator). It combines the Marine Predators Algorithm (MPA), introduced by Faramarzi et al. in \cite{AF2020}, with the predicted-candidate operator proposed by Oszust, Sroka, and Cymerys in \cite{MO2021}. 
MPA was ranked second out of nine methods on CEC~2017 (for $dim = 10$; limit: 50000 function evaluations, see Faramarzi et al. in \cite{AF2020}), behind LSHADE-cnEpSin. Post hoc tests revealed significant advantages over the tested:  Particle Swarm Optimization (PSO),  Genetic Algorithm (GA) and Covariance Matrix Adaptation Evolution 
Strategy (CMA-ES) implementations. No significant differences were detected from Success-History Based Parameter Adaptation 
Differential Evolution (SHADE) or Ensemble Sinusoidal Differential Covariance Matrix Adaptation with Euclidean Neighborhood (LSHADE-cnEpSin). MPA’s 3-phase search,  described by Faramarzi et al. in \cite{AF2020}, uses coordinate-wise Brownian and L\'evy updates. These features, along with the reported competitiveness, motivate the choice of MPA as the base algorithm. The hMPA case study examines sensitivity to bias, shift and rotation.
\Bem{This choice is deliberate.}The 
$h$-operator depends on numerical objective values and coordinate-wise candidate reconstruction. Bias, shift, and rotation can therefore probe distinct aspects of its operation. The objective is not to produce another ranking of metaheuristics, but to diagnose the sensitivity of a specific hybrid mechanism to controlled changes in problem representation.

The study makes three \NOWE{significant} contributions. First, it provides a faithful parameterized implementation of CEC~2017 with independent control of bias, shift, and rotation. 
Second, it adapts Deep Statistical Comparison (DSC) and its extended variant, extended Deep Statistical Comparison (eDSC), developed by Eftimov and Korošec in \cite{TE2019}, from algorithm-level comparison to configuration-level diagnosis.
The workflow combines distributions of final objective values and solution vectors, global and control comparisons, all 56 three-configuration subsets, convergence plots, and numerical handling of rank-deficient covariance structures. To the best of our knowledge, this is the first exhaustive DSC and eDSC analysis of all eight bias–shift–rotation configurations of a parameterized CEC~2017 benchmark. Third, the framework is evaluated on 29 admissible functions, four problem dimensions (i.e., $dim\in\{10, \ 30, \ 50, \ 100\}$), and 30 independent runs under a common objective-function evaluation budget. This scope provides a systematic and reproducible basis for diagnosing transformation sensitivity across functions and dimensions.

The paper is organized into nine sections: Section 2 reviews related work. Section 3 presents the parameterized benchmark. Sections 4 – 6 describe the experimental protocol, hMPA, and statistical analysis. Sections 7 and 8 present and discuss the results. Section 9 concludes the study.

\section{Related Work}

Opara and Arabas in \cite{KO2011}, Bartz-Beielstein et al. in \cite{TBB2020}, and Hansen et al. in \cite{NH2021} indicate that a reliable assessment of metaheuristics requires an experimental procedure that defines the research objective, the problem class, the cost measure, and the comparison rules. The effectiveness of a heuristic, as emphasized by Michalewicz and Fogel in \cite{ZM2004}, depends on the match between the method and the problem structure. Results obtained on test functions should therefore be interpreted within the scope of a given benchmark, and \NOWE{they} should not be treated as a basis for general conclusions about all classes of optimization problems. This dependence is confirmed by Piotrowski et al. in \cite{APP2023}, \Bem{showing} \NOWE{who demonstrate} that the choice of a test benchmark may substantially affect final algorithm rankings. In turn, the notion of the \textbf{best algorithm} (i.e., the method that is best with respect to the adopted effectiveness and cost criteria), as indicated by Opara and Arabas in \cite{KO2011}, requires defining the problem class, the evaluation budget, the number of independent runs, the method of aggregating results, and the statistical analysis. Hellwig and Beyer in \cite{MH2019} and Bartz-Beielstein et al. in \cite{TBB2020} develop this logic, pointing to the importance of a transparent procedure, appropriate problem selection, statistical analysis, and reproducibility. Under such assessment conditions, the number of objective function evaluations is, according to the work of Hansen et al. in  \cite{NH2021}, the basic cost measure in black-box optimization.

A separate line of research concerns the properties of test functions. Naser et al. in  \cite{MZN2025} classify benchmarks, among others, according to modality, dimensionality, separability, smoothness, constraints, and noise. Important properties, according to \Bem{the work of Opara and Arabas in} \cite{KO2011}, also include conditioning, non-separability, the number of local minima, and the location of the optimum. These properties may clearly change the observed effectiveness of algorithms. \textbf {Centre-bias}, i.e., preferential search of the centre of the feasible space, is particularly important. In turn, Ma et al. in \cite{ZM2023} show that centre-bias may overestimate the assessment of a method when the optimum is located in the centre, and may reduce its effectiveness after the optimum is shifted. They also indicate that good \Bem{results}\NOWE{performance} of the new metaheuristics on classical functions without shift and rotation do not necessarily \Bem{persist on more difficult transformed benchmarks.} \NOWE{hold for more challenging, transformed test sets.} This conclusion is strengthened by Kudela et al. in \cite{JKc2022}, showing that the central location of the optimum may favour algorithms with center-bias and make their reliable comparison with methods without this mechanism more difficult.

Michalewicz et al. in \cite{ZM2000} emphasized that comparisons of optimization methods should be based on scalable test cases in which important problem properties can be controlled. \Bem{Appropriate parameterization of test functions, in accordance with the methodology of Kudela and Matousek in \cite{JK2022}, makes it possible to regulate problem difficulty and reveal differences between algorithms that are not visible in simpler benchmark settings.} \NOWE{Proper parameterization of test functions, in accordance with the methodology of Kudela and Matousek in \cite{JK2022}, allows one to adjust the difficulty level of a task and reveal differences between algorithms that are not apparent in simpler test configurations. The spring benchmark by Sroka and Wierzchoń in \cite{GS2026a} compares baseline and hybrid metaheuristics. It independently controls material variability, manufacturing tolerances and the margin of requirements at fixed dimension.}

The CEC~2017 benchmark \cite{NHA2016} is a standard reference point for single-objective numerical optimization with real decision variables. Awad et al. in \cite{NHA2016} define this benchmark as a suite of unimodal ($1 - 3$), multimodal ($4 - 10$), hybrid ($11 - 20$), and composition ($21 - 30$) functions, evaluated, among others, for $dim\in\{10,\ 30, \ 50, \ 100\}$. From a diagnostic perspective, a limitation of the standard CEC~2017 benchmark is the lack of direct separation of the contribution of individual transformations. Parameterized benchmarking, in the approach of Opara et al. in \cite{KO2020}, addresses this limitation by controlling selected properties of test functions \Bem{in order} to improve the interpretation of algorithm behaviour. Mohamed et al. in \cite{AWM2023} apply a similar logic in the CEC~2021 benchmark, where bias, shift, and rotation are binary operators, and their combinations create different variants of test problems. Controlled landscape generators, analyzed by Yazdani et al. in \cite{DY2025}, make it possible to isolate problem properties and obtain an assessment with greater diagnostic value than in the case of a fixed suite of test functions.

The evaluation budget is an important component of the experimental protocol. The number of objective function calls, treated by Opara and Arabas in \cite{KO2011} as the basic cost measure in black-box optimization, is fixed in this study to ensure comparability across all benchmark configurations. Therefore, the final result should be interpreted within the adopted fixed computational budget.

Convergence analysis complements this fixed-budget perspective. 
The final objective value describes the outcome obtained at the end of the computational budget, but not the trajectory leading to it. Intermediate results, as emphasized by Derrac et al. in \cite{JD2014}, allow the rate of improvement, stagnation phases, and the moment at which differences between methods emerge to be assessed. Convergence plots are therefore an important complement to final quality measures, especially when the compared variants achieve similar final results (see Section \ref{Sec7.4}).

An important element of benchmarking is the choice of statistical analysis. Standard comparisons of metaheuristics rely on means, medians, standard deviations, and rank-based tests (see Sections: \ref{sec7.1}-\ref{sec7.3} and \ref{sec7.5}). \Bem{Although necessary, these tools may be insufficient when similar final objective values conceal differences in the underlying distributions of outcomes.}
\NOWE{Although these tools are essential, they may prove insufficient when similar target values mask differences in the underlying distributions of results.}

According to Eftimov and Koro{\v{s}}ec in \cite{TE2019}, DSC evaluates performance using the full distributions of objective values rather than individual descriptive statistics. eDSC complements this analysis by examining the distributions of the corresponding solutions in the decision space. \Bem{As shown by Koro{\v{s}}ec and Eftimov in \cite{PK2020},}\NOWE{Moreover, these authors noticed that} the joint use of DSC and eDSC can also support the diagnosis of exploration and exploitation throughout the optimization process, rather than only at its endpoint. Such analysis facilitates the interpretation of stability, result dispersion, and the relationship between exploration and exploitation, which Morales-Casta{\~n}eda et al. in \cite{BMC2020} associate with population diversification dynamics, and not only with the final quality of solutions.

Numerous variants of the Marine Predators Algorithm are also being developed in parallel. Faramarzi et al. in \cite{AF2020} proposed MPA as a population-based metaheuristic inspired by the foraging strategy of marine predators.\Bem{The algorithm uses search phases associated with Lévy and Brownian motions.} The review by Philibus et al. in \cite{EP2025} indicates that subsequent modifications of MPA usually focus on improving convergence, population diversification, and the balance between exploration and exploitation.

This group includes hMPA, in which Oszust, Sroka and Cymerys in \cite{MO2021} attach a mechanism of predicted solution candidates to the base algorithm. This variant is aimed at modifying population dynamics and the exploration-exploitation balance. The need to study such constructions under transformed objective functions is strengthened by the work of Sroka and Wierzchoń in \cite{GS2026}, indicating that transformations may differentiate the robustness and effectiveness of hybrid metaheuristics.

\section{Parameterized variant of CEC~2017}

In the experimental part, we use a parameterized variant of the CEC~2017 benchmark. 
We preserve the evaluation routines of the retained CEC~2017 functions, shift data, rotation matrices, bias values, and the common range of decision variables (according to the report by Awad et al. in \cite{NHA2016}), i.e., $x \in [-100, 100]^{dim}$. We modify only the activation status of 3-transformation components: bias, shift, and rotation. This parameterisation adapts the binary transformation logic introduced in CEC~2021 benchmark (Mohamed et al. in \cite{AWM2020}).
It provides an experimental tool for evaluating the sensitivity of hMPA (MPA) to transformation settings and their interactions across the retained CEC 2017 functions.
The \ \textbf{shift, rotation, bias} \ scheme for function $F_i$, where $i$ denotes the test function index, is showed schematically as follows:
\begin{equation}
\label{eq:shift_rotation_bias}
F_i(x) = f_{i} \left(M_i(x - o_i)\right) + F_i^*
\end{equation}
where $f_i$ is the internal structure of the function, $o_i$ is the shift vector, $M_i$ is the rotation matrix, and $F_i^*$ is the external bias. 

In the implementation of the CEC~2017 benchmark used in this study, we assume $F_i^*=100\cdot i$ (where $i = 1,3,4,\ldots, 30$). The function $F2$ is excluded because it cannot be evaluated in the baseline implementation. As a result, the proper test set consists of $29$ functions, for $i = 1,3,4,\ldots, 30$.

The activation of the transformation components illustrated in
Eq.~\eqref{eq:shift_rotation_bias} is described by the binary vector
$C = [C1, C2, C3]$. Its entries correspond to \texttt{bias}, \texttt{shift}, and
\texttt{rotation}, respectively.
A value of $1$ denotes the activation of a given transformation component $C1$, $C2$, or $C3$ according to the CEC~2017 benchmark (Awad et al. in \cite{NHA2016}). A value of $0$ denotes its deactivation. For the bias, this means omitting the component $F_i^*$ (i.e., setting $F_i^*=0$ where $i=1,3,4,\ldots,30$), for the shift, using a zero vector, and for the rotation, using an identity matrix with a $dim$ consistent with the given function.
Table~\ref{tab1} lists the 8 configurations used in the analysis.

\begin{table}[!t]
\centering
\caption{Transformation configurations in the parameterized CEC~2017 benchmark.}
\label{tab1}
\renewcommand{\arraystretch}{1.08}
\setlength{\tabcolsep}{3.8pt}
\setlength{\heavyrulewidth}{\lightrulewidth}
\begin{tabular}{@{}c l c c c l@{}}
\toprule
No. & Configuration & Bias & Shift & Rotation & Comparison purpose \\
\midrule
1. & $C(000)$ & $0$      & $0$   & $I$   & Baseline control; all transformations inactive \\
2. & $C(100)$ & $F_i^*$ & $0$   & $I$   & Isolated bias effect; shift and rotation disabled \\
3. & $C(010)$ & $0$      & $o_i$ & $I$   & Isolated shift effect; bias and rotation disabled \\
4. & $C(001)$ & $0$      & $0$   & $M_i$ & Isolated rotation effect; bias and shift disabled \\
5. & $C(110)$ & $F_i^*$ & $o_i$ & $I$   & Combined bias - shift effect; rotation disabled \\
6. & $C(101)$ & $F_i^*$ & $0$   & $M_i$ &  Combined bias - rotation effect; shift disabled\\
7. & $C(011)$ & $0$      & $o_i$ & $M_i$ & Combined shift - rotation effect; bias disabled \\
8. & $C(111)$ & $F_i^*$ & $o_i$ & $M_i$ & Full CEC~2017 setting \\
\bottomrule
\end{tabular}
\end{table}

The full configuration scheme $C$ ($2^3 = 8$), presented in Table~\ref{tab1}, is used as the organizing structure of the experimental analysis. In the CEC~2021 technical report, (see Mohamed et al. in \cite{AWM2020}), indicate that benchmark parameterization is intended to study combinations of bias, shift, and rotation. At the same time, the authors of the report emphasize that the presence of shift substantially changes the character of the test problem. When shift is not active, the optimum remains at the centre of the search space. This may favour algorithms directed toward the region close to zero. For this reason, in the CEC~2021 competition protocol, Mohamed et al. in \cite{AWM2020} removed some scenarios without shift. Variants without shift were recommended to be treated as a tool for studying the influence of shift, and not as a basis for the standard ranking assessment of algorithms \cite{AWM2020}. 

In this study, configurations without shift serve precisely this diagnostic role. They assess the effects of shift data and their interactions with external bias and rotation.
The remaining configurations serve as ablation variants (see Oszust et al. in \cite{MO2021}), with configuration $C(111)$ providing the baseline.

\Bem{Configuration $C(111)$ corresponds to the standard setting of the CEC - 2017 benchmark \cite{NHA2016}. In this variant, bias, shift, and rotation are active. The remaining configurations serve as ablation variants, enabling the influence of individual transformation components and their combinations on the behavior of hMPA to be assessed \cite{MO2021}.}  


The function indices of the \textbf{CEC 2017 \texttt{C++} implementation} (repository provided by Awad et al. in \cite{NHA2016}) are retained. For the hybrid functions $F11$-$F20$, the selected shift and rotation act on the full vector before the original permutation and partitioning. The same transformation order applies to the hybrid components of $F{29}$ and $F{30}$. For the composition functions $F21$–$F30$, the settings of $C2$ and $C3$ are shared by all components. These parameters select the original or zero shift vector and the original or identity rotation matrix for each component, respectively. The selected shift vectors also define the centres used to calculate composition weights. The component functions, evaluation order, partition sizes, scaling constants, internal component biases and aggregation rules remain unchanged. The weight-width parameters, weight normalisation rules, zero-distance handling and uniform-weight fallback are also retained. The parameter $C1$ controls only the external additive constant. The internal rules for applying shifts and rotations remain unchanged. Setting $C2 = 0$ replaces the shift data with zeros and also disables sign reversals  (the \texttt{if (Os[i] < 0.0)} branch aligns with CEC 2021 repository provided by Mohamed et al. in \cite{AWM2020}) conditioned on negative shift entries. This additional effect occurs in $F7$ \textbf{(Shifted and Rotated Lunacek Bi-Rastrigin's Function)} and in the Lunacek component of $F13$ \textbf{(Hybrid Function 3)}. The parameterised implementation uses the CEC 2017 (see Awad et al. in \cite{NHA2016}) reference evaluation procedure for $F6$ 
\textbf{(Shifted and Rotated Schaffer's $F7$ Function)}. This procedure uses pre-rotation coordinates. Therefore, for fixed input coordinates and fixed values of $C1$ and $C2$, the returned value of $F6$ is independent of $C3$.

\NOWE{\section{Experimental setup}}

The experiment was conducted for hMPA. We based the MPA implementation on the publicly available code of Faramarzi et al. in \cite{AF2020}, including population initialization, the marine memory mechanism, the three position-update phases, the FADs effect, and the parameters $FADs = 0.2$ and $P = 0.5$. We prepared the hMPA variant by integrating this implementation with the predicted-candidate operator described by Oszust, Sroka, and Cymerys in \cite{MO2021}.

We performed the optimization computations, result recording, and convergence plotting in MATLAB R2023b. The tests were run on a DELL computer with Windows 10 Home, version 22H2. The computer was equipped with an Intel Core i7 - 7500U processor with a base frequency of 2.70 GHz and 8 GB of RAM. The DSC and eDSC analyses were performed separately in \texttt{R} (version 4.5.3).

We based the implementation of the CEC~2017 functions on the definitions and C and MATLAB codes provided by Awad et al. in \cite{NHA2016} together with the technical report. To this implementation, we added our own parameterization layer, denoted by the prefixes \texttt{cec2017\_param\_} and \texttt{cec17\_param\_func\_}, which implemented the configurations $C$ without modifying the definitions of the benchmark functions CEC~2017 (Awad et al. in \cite{NHA2016}).

The population size was fixed at 20 individuals in all configurations, dimensions, and runs. The number of independent runs was 30, and the parameter of the hybrid operator was set to $newN_{opar} = 0.1$ (Oszust et al. in \cite{MO2021}). This means that $10\%$ of the population was replaced by predicted solution candidates. The fixed population size reduces additional experimental variability. It therefore allows differences in the results to be interpreted mainly with respect to the configuration $C$, dimensionality, and evaluation budget. 

Under a limit based on the number of objective function calls, a moderate population size reduces the cost of a single update cycle relative to larger populations. This makes it possible to observe more update cycles within the same $FES_{\max}$. This setting is consistent with the observation of Chen et al. in \cite{TCh2012} that increasing the population size does not necessarily improve the effectiveness of evolutionary algorithms, especially under a limited computational budget.

\NOWE{In summary,} we analyzed 29 CEC~2017 functions, eight configurations $C$, and four dimensions: $dim\in\{10,\ 30,\  50,\  100\}$. The basic variant was run under the adopted evaluation budget $FES_{\max} = 10000\cdot dim$. No additional independent limit on the number of generations was imposed. Each run terminated after the budget $FES_{\max}$. Thus, the main constraint of the experiment was the number of objective function evaluations, rather than an arbitrarily fixed number of iterations.

For each run, we recorded the best solution \texttt{outputsX}, the final objective-function value \texttt{outputsF}, and the convergence trajectory \texttt{outputsConv}.  These data were used to analyse the distributions of final solutions and final objective function values, as well as convergence dynamics. In the eDSC analysis (see Section \ref{sec7}), the final solution vectors were aligned by subtracting the corresponding reference optimum for each function and configuration. All values reported in this paper are based on our own experimental runs.

For one algorithm, the full basic experimental design includes:
$1.3224 \times 10^{10}$ objective function evaluations.

\vspace{0.25cm}

\section{Hybridization mechanism in hMPA}
\label{Hybr}
\NOWE{The notation hMPA used in this study follows the convention introduced in \cite{MO2021} by Oszust, Sroka, and Cymerys. It should not be confused with HMPA, introduced in \cite{SB2020} by Barshandeh, Piri, and Sangani. Their HMPA is a distinct hybrid multi-population algorithm that combines artificial ecosystem-based optimization with Harris hawks optimization. Throughout this study, hMPA refers exclusively to the predicted-candidate variant defined in \cite{MO2021}. The range of MPA variants and applications reviewed by Mugemanyi et al. in \cite{SM2023} provides relevant context for this study.

\Bem{The choice of MPA as the base algorithm is supported by three considerations:}
\vspace{0.2cm}
The selection of MPA as the base algorithm is motivated by three main considerations:
\begin{itemize}
\item Oszust et al. in \cite{MO2021} applied the predicted-candidate operator to $10$ population-based metaheuristics, including MPA. Their results showed that the operator can improve MPA. However, its effect depends on the test problem. This makes hMPA a nontrivial case for studying the response of an external hybridization mechanism.

\item Subsequent modifications of MPA confirm that the base algorithm still provides room for improvement. Hassan et al. in \cite{MHH2022} introduced a modified MPA based on comprehensive learning. The added mechanism was intended to improve population diversity and convergence. The method was evaluated on $28$ CEC~2017 functions in $50$ dimensions. It generally outperformed the original MPA.

\item The broader assessment conducted by Ma et al. in \cite{ZM2023} showed that MPA remained competitive among several recent nature-inspired algorithms. However, the best Differential Evolution (DE) variants, including LSHADE (SHADE with Linear Population Size Reduction)-based methods, were more consistently competitive on 
CEC~2017. Many nature-inspired algorithms introduced around $2019$--$2020$ also failed to surpass these established DE methods.
\end{itemize} 

Taken together, these findings support MPA as a capable, improvable base algorithm for assessing its response to an external hybridization mechanism.}
\Bem{Taken together, these findings justify the selection of MPA as a capable yet improvable base algorithm for examining the response of an external hybridization mechanism.}

The predicted-candidate solution operator is an external hybridization module. It does not replace the internal search rules of MPA. It is executed as an additional population-replacement step. The input data of the operator are the current population $X =\{x^{(1)},x^{(2)},$
$\ldots,x^{(n)}\}$ and the corresponding set of objective function values $F = \{f(x^{(1)}),f(x^{(2)}),$
$\ldots,f(x^{(n)})\}$, where $n$ denotes the population size and $x^{(j)}\in\mathbb{R}^{dim}$ is the $j$ - th individual. The operator uses only the pairs $(x^{(j)},f(x^{(j)}))$, for $j = 1,\ldots,n$. It does not require an explicit form of the objective function, gradient information, or an external problem model.

The mechanism of the operator $h$ consists of three stages. First, the target value $\tilde{f}$ is determined. Second, the corresponding decision vector is reconstructed by calibration. Third, the $g$ weakest individuals in the population are replaced. In the work of Oszust et al. in \cite{MO2021}, the target value is defined with respect to the mean objective function value in the current population: $\tilde{f} = \rho_t \bar{f}$,
$\bar{f}  =  \frac{1}{n}\sum_{j=1}^{n} f(x^{(j)})$,
$\rho_t \sim \mathcal{U}(a_t, b_t)$,
where $\rho_t$ is a random coefficient from the uniform distribution on the interval $(a_t, b_t)$ and $t$ is the current iteration of the algorithm. 

In the original scheme proposed \Bem{by Oszust et al.} in \cite{MO2021}, this interval successively takes the ranges $(0.5, 1.5)$, $(0.95, 1.05)$, and $(0.99, 1.01)$. The first range permits larger deviations from the population mean. It therefore supports broader exploration. The latter ranges concentrate the target value around the mean. They progressively strengthen the use of information accumulated in the population.

After $\tilde f$ has been determined, the calibration problem is solved. In standard evaluation, the decision vector $\textbf{x}$ is known and the value $f(\textbf{x})$ is computed. In the operator $h$, this direction is reversed. The target level $\tilde f$ is known, and the corresponding candidate $\tilde{x}$ is determined. For each coordinate, a separate regression model relates the values of this coordinate in the population to the objective function values. In the linear variant, it has the form:
$f(x^{(j)}) \approx \beta_{0,k} + \beta_{1,k}x^{(j)}_k$,
for $j = 1,\ldots,n$, where $x^{(j)}_k$ is the $k$-th coordinate of the $j$-th individual, and $\beta_{0,k}$ and $\beta_{1,k}$ are the model coefficients. These coefficients are estimated by the least squares method. The model is then inverted:
$\tilde{x}_k =  \frac{\tilde{f}-\widehat{\beta}_{0,k}}{\widehat{\beta}_{1,k}}$,
$k = 1,\ldots, dim$.
Repeating this procedure for all coordinates gives the predicted candidate:
$\tilde{x} =  (\tilde{x}_1,\tilde{x}_2,\ldots,\tilde{x}_{dim})$.

The predicted candidate is neither a copy of the best individual nor a simple random perturbation. \Bem{It is reconstructed from the current relationships between the positions of individuals in the decision space and their objective-function values.} \NOWE{ It is reconstructed based on the current relationships between the positions of individual agents in the decision space and the values of their objective functions.} The operator $h$ therefore uses information already accumulated in the population. 

In hMPA, as defined by Oszust et al. in \cite{MO2021}, the operator is executed following the standard MPA update and evaluation of the population. The individuals are then ordered according to the objective function value. The $g$ worst solutions are replaced with candidates obtained by calibration. In a minimization problem, these are the individuals with the highest 
objective-function values.
There are $g = \lfloor \text{newN}_{\text{opar}} \cdot n \rfloor$ of them, where the floor function is denoted by $\lfloor \cdot \rfloor$ and $\text{newN}_{\text{opar}}$ is the fraction of the population subject to replacement. The sensitivity analysis reported in \cite{MO2021} showed that high values of this parameter mainly worsened the hMPA results. The value adopted in this study is reported in the experimental protocol.
The remaining individuals preserve the result of the standard MPA cycle defined by Faramarzi et al. in \cite{AF2020}.

The relationship between the operator and the benchmark configurations $C$ follows from the construction of the operator $h$ (Oszust et al. in \cite{MO2021}) and the activation scheme of the bias, shift, and rotation components in the parameterized CEC~2017 variant (Awad et al. in \cite{NHA2016}). The individuals selected for replacement are determined from the ranking of 
objective-function values. Positive scaling of the function and adding a constant do not change this ranking. However, the calibration stage uses the numerical objective-function values and the mean $\bar f$. It may therefore respond to the activation of the bias even when the replacement ranking remains unchanged. The shift changes the location of the optimum with respect to the coordinate system. Rotation mixes the decision variables. The operator, in contrast, constructs a separate one-dimensional model for each variable. The configurations $C$ can therefore diagnose the empirical response of hMPA to changes in the objective level, optimum location, and coordinate dependencies.

\Bem{The additional overhead of the operator can be analyzed independently of the complexity of MPA.} \NOWE{The additional operator workload can be analyzed independently of the complexity of the MPA.} According to the analysis of Oszust, Sroka, and Cymerys in \cite{MO2021}, sorting the population has computational complexity $O(n\cdot\log n)$. Determining the target values for $g$ candidates requires $O(g \cdot n)$ operations. Generating predicted candidates by regression requires $O(g\cdot dim\cdot n\cdot u)$ operations, where $u$ denotes the degree of the regression model. Replacing the selected individuals has complexity $O(g\cdot dim)$. In the applied linear version, $u = 1$ is assumed. The additional cost therefore results from sorting, simple regression computations, and replacing part of the decision vectors. The operator changes the way in which part of the population is generated without requiring additional objective function calls.

\section{Statistical adaptation of DSC and eDSC}
\label{Sec:6}

\NOWE{The DSC and eDSC methods, developed by Eftimov and Koro{\v{s}}ec in \cite{TE2019}, were originally intended to compare stochastic optimization algorithms. In this study, they are deliberately adapted to compare benchmark configurations. The hMPA algorithm remains fixed. Each configuration yields repeated-run distributions on which the DSC and eDSC ranking procedures can operate. The resulting ranks therefore describe configuration-dependent differences in the results of hMPA, not differences between algorithms.}

The implementation was based on the DSC and eDSC scripts provided by the authors. The empirical-distribution comparison and ranking logic implemented in the authors' scripts was retained.
\NOWE{The input layer developed for the \textbf{Black-Box Benchmarking (BBOB)} output format has been replaced with procedures for reading results from MATLAB \texttt{.mat} files.}
\Bem{The input layer developed for the Black-Box Benchmarking (BBOB) output format was replaced with procedures for reading MATLAB \texttt{.mat} results.} 
The reconstruction of solution matrices, the scope of comparisons, the numerical handling of covariance matrices, and the reporting of results were also adapted. This adaptation was developed specifically for the complete set of 8-configurations analyzed in this study. To the best of our knowledge, this is the first application of the DSC and eDSC workflow to an exhaustive analysis of all 8: 
\textbf{bias, shift, rotation configurations} of a parameterized CEC~2017 benchmark (Awad et al. in \cite{NHA2016}).

In the DSC part (Eftimov and Koro{\v{s}}ec in \cite{TE2019}), the final objective-function values stored in \texttt{outputsF} were used. The \texttt{RankY} values were determined on this basis. In the eDSC part, the final solution vectors stored in \texttt{outputsX} were used. These data were reconstructed as matrices $X_{i,l}\in\mathbb{R}^{q\times dim}$, where $q$ is the number of independent runs, $dim$ denotes the problem dimensionality, $i$ denotes the test function, and $l$ denotes the analyzed comparison group. Each row represented one final solution in the decision space. Before the statistical analysis, the consistency of the number of runs and problem dimensionality was verified across the compared data sets. \Bem{The routines for the exhaustive three - configuration and control protocols additionally checked the completeness and range of the assigned ranks.} \NOWE{The procedures for the triplets of protocols, both configuration and control protocols, also verified the completeness and scope of the assigned ranks.} For each test function, the ranks were also verified to sum to $\frac{m \cdot (m+1)}{2}$, where $m$ is the number of compared configurations.

In the DSC part, pairwise Anderson-Darling tests (e.g., Engmann and Cousineau in \cite{SE2011}) were applied to the distributions of final objective function values. The resulting $p$-values were placed in a similarity matrix. 
The Bonferroni threshold $\alpha/\binom{m}{2}$ (where $\alpha=0.05$ denotes the nominal significance level, e.g., Eftimov and Koro{\v{s}}ec in \cite{TE2019}) was then applied.
After binarization of the similarity matrix, the transitivity of the resulting similarity relation was checked.

In the eDSC part \Bem{(Eftimov and Koro{\v{s}}ec in \cite{TE2019}),} the multivariate $\mathcal{E}$-test (e.g., Sz\'ekely and Rizzo in \cite{GJS2004}) was applied to the final solution vectors. The resulting $p$-values were processed using the same Bonferroni threshold and transitivity check as in the DSC part. Spatial dispersion was described using the covariance hypervolume of the final-solution distribution in the search space. The setting $\nu=0$ was adopted (so that smaller covariance hypervolumes were preferred in the spatial ranking). Therefore, \texttt{RankX} was interpreted as a measure of the compactness and repeatability of the final solutions. It was not interpreted as a measure rewarding greater exploratory dispersion.

For both ranking schemes, a transitive similarity relation partitioned the configurations into disjoint groups of statistically similar distributions. All configurations within a group received the same averaged rank. For eDSC, the hypervolume of each group was obtained from one member sampled uniformly at random (see Eftimov and Koro{\v{s}}ec in \cite{TE2019}). A reproducible seed was derived separately for each group from the problem dimension, function index and configuration labels. All group members received the resulting rank. In the case of DSC, the first-indexed member remained the deterministic representative (as in the implementation provided for DSC and eDSC, see Eftimov and Koro{\v{s}}ec in \cite{TE2019}).

If transitivity was not satisfied, ranks were assigned directly from the mean final objective-function values in DSC and from the covariance hypervolumes in eDSC. Averaged ranks were used for equal values. Lower values of \texttt{RankY} indicated better objective-value results, whereas lower values of \texttt{RankX} indicated more compact final-solution distributions.

Degenerate and rank-deficient covariance structures were handled explicitly. This treatment was particularly relevant for $dim \geq q$, where the empirical covariance matrix is necessarily rank deficient.
After calculating the empirical covariance matrix $\Sigma$, it was replaced by its symmetrized form:
$\Sigma\leftarrow(\Sigma+\Sigma^{T})/2$, with $T$ denoting transposition. The calculation was terminated if the matrix contained non-finite values. An absolute numerical tolerance of $10^{-10}$ was adopted (to suppress numerical artifacts 
in nearly degenerate covariance matrices). 
If $\max_{1\leq r,s\leq dim}\left|\Sigma_{rs}\right|<10^{-10}$,
the condition of numerically negligible spatial dispersion was satisfied. Here, $r$ and $s$ denote the row and column indices of $\Sigma$, respectively. The covariance hypervolume was then set to zero.

In the remaining cases, the nearest positive-definite approximation
$\widetilde{\Sigma}$ was obtained using the \texttt{R} (version 4.5.3) function
\texttt{Matrix::nearPD} with \texttt{corr = FALSE},
\texttt{keepDiag = FALSE}, \texttt{do2eigen = TRUE},
\texttt{eig.tol = $10^{-10}$}, \texttt{posd.tol = $10^{-10}$},
and \texttt{maxit = 1000}. 
The covariance hypervolume was then calculated as
$V=\sqrt{\prod_{r=1}^{dim}\lambda_r}$,
with $\lambda_r$ denoting the eigenvalues of
$\widetilde{\Sigma}$. Residual negative eigenvalues caused by
finite-precision arithmetic were set to zero before evaluating the product.

If the \texttt{Matrix::nearPD} function failed to converge, the analysis stopped and an error message was displayed. No alternative hypervolume measure or rank was substituted. Only the calculation of the covariance hypervolume was affected. The multivariate $\mathcal{E}$-test, the empirical distribution comparisons and the similarity relation remained unchanged. The DSC and eDSC ranking principles were also preserved.

We used three analysis protocols. The first compared all 8-configurations globally. The second covered all
$\binom{8}{3}=56$ three-configuration subsets. The third compared $C(000)$ with each remaining configuration. Unlike the illustrative eDSC experiments of Eftimov and Koro{\v{s}}ec in \cite{TE2019}, no 3-configuration subsets were sampled. The complete set of $56$ subsets was analyzed.

For each problem dimensionality, the \texttt{RankY} and \texttt{RankX} matrices from the global and exhaustive three-configuration protocols were analyzed across the test functions using the Friedman test. The resulting $p$-values were denoted by $p_{valueY}$ and $p_{valueX}$, respectively.

In the control protocol, $C(000)$ was compared with each remaining configuration across the test functions using the paired Wilcoxon signed-rank test. Asymptotic $p$-values were computed with \texttt{exact = FALSE}. If all paired rank differences were zero, $p$ was set to $1$. Seven pairwise $p$-values were obtained for each of \texttt{RankY} and \texttt{RankX} and adjusted separately using the Bonferroni correction.

For each ranking analysis, a combined significance value was also calculated, as defined by Eftimov and Koro{\v{s}}ec in \cite{TE2019}:
$p_{value}=1-\prod_{j=1}^{7}(1-p_j)$, where $p_j$ is the unadjusted $p$-value from the $j$-th Wilcoxon comparison. The resulting $p_{value}$ summarizes the seven comparisons against $C(000)$ and complements the Bonferroni-adjusted pairwise results.

The authors' eDSC script used $R=199$ resampling replications for each pairwise $\mathcal{E}$-test, whereas $R=4999$ was used in this study. This increased the nominal Monte Carlo (e.g., Fishman in \cite{GSF2013}) resolution of the estimated $p$-values from $1/(199+1)=0.005$ to $1/(4999+1)=0.0002$. The finer resolution allowed estimated $p$-values to fall below the Bonferroni threshold used in the global comparison of eight configurations, for which $\alpha/\binom{8}{2}=0.05/28\approx1.79\times10^{-3}$. The resampling seed was set to $1234$ before each test to ensure reproducibility.
All Anderson-Darling tests were completed successfully.

\section{Results}
\label{sec7}
This section presents the empirical results for all four problem
dimensionalities. The analysis begins with the global DSC and eDSC (e.g., Eftimov and Koro{\v{s}}ec in \cite{TE2019})
comparison of all 8-configurations, followed by the exhaustive
3-configuration and control protocols. The convergence plots then complement the distributional analysis by characterizing the optimization dynamics. The box plots further illustrate variability in final performance.

\subsection{Global DSC and eDSC results}
\label{sec7.1}
The global analysis first examines whether the eight configurations
differ across the 29 test functions. Table~\ref{tab:friedman}
reports the Friedman-test $p$-values (e.g., Derrac in \cite{JD2011}) for the DSC and eDSC rankings
at $dim\in\{10, \ 30, \ 50, \ 100\}$, computed using the \texttt{stats package} in \texttt{R} (version 4.5.3).

\begin{table}[ht]
\centering
\caption{Friedman test $p$-values for DSC and eDSC rankings across dimensions.}
\label{tab:friedman}
\begin{tabular}{cccccc}
\midrule
 & \multicolumn{2}{c}{\textbf{Algorithm: hMPA}} & & \multicolumn{2}{c}{\textbf{Algorithm: MPA}} \\
\cmidrule{2-3} \cmidrule{5-6}
$dim$ & DSC & eDSC & & DSC & eDSC \\
\midrule
10 & 3.900e-32 & 8.147e-22 & & 8.112E-31 & 8.020E-23 \\
30  & 4.568e-25 & 2.083e-29 & & 1.999E-25 & 6.854E-30\\
50  & 2.884e-26 & 2.517e-27 && 1.451E-25 & 4.412E-25 \\
100 & 1.560e-24 & 7.828e-24 & & 3.392E-28 & 5.863E-23\\
\midrule
\end{tabular}
\end{table}


As can be seen in Table~\ref{tab:friedman}, all the obtained $p$-values are strictly below 0.05 across all dimensions.
For both algorithms, the DSC $p$-values are lowest at $dim=10$. The eDSC $p$-values are lowest at $dim=30$, increasing at $dim=50$ and $dim=100$.

Tables~\ref{tab:dsc_ranks} and~\ref{tab:edsc_ranks} report the function-wise DSC and eDSC ranks together with their mean ranks.
\begin{table}[h!]
\setlength{\heavyrulewidth}{0.3pt}
\setlength{\lightrulewidth}{0.3pt}
\setlength{\cmidrulewidth}{0.3pt}
\centering
\scriptsize
\setlength{\tabcolsep}{6pt}
\caption{Function-wise DSC ranks. Lower values indicate better objective-value performance.}
\label{tab:dsc_ranks}
\resizebox{\linewidth}{!}{
\begin{tabular}{lcccccccc@{\hspace{0.75cm}}lcccccccc}
\toprule
\multicolumn{9}{c}{\textbf{Algorithm: hMPA}\hfill\textbf{$dim=10$}\hfill \qquad\qquad} & \multicolumn{9}{c}{\textbf{$dim=30$}} \\
\cmidrule(lr){1-9} \cmidrule(lr){10-18}
Functions & C(000) & C(001) & C(010) & C(011) & C(100) & C(101) & C(110) & C(111) & Functions & C(000) & C(001) & C(010) & C(011) & C(100) & C(101) & C(110) & C(111)
\\
\midrule
F1 & 2.00 & 2.00 & 5.00 & 7.00 & 2.00 & 4.00 & 6.00 & 8.00 & F1 & 2.50 & 2.50 & 6.00 & 8.00 & 2.50 & 2.50 & 5.00 & 7.00
\\
F3 & 2.50 & 2.50 & 5.50 & 5.50 & 2.50 & 2.50 & 7.50 & 7.50 & F3 & 2.50 & 2.50 & 5.00 & 8.00 & 2.50 & 2.50 & 6.00 & 7.00
\\
F4 & 2.50 & 2.50 & 5.50 & 5.50 & 2.50 & 2.50 & 7.50 & 7.50 & F4 & 2.50 & 2.50 & 5.00 & 8.00 & 2.50 & 2.50 & 6.00 & 7.00
\\
F5 & 2.50 & 2.50 & 5.50 & 7.50 & 2.50 & 2.50 & 5.50 & 7.50 & F5 & 2.50 & 2.50 & 5.50 & 7.50 & 2.50 & 2.50 & 5.50 & 7.50
\\
F6 & 2.50 & 2.50 & 6.50 & 6.50 & 2.50 & 2.50 & 6.50 & 6.50 & F6 & 2.50 & 2.50 & 6.50 & 6.50 & 2.50 & 2.50 & 6.50 & 6.50
\\
F7 & 2.50 & 2.50 & 5.50 & 7.50 & 2.50 & 2.50 & 5.50 & 7.50 & F7 & 2.50 & 2.50 & 5.50 & 7.50 & 2.50 & 2.50 & 5.50 & 7.50
\\
F8 & 2.50 & 2.50 & 5.50 & 7.50 & 2.50 & 2.50 & 5.50 & 7.50 & F8 & 2.50 & 2.50 & 5.50 & 7.50 & 2.50 & 2.50 & 5.50 & 7.50
\\
F9 & 1.00 & 3.00 & 4.00 & 8.00 & 2.00 & 6.00 & 5.00 & 7.00 & F9 & 1.00 & 6.00 & 5.00 & 7.00 & 2.00 & 3.00 & 4.00 & 8.00
\\
F10 & 1.50 & 3.50 & 5.50 & 7.50 & 1.50 & 3.50 & 5.50 & 7.50 & F10 & 2.50 & 2.50 & 5.50 & 7.50 & 2.50 & 2.50 & 5.50 & 7.50
\\
F11 & 2.50 & 2.50 & 5.00 & 7.50 & 2.50 & 2.50 & 6.00 & 7.50 & F11 & 2.50 & 2.50 & 5.50 & 7.50 & 2.50 & 2.50 & 5.50 & 7.50
\\
F12 & 1.50 & 4.50 & 4.50 & 7.50 & 1.50 & 4.50 & 4.50 & 7.50 & F12 & 1.50 & 3.50 & 5.50 & 7.50 & 1.50 & 3.50 & 5.50 & 7.50
\\
F13 & 1.50 & 3.00 & 6.00 & 5.00 & 1.50 & 4.00 & 8.00 & 7.00 & F13 & 2.50 & 2.50 & 5.50 & 7.50 & 2.50 & 2.50 & 5.50 & 7.50
\\
F14 & 1.50 & 5.00 & 4.00 & 8.00 & 1.50 & 6.00 & 3.00 & 7.00 & F14 & 1.50 & 5.50 & 3.50 & 7.50 & 1.50 & 5.50 & 3.50 & 7.50
\\
F15 & 1.00 & 6.00 & 3.00 & 7.00 & 2.00 & 5.00 & 4.00 & 8.00 & F15 & 1.00 & 5.00 & 3.00 & 7.00 & 2.00 & 6.00 & 4.00 & 8.00
\\
F16 & 2.50 & 2.50 & 6.50 & 6.50 & 2.50 & 2.50 & 6.50 & 6.50 & F16 & 3.50 & 1.50 & 5.50 & 7.50 & 3.50 & 1.50 & 5.50 & 7.50
\\
F17 & 2.50 & 2.50 & 5.50 & 7.50 & 2.50 & 2.50 & 5.50 & 7.50 & F17 & 2.50 & 2.50 & 5.50 & 7.50 & 2.50 & 2.50 & 5.50 & 7.50
\\
F18 & 1.00 & 5.00 & 3.00 & 7.00 & 2.00 & 6.00 & 4.00 & 8.00 & F18 & 1.50 & 5.50 & 3.50 & 7.50 & 1.50 & 5.50 & 3.50 & 7.50
\\
F19 & 1.50 & 3.00 & 6.00 & 6.00 & 1.50 & 6.00 & 6.00 & 6.00 & F19 & 1.50 & 3.50 & 5.50 & 7.50 & 1.50 & 3.50 & 5.50 & 7.50
\\
F20 & 2.50 & 2.50 & 5.50 & 7.50 & 2.50 & 2.50 & 5.50 & 7.50 & F20 & 2.50 & 2.50 & 5.50 & 7.50 & 2.50 & 2.50 & 5.50 & 7.50
\\
F21 & 1.50 & 1.50 & 6.50 & 6.50 & 3.50 & 3.50 & 6.50 & 6.50 & F21 & 1.50 & 1.50 & 5.00 & 7.50 & 3.50 & 3.50 & 6.00 & 7.50
\\
F22 & 1.50 & 1.50 & 6.50 & 6.50 & 3.50 & 3.50 & 6.50 & 6.50 & F22 & 1.50 & 1.50 & 7.00 & 8.00 & 3.50 & 3.50 & 5.00 & 6.00
\\
F23 & 1.00 & 4.00 & 5.00 & 8.00 & 2.00 & 3.00 & 6.00 & 7.00 & F23 & 1.50 & 7.50 & 3.50 & 5.50 & 1.50 & 7.50 & 3.50 & 5.50
\\
F24 & 2.50 & 2.50 & 6.50 & 6.50 & 2.50 & 2.50 & 6.50 & 6.50 & F24 & 2.50 & 2.50 & 5.50 & 7.50 & 2.50 & 2.50 & 5.50 & 7.50
\\
F25 & 2.50 & 2.50 & 5.50 & 7.50 & 2.50 & 2.50 & 5.50 & 7.50 & F25 & 2.50 & 2.50 & 7.50 & 5.50 & 2.50 & 2.50 & 7.50 & 5.50
\\
F26 & 2.50 & 2.50 & 5.50 & 7.50 & 2.50 & 2.50 & 5.50 & 7.50 & F26 & 2.50 & 2.50 & 6.50 & 6.50 & 2.50 & 2.50 & 6.50 & 6.50
\\
F27 & 3.50 & 1.50 & 7.50 & 5.50 & 3.50 & 1.50 & 7.50 & 5.50 & F27 & 7.50 & 5.50 & 1.50 & 3.50 & 7.50 & 5.50 & 1.50 & 3.50
\\
F28 & 2.50 & 2.50 & 7.50 & 5.50 & 2.50 & 2.50 & 7.50 & 5.50 & F28 & 2.50 & 2.50 & 7.50 & 5.50 & 2.50 & 2.50 & 7.50 & 5.50
\\
F29 & 1.00 & 4.00 & 6.00 & 8.00 & 2.00 & 3.00 & 5.00 & 7.00 & F29 & 1.50 & 7.50 & 3.50 & 5.50 & 1.50 & 7.50 & 3.50 & 5.50
\\
F30 & 2.00 & 3.50 & 5.50 & 7.50 & 1.00 & 3.50 & 5.50 & 7.50 & F30 & 1.50 & 3.50 & 7.50 & 5.50 & 1.50 & 3.50 & 7.50 & 5.50
\\
\midrule
Mean rank & 2.00 & 2.97 & 5.50 & 6.93 & 2.28 & 3.38 & 5.84 & 7.10 & Mean rank & 2.29 & 3.36 & 5.28 & 6.98 & 2.50 & 3.43 & 5.28 & 6.88
\\
\bottomrule
\end{tabular}
}
\end{table}

\begin{table}
\vspace{0.4cm}
\centering\noindent\textit{Table~\ref{tab:dsc_ranks} (continued).}\par\vspace{2pt}
\resizebox{\linewidth}{!}{
\begin{tabular}{lcccccccc@{\hspace{0.75cm}}lcccccccc}
\toprule
\multicolumn{9}{c}{\textbf{Algorithm: hMPA}\hfill\textbf{$dim=50$}\hfill \qquad\qquad} & \multicolumn{9}{c}{\rlap\textbf{$dim=100$}} \\
\cmidrule(lr){1-9} \cmidrule(lr){10-18}
Functions & C(000) & C(001) & C(010) & C(011) & C(100) & C(101) & C(110) & C(111) & Functions & C(000) & C(001) & C(010) & C(011) & C(100) & C(101) & C(110) & C(111)
\\
\midrule
F1 & 2.50 & 2.50 & 6.50 & 6.50 & 2.50 & 2.50 & 6.50 & 6.50 & F1 & 2.50 & 2.50 & 5.50 & 7.50 & 2.50 & 2.50 & 5.50 & 7.50
\\
F3 & 2.50 & 2.50 & 5.50 & 7.50 & 2.50 & 2.50 & 5.50 & 7.50 & F3 & 2.50 & 2.50 & 5.50 & 7.50 & 2.50 & 2.50 & 5.50 & 7.50
\\
F4 & 2.50 & 2.50 & 5.00 & 7.00 & 2.50 & 2.50 & 6.00 & 8.00 & F4 & 2.50 & 2.50 & 5.50 & 7.50 & 2.50 & 2.50 & 5.50 & 7.50
\\
F5 & 2.50 & 2.50 & 5.50 & 7.50 & 2.50 & 2.50 & 5.50 & 7.50 & F5 & 2.50 & 2.50 & 5.50 & 7.50 & 2.50 & 2.50 & 5.50 & 7.50
\\
F6 & 2.50 & 2.50 & 6.50 & 6.50 & 2.50 & 2.50 & 6.50 & 6.50 & F6 & 2.50 & 2.50 & 6.50 & 6.50 & 2.50 & 2.50 & 6.50 & 6.50
\\
F7 & 2.50 & 2.50 & 5.50 & 7.50 & 2.50 & 2.50 & 5.50 & 7.50 & F7 & 2.50 & 2.50 & 5.50 & 7.50 & 2.50 & 2.50 & 5.50 & 7.50
\\
F8 & 2.50 & 2.50 & 5.50 & 7.50 & 2.50 & 2.50 & 5.50 & 7.50 & F8 & 2.50 & 2.50 & 5.50 & 7.50 & 2.50 & 2.50 & 5.50 & 7.50
\\
F9 & 1.00 & 3.50 & 5.50 & 7.50 & 2.00 & 3.50 & 5.50 & 7.50 & F9 & 1.50 & 3.50 & 5.50 & 7.50 & 1.50 & 3.50 & 5.50 & 7.50
\\
F10 & 2.50 & 2.50 & 5.50 & 7.50 & 2.50 & 2.50 & 5.50 & 7.50 & F10 & 2.50 & 2.50 & 5.50 & 7.50 & 2.50 & 2.50 & 5.50 & 7.50
\\
F11 & 2.50 & 2.50 & 5.50 & 7.50 & 2.50 & 2.50 & 5.50 & 7.50 & F11 & 2.50 & 2.50 & 5.50 & 7.50 & 2.50 & 2.50 & 5.50 & 7.50
\\
F12 & 2.50 & 2.50 & 5.50 & 7.50 & 2.50 & 2.50 & 5.50 & 7.50 & F12 & 2.50 & 2.50 & 5.50 & 7.50 & 2.50 & 2.50 & 5.50 & 7.50
\\
F13 & 2.50 & 2.50 & 5.50 & 7.50 & 2.50 & 2.50 & 5.50 & 7.50 & F13 & 2.50 & 2.50 & 6.00 & 8.00 & 2.50 & 2.50 & 5.00 & 7.00
\\
F14 & 1.50 & 5.50 & 3.50 & 7.50 & 1.50 & 5.50 & 3.50 & 7.50 & F14 & 1.50 & 5.50 & 3.50 & 7.50 & 1.50 & 5.50 & 3.50 & 7.50
\\
F15 & 1.00 & 2.00 & 5.00 & 7.00 & 4.00 & 3.00 & 6.00 & 8.00 & F15 & 2.50 & 2.50 & 5.50 & 7.50 & 2.50 & 2.50 & 5.50 & 7.50
\\
F16 & 3.00 & 1.00 & 5.00 & 7.00 & 4.00 & 2.00 & 6.00 & 8.00 & F16 & 2.00 & 1.00 & 5.00 & 8.00 & 4.00 & 3.00 & 6.00 & 7.00
\\
F17 & 2.50 & 2.50 & 5.50 & 7.50 & 2.50 & 2.50 & 5.50 & 7.50 & F17 & 2.50 & 2.50 & 5.50 & 7.50 & 2.50 & 2.50 & 5.50 & 7.50
\\
F18 & 1.50 & 5.50 & 3.00 & 7.50 & 1.50 & 5.50 & 4.00 & 7.50 & F18 & 1.50 & 5.50 & 3.50 & 8.00 & 1.50 & 5.50 & 3.50 & 7.00
\\
F19 & 2.50 & 2.50 & 5.50 & 7.50 & 2.50 & 2.50 & 5.50 & 7.50 & F19 & 2.50 & 2.50 & 5.50 & 7.50 & 2.50 & 2.50 & 5.50 & 7.50
\\
F20 & 2.00 & 3.00 & 4.00 & 7.00 & 1.00 & 5.00 & 6.00 & 8.00 & F20 & 1.50 & 3.50 & 5.50 & 7.50 & 1.50 & 3.50 & 5.50 & 7.50
\\
F21 & 1.50 & 1.50 & 5.50 & 7.50 & 3.50 & 3.50 & 5.50 & 7.50 & F21 & 1.50 & 1.50 & 5.50 & 7.50 & 3.50 & 3.50 & 5.50 & 7.50
\\
F22 & 3.50 & 3.50 & 5.00 & 8.00 & 1.50 & 1.50 & 6.00 & 7.00 & F22 & 1.50 & 1.50 & 5.50 & 7.50 & 3.50 & 3.50 & 5.50 & 7.50
\\
F23 & 1.50 & 7.50 & 3.50 & 5.50 & 1.50 & 7.50 & 3.50 & 5.50 & F23 & 5.50 & 7.50 & 1.50 & 3.50 & 5.50 & 7.50 & 1.50 & 3.50
\\
F24 & 2.50 & 2.50 & 5.50 & 7.50 & 2.50 & 2.50 & 5.50 & 7.50 & F24 & 2.50 & 2.50 & 5.50 & 7.50 & 2.50 & 2.50 & 5.50 & 7.50
\\
F25 & 2.50 & 2.50 & 6.50 & 6.50 & 2.50 & 2.50 & 6.50 & 6.50 & F25 & 2.50 & 2.50 & 5.50 & 7.50 & 2.50 & 2.50 & 5.50 & 7.50
\\
F26 & 2.50 & 2.50 & 5.50 & 7.50 & 2.50 & 2.50 & 5.50 & 7.50 & F26 & 2.50 & 2.50 & 5.50 & 7.50 & 2.50 & 2.50 & 5.50 & 7.50
\\
F27 & 7.50 & 5.50 & 1.50 & 3.50 & 7.50 & 5.50 & 1.50 & 3.50 & F27 & 7.50 & 5.50 & 1.50 & 3.50 & 7.50 & 5.50 & 1.50 & 3.50
\\
F28 & 2.50 & 2.50 & 6.50 & 6.50 & 2.50 & 2.50 & 6.50 & 6.50 & F28 & 2.50 & 2.50 & 7.50 & 5.50 & 2.50 & 2.50 & 7.50 & 5.50
\\
F29 & 3.50 & 7.50 & 1.50 & 5.50 & 3.50 & 7.50 & 1.50 & 5.50 & F29 & 3.50 & 7.50 & 1.50 & 5.50 & 3.50 & 7.50 & 1.50 & 5.50
\\
F30 & 2.50 & 2.50 & 5.50 & 7.50 & 2.50 & 2.50 & 5.50 & 7.50 & F30 & 2.50 & 2.50 & 5.50 & 7.50 & 2.50 & 2.50 & 5.50 & 7.50
\\
\midrule
Mean rank & 2.50 & 3.14 & 5.02 & 7.03 & 2.64 & 3.28 & 5.26 & 7.14 & Mean rank & 2.59 & 3.10 & 5.05 & 7.10 & 2.79 & 3.31 & 5.05 & 7.00
\\
\bottomrule
\end{tabular}
}

\vspace{0.4cm}

\resizebox{\linewidth}{!}{
\begin{tabular}{lcccccccc@{\hspace{0.75cm}}lcccccccc}
\toprule
\multicolumn{9}{c}{\textbf{Algorithm: MPA}\hfill\textbf{$dim=10$}\hfill \qquad\qquad} & \multicolumn{9}{c}{\textbf{$dim=30$}} \\
\cmidrule(lr){1-9} \cmidrule(lr){10-18}
Functions & C(000) & C(001) & C(010) & C(011) & C(100) & C(101) & C(110) & C(111) & Functions & C(000) & C(001) & C(010) & C(011) & C(100) & C(101) & C(110) & C(111)
\\
\midrule
F1 & 3.00 & 4.00 & 6.00 & 7.00 & 1.50 & 1.50 & 5.00 & 8.00 & F1 & 2.00 & 4.00 & 5.00 & 7.00 & 2.00 & 2.00 & 6.00 & 8.00
\\
F3 & 4.00 & 3.00 & 7.50 & 7.50 & 1.50 & 1.50 & 5.50 & 5.50 & F3 & 3.00 & 4.00 & 5.50 & 7.50 & 1.50 & 1.50 & 5.50 & 7.50
\\
F4 & 2.50 & 2.50 & 6.50 & 6.50 & 2.50 & 2.50 & 6.50 & 6.50 & F4 & 2.50 & 2.50 & 5.50 & 7.50 & 2.50 & 2.50 & 5.50 & 7.50
\\
F5 & 2.50 & 2.50 & 5.50 & 7.50 & 2.50 & 2.50 & 5.50 & 7.50 & F5 & 2.50 & 2.50 & 5.50 & 7.50 & 2.50 & 2.50 & 5.50 & 7.50
\\
F6 & 3.50 & 3.50 & 6.50 & 6.50 & 1.50 & 1.50 & 6.50 & 6.50 & F6 & 2.50 & 2.50 & 6.50 & 6.50 & 2.50 & 2.50 & 6.50 & 6.50
\\
F7 & 2.50 & 2.50 & 5.50 & 7.50 & 2.50 & 2.50 & 5.50 & 7.50 & F7 & 2.50 & 2.50 & 5.50 & 7.50 & 2.50 & 2.50 & 5.50 & 7.50
\\
F8 & 2.50 & 2.50 & 5.50 & 7.50 & 2.50 & 2.50 & 5.50 & 7.50 & F8 & 2.50 & 2.50 & 5.50 & 7.50 & 2.50 & 2.50 & 5.50 & 7.50
\\
F9 & 1.00 & 7.00 & 3.00 & 8.00 & 2.00 & 5.00 & 4.00 & 6.00 & F9 & 2.00 & 5.00 & 4.00 & 7.00 & 1.00 & 3.00 & 6.00 & 8.00
\\
F10 & 1.50 & 3.50 & 5.50 & 7.50 & 1.50 & 3.50 & 5.50 & 7.50 & F10 & 2.50 & 2.50 & 5.50 & 7.50 & 2.50 & 2.50 & 5.50 & 7.50
\\
F11 & 3.00 & 4.00 & 5.50 & 7.50 & 1.50 & 1.50 & 5.50 & 7.50 & F11 & 2.50 & 2.50 & 5.50 & 7.50 & 2.50 & 2.50 & 5.50 & 7.50
\\
F12 & 2.00 & 4.00 & 3.00 & 7.00 & 1.00 & 5.00 & 6.00 & 8.00 & F12 & 2.50 & 4.00 & 5.50 & 7.50 & 1.00 & 2.50 & 5.50 & 7.50
\\
F13 & 2.00 & 4.00 & 8.00 & 7.00 & 1.00 & 3.00 & 5.00 & 6.00 & F13 & 2.50 & 2.50 & 5.50 & 7.50 & 2.50 & 2.50 & 5.50 & 7.50
\\
F14 & 2.00 & 5.50 & 3.50 & 7.50 & 1.00 & 5.50 & 3.50 & 7.50 & F14 & 1.50 & 5.50 & 3.50 & 7.50 & 1.50 & 5.50 & 3.50 & 7.50
\\
F15 & 1.50 & 5.50 & 5.50 & 5.50 & 1.50 & 5.50 & 5.50 & 5.50 & F15 & 2.50 & 2.50 & 5.50 & 7.50 & 2.50 & 2.50 & 5.50 & 7.50
\\
F16 & 2.50 & 2.50 & 5.50 & 7.50 & 2.50 & 2.50 & 5.50 & 7.50 & F16 & 1.50 & 3.50 & 5.50 & 7.50 & 1.50 & 3.50 & 5.50 & 7.50
\\
F17 & 2.50 & 2.50 & 5.50 & 7.50 & 2.50 & 2.50 & 5.50 & 7.50 & F17 & 1.50 & 3.00 & 5.00 & 7.00 & 1.50 & 4.00 & 6.00 & 8.00
\\
F18 & 1.00 & 6.00 & 3.00 & 7.00 & 2.00 & 5.00 & 4.00 & 8.00 & F18 & 1.50 & 5.50 & 3.50 & 7.50 & 1.50 & 5.50 & 3.50 & 7.50
\\
F19 & 2.00 & 5.50 & 5.50 & 5.50 & 1.00 & 5.50 & 5.50 & 5.50 & F19 & 1.50 & 3.50 & 5.50 & 7.50 & 1.50 & 3.50 & 5.50 & 7.50
\\
F20 & 3.00 & 4.00 & 5.00 & 7.00 & 1.50 & 1.50 & 6.00 & 8.00 & F20 & 2.00 & 4.00 & 5.00 & 8.00 & 1.00 & 3.00 & 6.00 & 7.00
\\
F21 & 1.50 & 1.50 & 6.50 & 6.50 & 3.50 & 3.50 & 6.50 & 6.50 & F21 & 1.50 & 1.50 & 5.50 & 7.50 & 3.50 & 3.50 & 5.50 & 7.50
\\
F22 & 1.50 & 1.50 & 6.50 & 6.50 & 3.50 & 3.50 & 6.50 & 6.50 & F22 & 1.50 & 1.50 & 7.00 & 6.00 & 3.50 & 3.50 & 8.00 & 5.00
\\
F23 & 1.50 & 3.50 & 5.50 & 7.50 & 1.50 & 3.50 & 5.50 & 7.50 & F23 & 1.50 & 7.50 & 3.50 & 5.50 & 1.50 & 7.50 & 3.50 & 5.50
\\
F24 & 2.50 & 2.50 & 6.50 & 6.50 & 2.50 & 2.50 & 6.50 & 6.50 & F24 & 2.50 & 2.50 & 5.50 & 7.50 & 2.50 & 2.50 & 5.50 & 7.50
\\
F25 & 2.50 & 2.50 & 7.50 & 5.50 & 2.50 & 2.50 & 7.50 & 5.50 & F25 & 2.50 & 2.50 & 7.50 & 5.50 & 2.50 & 2.50 & 7.50 & 5.50
\\
F26 & 2.50 & 2.50 & 5.50 & 7.50 & 2.50 & 2.50 & 5.50 & 7.50 & F26 & 2.50 & 2.50 & 6.50 & 6.50 & 2.50 & 2.50 & 6.50 & 6.50
\\
F27 & 3.00 & 2.00 & 8.00 & 5.00 & 4.00 & 1.00 & 7.00 & 6.00 & F27 & 7.50 & 5.50 & 1.50 & 3.50 & 7.50 & 5.50 & 1.50 & 3.50
\\
F28 & 2.50 & 2.50 & 5.50 & 7.50 & 2.50 & 2.50 & 5.50 & 7.50 & F28 & 3.00 & 1.00 & 8.00 & 5.00 & 4.00 & 2.00 & 7.00 & 6.00
\\
F29 & 1.50 & 3.50 & 5.50 & 7.50 & 1.50 & 3.50 & 5.50 & 7.50 & F29 & 1.50 & 7.50 & 3.50 & 5.50 & 1.50 & 7.50 & 3.50 & 5.50
\\
F30 & 1.50 & 3.50 & 5.50 & 7.50 & 1.50 & 3.50 & 5.50 & 7.50 & F30 & 1.50 & 3.50 & 7.50 & 5.50 & 1.50 & 3.50 & 7.50 & 5.50
\\
\midrule
Mean rank & 2.24 & 3.45 & 5.66 & 6.97 & 2.03 & 3.07 & 5.62 & 6.97 & Mean rank & 2.31 & 3.45 & 5.33 & 6.84 & 2.31 & 3.34 & 5.50 & 6.91
\\
\bottomrule
\end{tabular}
}
\end{table}

\vspace{0.4cm}
\begin{table}
\centering
\noindent\textit{Table~\ref{tab:dsc_ranks} (continued).}\par\vspace{2pt}
\resizebox{\linewidth}{!}{
\begin{tabular}{lcccccccc@{\hspace{0.75cm}}lcccccccc}
\toprule
\multicolumn{9}{c}{\textbf{Algorithm: MPA}\hfill\textbf{$dim=50$}\hfill \qquad\qquad} & \multicolumn{9}{c}{\textbf{$dim=100$}} \\
\cmidrule(lr){1-9} \cmidrule(lr){10-18}
Functions & C(000) & C(001) & C(010) & C(011) & C(100) & C(101) & C(110) & C(111) & Functions & C(000) & C(001) & C(010) & C(011) & C(100) & C(101) & C(110) & C(111)
\\
\midrule
F1 & 2.50 & 2.50 & 6.50 & 6.50 & 2.50 & 2.50 & 6.50 & 6.50 & F1 & 2.50 & 2.50 & 5.50 & 7.50 & 2.50 & 2.50 & 5.50 & 7.50
\\
F3 & 3.00 & 4.00 & 5.50 & 7.50 & 1.50 & 1.50 & 5.50 & 7.50 & F3 & 3.00 & 4.00 & 5.50 & 7.50 & 1.50 & 1.50 & 5.50 & 7.50
\\
F4 & 2.50 & 2.50 & 5.00 & 7.00 & 2.50 & 2.50 & 6.00 & 8.00 & F4 & 2.50 & 2.50 & 6.50 & 6.50 & 2.50 & 2.50 & 6.50 & 6.50
\\
F5 & 2.50 & 2.50 & 5.50 & 7.50 & 2.50 & 2.50 & 5.50 & 7.50 & F5 & 2.50 & 2.50 & 5.50 & 7.50 & 2.50 & 2.50 & 5.50 & 7.50
\\
F6 & 2.50 & 2.50 & 6.50 & 6.50 & 2.50 & 2.50 & 6.50 & 6.50 & F6 & 2.50 & 2.50 & 6.50 & 6.50 & 2.50 & 2.50 & 6.50 & 6.50
\\
F7 & 2.50 & 2.50 & 5.50 & 7.50 & 2.50 & 2.50 & 5.50 & 7.50 & F7 & 2.50 & 2.50 & 5.50 & 7.50 & 2.50 & 2.50 & 5.50 & 7.50
\\
F8 & 2.50 & 2.50 & 5.50 & 7.50 & 2.50 & 2.50 & 5.50 & 7.50 & F8 & 2.50 & 2.50 & 5.50 & 7.50 & 2.50 & 2.50 & 5.50 & 7.50
\\
F9 & 1.50 & 3.50 & 5.50 & 7.50 & 1.50 & 3.50 & 5.50 & 7.50 & F9 & 1.50 & 3.50 & 5.50 & 7.50 & 1.50 & 3.50 & 5.50 & 7.50
\\
F10 & 2.50 & 2.50 & 5.50 & 7.50 & 2.50 & 2.50 & 5.50 & 7.50 & F10 & 2.50 & 2.50 & 5.50 & 7.50 & 2.50 & 2.50 & 5.50 & 7.50
\\
F11 & 2.50 & 2.50 & 5.50 & 7.50 & 2.50 & 2.50 & 5.50 & 7.50 & F11 & 2.50 & 2.50 & 5.50 & 7.50 & 2.50 & 2.50 & 5.50 & 7.50
\\
F12 & 2.00 & 4.00 & 5.50 & 7.50 & 2.00 & 2.00 & 5.50 & 7.50 & F12 & 2.50 & 2.50 & 5.50 & 7.50 & 2.50 & 2.50 & 5.50 & 7.50
\\
F13 & 2.50 & 2.50 & 5.50 & 7.50 & 2.50 & 2.50 & 5.50 & 7.50 & F13 & 2.50 & 2.50 & 6.50 & 6.50 & 2.50 & 2.50 & 6.50 & 6.50
\\
F14 & 1.50 & 5.50 & 3.50 & 7.50 & 1.50 & 5.50 & 3.50 & 7.50 & F14 & 1.50 & 5.50 & 3.50 & 7.50 & 1.50 & 5.50 & 3.50 & 7.50
\\
F15 & 2.50 & 2.50 & 5.50 & 7.50 & 2.50 & 2.50 & 5.50 & 7.50 & F15 & 2.50 & 2.50 & 5.50 & 7.50 & 2.50 & 2.50 & 5.50 & 7.50
\\
F16 & 3.50 & 1.50 & 5.50 & 7.50 & 3.50 & 1.50 & 5.50 & 7.50 & F16 & 3.00 & 2.00 & 6.00 & 8.00 & 4.00 & 1.00 & 5.00 & 7.00
\\
F17 & 2.50 & 2.50 & 5.50 & 7.50 & 2.50 & 2.50 & 5.50 & 7.50 & F17 & 2.50 & 2.50 & 5.50 & 7.50 & 2.50 & 2.50 & 5.50 & 7.50
\\
F18 & 1.50 & 5.50 & 3.50 & 7.50 & 1.50 & 5.50 & 3.50 & 7.50 & F18 & 1.50 & 5.50 & 3.50 & 7.50 & 1.50 & 5.50 & 3.50 & 7.50
\\
F19 & 2.50 & 2.50 & 5.50 & 7.50 & 2.50 & 2.50 & 5.50 & 7.50 & F19 & 2.50 & 2.50 & 5.50 & 7.50 & 2.50 & 2.50 & 5.50 & 7.50
\\
F20 & 1.00 & 5.00 & 3.00 & 8.00 & 2.00 & 6.00 & 4.00 & 7.00 & F20 & 2.50 & 2.50 & 5.50 & 7.50 & 2.50 & 2.50 & 5.50 & 7.50
\\
F21 & 1.50 & 1.50 & 5.50 & 7.50 & 3.50 & 3.50 & 5.50 & 7.50 & F21 & 1.00 & 4.00 & 5.50 & 7.50 & 2.50 & 2.50 & 5.50 & 7.50
\\
F22 & 3.50 & 3.50 & 5.00 & 7.00 & 1.50 & 1.50 & 6.00 & 8.00 & F22 & 1.50 & 1.50 & 5.50 & 7.50 & 3.50 & 3.50 & 5.50 & 7.50
\\
F23 & 1.50 & 7.50 & 3.50 & 5.50 & 1.50 & 7.50 & 3.50 & 5.50 & F23 & 2.50 & 2.50 & 5.50 & 7.50 & 2.50 & 2.50 & 5.50 & 7.50
\\
F24 & 2.50 & 2.50 & 5.50 & 7.50 & 2.50 & 2.50 & 5.50 & 7.50 & F24 & 2.50 & 2.50 & 5.50 & 7.50 & 2.50 & 2.50 & 5.50 & 7.50
\\
F25 & 2.50 & 2.50 & 6.50 & 6.50 & 2.50 & 2.50 & 6.50 & 6.50 & F25 & 2.50 & 2.50 & 5.50 & 7.50 & 2.50 & 2.50 & 5.50 & 7.50
\\
F26 & 2.50 & 2.50 & 7.00 & 8.00 & 2.50 & 2.50 & 6.00 & 5.00 & F26 & 2.50 & 2.50 & 5.50 & 7.50 & 2.50 & 2.50 & 5.50 & 7.50
\\
F27 & 7.50 & 5.50 & 1.50 & 3.50 & 7.50 & 5.50 & 1.50 & 3.50 & F27 & 7.50 & 5.50 & 1.50 & 3.50 & 7.50 & 5.50 & 1.50 & 3.50
\\
F28 & 2.50 & 2.50 & 6.50 & 6.50 & 2.50 & 2.50 & 6.50 & 6.50 & F28 & 2.50 & 2.50 & 7.50 & 5.50 & 2.50 & 2.50 & 7.50 & 5.50
\\
F29 & 3.50 & 7.50 & 1.50 & 5.50 & 3.50 & 7.50 & 1.50 & 5.50 & F29 & 3.50 & 7.50 & 1.50 & 5.50 & 3.50 & 7.50 & 1.50 & 5.50
\\
F30 & 2.50 & 2.50 & 5.50 & 7.50 & 2.50 & 2.50 & 5.50 & 7.50 & F30 & 2.50 & 2.50 & 5.50 & 7.50 & 2.50 & 2.50 & 5.50 & 7.50
\\
\midrule
Mean rank & 2.55 & 3.34 & 5.09 & 7.09 & 2.53 & 3.22 & 5.16 & 7.02 & Mean rank & 2.55 & 3.07 & 5.28 & 7.14 & 2.66 & 2.97 & 5.24 & 7.10
\\
\bottomrule
\end{tabular}
}
\vspace{0.5cm}
\setlength{\heavyrulewidth}{0.3pt}
\setlength{\lightrulewidth}{0.3pt}
\setlength{\cmidrulewidth}{0.3pt}
\centering
\scriptsize
\setlength{\tabcolsep}{6pt}
\caption{Function-wise eDSC ranks. Lower values indicate greater final-solution compactness.}
\label{tab:edsc_ranks}
\resizebox{\linewidth}{!}{%
\begin{tabular}{lcccccccc@{\hspace{0.75cm}}lcccccccc}
\toprule
\multicolumn{9}{c}{\textbf{Algorithm: hMPA}\hfill\textbf{$dim=10$}\hfill \qquad\qquad} & \multicolumn{9}{c}{\textbf{$dim=30$}} \\
\cmidrule(lr){1-9} \cmidrule(lr){10-18}
Functions & C(000) & C(001) & C(010) & C(011) & C(100) & C(101) & C(110) & C(111) & Functions & C(000) & C(001) & C(010) & C(011) & C(100) & C(101) & C(110) & C(111)
\\
\midrule
F1 & 2.50 & 2.50 & 5.00 & 7.00 & 2.50 & 2.50 & 6.00 & 8.00 & F1 & 2.50 & 2.50 & 5.50 & 7.50 & 2.50 & 2.50 & 5.50 & 7.50
\\
F3 & 2.50 & 2.50 & 5.50 & 5.50 & 2.50 & 2.50 & 7.50 & 7.50 & F3 & 2.50 & 2.50 & 5.00 & 8.00 & 2.50 & 2.50 & 6.00 & 7.00
\\
F4 & 2.50 & 2.50 & 5.00 & 6.00 & 2.50 & 2.50 & 7.00 & 8.00 & F4 & 2.50 & 2.50 & 5.50 & 7.50 & 2.50 & 2.50 & 5.50 & 7.50
\\
F5 & 2.50 & 2.50 & 5.50 & 7.50 & 2.50 & 2.50 & 5.50 & 7.50 & F5 & 2.50 & 2.50 & 5.50 & 7.50 & 2.50 & 2.50 & 5.50 & 7.50
\\
F6 & 2.50 & 2.50 & 6.50 & 6.50 & 2.50 & 2.50 & 6.50 & 6.50 & F6 & 2.50 & 2.50 & 6.50 & 6.50 & 2.50 & 2.50 & 6.50 & 6.50
\\
F7 & 1.50 & 1.50 & 5.50 & 7.50 & 3.50 & 3.50 & 5.50 & 7.50 & F7 & 2.50 & 2.50 & 5.50 & 7.50 & 2.50 & 2.50 & 5.50 & 7.50
\\
F8 & 1.50 & 1.50 & 5.50 & 7.50 & 3.50 & 3.50 & 5.50 & 7.50 & F8 & 2.50 & 2.50 & 5.50 & 7.50 & 2.50 & 2.50 & 5.50 & 7.50
\\
F9 & 2.00 & 2.00 & 4.00 & 8.00 & 2.00 & 5.00 & 6.00 & 7.00 & F9 & 1.00 & 6.00 & 3.00 & 8.00 & 2.00 & 5.00 & 4.00 & 7.00
\\
F10 & 1.50 & 3.50 & 5.50 & 7.50 & 1.50 & 3.50 & 5.50 & 7.50 & F10 & 1.50 & 3.00 & 6.00 & 8.00 & 1.50 & 4.00 & 5.00 & 7.00
\\
F11 & 2.50 & 2.50 & 5.50 & 7.50 & 2.50 & 2.50 & 5.50 & 7.50 & F11 & 2.50 & 2.50 & 5.50 & 7.50 & 2.50 & 2.50 & 5.50 & 7.50
\\
F12 & 1.50 & 4.00 & 5.00 & 8.00 & 1.50 & 6.00 & 3.00 & 7.00 & F12 & 2.00 & 4.00 & 5.00 & 7.00 & 2.00 & 2.00 & 6.00 & 8.00
\\
F13 & 1.50 & 3.50 & 5.50 & 7.50 & 1.50 & 3.50 & 5.50 & 7.50 & F13 & 2.50 & 2.50 & 5.50 & 7.50 & 2.50 & 2.50 & 5.50 & 7.50
\\
F14 & 1.00 & 4.00 & 5.00 & 8.00 & 2.00 & 6.00 & 3.00 & 7.00 & F14 & 1.00 & 3.50 & 6.50 & 6.50 & 2.00 & 3.50 & 6.50 & 6.50
\\
F15 & 1.50 & 5.50 & 3.50 & 7.50 & 1.50 & 5.50 & 3.50 & 7.50 & F15 & 1.50 & 3.50 & 5.50 & 7.50 & 1.50 & 3.50 & 5.50 & 7.50
\\
F16 & 2.50 & 2.50 & 5.50 & 7.50 & 2.50 & 2.50 & 5.50 & 7.50 & F16 & 4.00 & 1.00 & 5.00 & 7.00 & 3.00 & 2.00 & 6.00 & 8.00
\\
F17 & 1.00 & 4.00 & 5.00 & 7.00 & 2.00 & 3.00 & 6.00 & 8.00 & F17 & 1.50 & 3.00 & 6.00 & 7.00 & 1.50 & 4.00 & 5.00 & 8.00
\\
F18 & 1.00 & 5.00 & 4.00 & 7.00 & 2.00 & 6.00 & 3.00 & 8.00 & F18 & 1.50 & 5.50 & 3.50 & 7.50 & 1.50 & 5.50 & 3.50 & 7.50
\\
F19 & 1.00 & 5.00 & 3.00 & 6.00 & 2.00 & 7.00 & 4.00 & 8.00 & F19 & 1.00 & 4.00 & 6.00 & 7.00 & 2.00 & 3.00 & 5.00 & 8.00
\\
F20 & 1.50 & 1.50 & 6.50 & 6.50 & 3.50 & 3.50 & 6.50 & 6.50 & F20 & 2.50 & 2.50 & 6.50 & 6.50 & 2.50 & 2.50 & 6.50 & 6.50
\\
F21 & 2.50 & 2.50 & 6.50 & 6.50 & 2.50 & 2.50 & 6.50 & 6.50 & F21 & 2.50 & 2.50 & 5.50 & 7.50 & 2.50 & 2.50 & 5.50 & 7.50
\\
F22 & 2.50 & 2.50 & 6.50 & 6.50 & 2.50 & 2.50 & 6.50 & 6.50 & F22 & 2.50 & 2.50 & 6.50 & 6.50 & 2.50 & 2.50 & 6.50 & 6.50
\\
F23 & 1.50 & 7.50 & 3.50 & 5.50 & 1.50 & 7.50 & 3.50 & 5.50 & F23 & 1.50 & 5.50 & 3.50 & 7.50 & 1.50 & 5.50 & 3.50 & 7.50
\\
F24 & 2.50 & 2.50 & 6.50 & 6.50 & 2.50 & 2.50 & 6.50 & 6.50 & F24 & 2.50 & 2.50 & 5.50 & 7.50 & 2.50 & 2.50 & 5.50 & 7.50
\\
F25 & 4.50 & 4.50 & 1.50 & 7.50 & 4.50 & 4.50 & 1.50 & 7.50 & F25 & 2.50 & 2.50 & 5.50 & 7.50 & 2.50 & 2.50 & 5.50 & 7.50
\\
F26 & 2.50 & 2.50 & 5.50 & 7.50 & 2.50 & 2.50 & 5.50 & 7.50 & F26 & 2.50 & 2.50 & 6.50 & 6.50 & 2.50 & 2.50 & 6.50 & 6.50
\\
F27 & 7.50 & 3.50 & 1.50 & 5.50 & 7.50 & 3.50 & 1.50 & 5.50 & F27 & 3.50 & 7.50 & 1.50 & 5.50 & 3.50 & 7.50 & 1.50 & 5.50
\\
F28 & 4.50 & 4.50 & 7.50 & 1.50 & 4.50 & 4.50 & 7.50 & 1.50 & F28 & 2.50 & 2.50 & 5.50 & 7.50 & 2.50 & 2.50 & 5.50 & 7.50
\\
F29 & 1.50 & 7.50 & 3.50 & 5.50 & 1.50 & 7.50 & 3.50 & 5.50 & F29 & 3.50 & 7.50 & 1.50 & 5.50 & 3.50 & 7.50 & 1.50 & 5.50
\\
F30 & 1.50 & 5.50 & 3.50 & 7.50 & 1.50 & 5.50 & 3.50 & 7.50 & F30 & 1.00 & 3.00 & 6.00 & 7.00 & 2.00 & 4.00 & 5.00 & 8.00
\\
\midrule
Mean rank & 2.24 & 3.50 & 4.91 & 6.74 & 2.59 & 4.02 & 5.05 & 6.95 & Mean rank & 2.22 & 3.34 & 5.19 & 7.16 & 2.33 & 3.34 & 5.19 & 7.22
\\
\bottomrule
\end{tabular}
}
\vspace{0.4cm}

\resizebox{\linewidth}{!}{
\begin{tabular}{lcccccccc@{\hspace{0.75cm}}lcccccccc}
\toprule
\multicolumn{9}{c}{\textbf{Algorithm: hMPA}\hfill\textbf{$dim=50$}\hfill \qquad\qquad} & \multicolumn{9}{c}{\textbf{$dim=100$}} \\
\cmidrule(lr){1-9} \cmidrule(lr){10-18}
Functions & C(000) & C(001) & C(010) & C(011) & C(100) & C(101) & C(110) & C(111) & Functions & C(000) & C(001) & C(010) & C(011) & C(100) & C(101) & C(110) & C(111)
\\
\midrule
F1 & 2.50 & 2.50 & 5.50 & 7.50 & 2.50 & 2.50 & 5.50 & 7.50 & F1 & 2.50 & 2.50 & 7.50 & 5.50 & 2.50 & 2.50 & 7.50 & 5.50
\\
F3 & 2.50 & 2.50 & 5.50 & 7.50 & 2.50 & 2.50 & 5.50 & 7.50 & F3 & 2.50 & 2.50 & 5.50 & 7.50 & 2.50 & 2.50 & 5.50 & 7.50
\\
F4 & 2.50 & 2.50 & 5.50 & 7.50 & 2.50 & 2.50 & 5.50 & 7.50 & F4 & 2.50 & 2.50 & 5.50 & 7.50 & 2.50 & 2.50 & 5.50 & 7.50
\\
F5 & 2.50 & 2.50 & 5.50 & 7.50 & 2.50 & 2.50 & 5.50 & 7.50 & F5 & 2.50 & 2.50 & 7.50 & 5.50 & 2.50 & 2.50 & 7.50 & 5.50
\\
F6 & 2.50 & 2.50 & 6.50 & 6.50 & 2.50 & 2.50 & 6.50 & 6.50 & F6 & 2.50 & 2.50 & 6.50 & 6.50 & 2.50 & 2.50 & 6.50 & 6.50
\\
F7 & 2.50 & 2.50 & 5.50 & 7.50 & 2.50 & 2.50 & 5.50 & 7.50 & F7 & 2.50 & 2.50 & 5.50 & 7.50 & 2.50 & 2.50 & 5.50 & 7.50
\\
F8 & 2.50 & 2.50 & 5.50 & 7.50 & 2.50 & 2.50 & 5.50 & 7.50 & F8 & 2.50 & 2.50 & 5.50 & 7.50 & 2.50 & 2.50 & 5.50 & 7.50
\\
F9 & 1.50 & 3.50 & 5.50 & 7.50 & 1.50 & 3.50 & 5.50 & 7.50 & F9 & 1.50 & 3.50 & 7.50 & 5.50 & 1.50 & 3.50 & 7.50 & 5.50
\\
F10 & 2.00 & 2.00 & 5.00 & 8.00 & 2.00 & 4.00 & 6.00 & 7.00 & F10 & 2.50 & 2.50 & 5.50 & 7.50 & 2.50 & 2.50 & 5.50 & 7.50
\\
F11 & 2.50 & 2.50 & 5.50 & 7.50 & 2.50 & 2.50 & 5.50 & 7.50 & F11 & 2.50 & 2.50 & 5.50 & 7.50 & 2.50 & 2.50 & 5.50 & 7.50
\\
F12 & 2.50 & 2.50 & 5.50 & 7.50 & 2.50 & 2.50 & 5.50 & 7.50 & F12 & 2.50 & 2.50 & 5.50 & 7.50 & 2.50 & 2.50 & 5.50 & 7.50
\\
F13 & 2.50 & 2.50 & 5.50 & 7.50 & 2.50 & 2.50 & 5.50 & 7.50 & F13 & 2.50 & 2.50 & 5.50 & 7.50 & 2.50 & 2.50 & 5.50 & 7.50
\\
F14 & 1.00 & 5.50 & 3.50 & 7.50 & 2.00 & 5.50 & 3.50 & 7.50 & F14 & 1.50 & 3.50 & 5.50 & 7.50 & 1.50 & 3.50 & 5.50 & 7.50
\\
F15 & 3.00 & 2.00 & 5.00 & 7.00 & 4.00 & 1.00 & 6.00 & 8.00 & F15 & 2.50 & 2.50 & 5.50 & 7.50 & 2.50 & 2.50 & 5.50 & 7.50
\\
F16 & 4.00 & 1.00 & 7.00 & 6.00 & 3.00 & 2.00 & 8.00 & 5.00 & F16 & 2.50 & 2.50 & 7.50 & 5.50 & 2.50 & 2.50 & 7.50 & 5.50
\\
F17 & 2.50 & 2.50 & 7.50 & 5.50 & 2.50 & 2.50 & 7.50 & 5.50 & F17 & 2.50 & 2.50 & 7.00 & 6.00 & 2.50 & 2.50 & 8.00 & 5.00
\\
F18 & 1.50 & 5.50 & 3.50 & 7.50 & 1.50 & 5.50 & 3.50 & 7.50 & F18 & 1.50 & 3.50 & 5.50 & 7.50 & 1.50 & 3.50 & 5.50 & 7.50
\\
F19 & 2.50 & 2.50 & 5.50 & 7.50 & 2.50 & 2.50 & 5.50 & 7.50 & F19 & 2.50 & 2.50 & 7.00 & 8.00 & 2.50 & 2.50 & 5.00 & 6.00
\\
F20 & 2.50 & 2.50 & 5.50 & 7.50 & 2.50 & 2.50 & 5.50 & 7.50 & F20 & 2.50 & 2.50 & 7.50 & 5.50 & 2.50 & 2.50 & 7.50 & 5.50
\\
F21 & 2.50 & 2.50 & 5.50 & 7.50 & 2.50 & 2.50 & 5.50 & 7.50 & F21 & 3.50 & 3.50 & 5.50 & 7.50 & 1.50 & 1.50 & 5.50 & 7.50
\\
F22 & 2.50 & 2.50 & 5.50 & 7.50 & 2.50 & 2.50 & 5.50 & 7.50 & F22 & 2.50 & 2.50 & 5.50 & 7.50 & 2.50 & 2.50 & 5.50 & 7.50
\\
F23 & 1.50 & 3.50 & 5.50 & 7.50 & 1.50 & 3.50 & 5.50 & 7.50 & F23 & 1.50 & 7.50 & 3.50 & 5.50 & 1.50 & 7.50 & 3.50 & 5.50
\\
F24 & 2.50 & 2.50 & 5.00 & 7.00 & 2.50 & 2.50 & 6.00 & 8.00 & F24 & 2.50 & 2.50 & 5.50 & 7.50 & 2.50 & 2.50 & 5.50 & 7.50
\\
F25 & 2.50 & 2.50 & 5.50 & 7.50 & 2.50 & 2.50 & 5.50 & 7.50 & F25 & 2.50 & 2.50 & 7.50 & 5.50 & 2.50 & 2.50 & 7.50 & 5.50
\\
F26 & 2.50 & 2.50 & 7.50 & 5.50 & 2.50 & 2.50 & 7.50 & 5.50 & F26 & 2.50 & 2.50 & 5.50 & 7.50 & 2.50 & 2.50 & 5.50 & 7.50
\\
F27 & 3.00 & 4.00 & 1.00 & 5.00 & 7.00 & 8.00 & 2.00 & 6.00 & F27 & 5.00 & 7.00 & 2.00 & 4.00 & 6.00 & 8.00 & 1.00 & 3.00
\\
F28 & 2.50 & 2.50 & 5.50 & 7.50 & 2.50 & 2.50 & 5.50 & 7.50 & F28 & 2.50 & 2.50 & 7.50 & 5.50 & 2.50 & 2.50 & 7.50 & 5.50
\\
F29 & 5.50 & 7.50 & 1.50 & 3.50 & 5.50 & 7.50 & 1.50 & 3.50 & F29 & 5.50 & 7.50 & 1.50 & 3.50 & 5.50 & 7.50 & 1.50 & 3.50
\\
F30 & 2.50 & 2.50 & 5.50 & 7.50 & 2.50 & 2.50 & 5.50 & 7.50 & F30 & 4.00 & 3.00 & 6.00 & 8.00 & 2.00 & 1.00 & 5.00 & 7.00
\\
\midrule
Mean rank & 2.52 & 2.91 & 5.24 & 7.03 & 2.69 & 3.12 & 5.41 & 7.07 & Mean rank & 2.64 & 3.16 & 5.81 & 6.64 & 2.53 & 3.05 & 5.71 & 6.47
\\
\bottomrule
\end{tabular}
}
\end{table}

\clearpage
\begin{table}[]
\setlength{\heavyrulewidth}{0.3pt}
\setlength{\lightrulewidth}{0.3pt}
\setlength{\cmidrulewidth}{0.3pt}
\centering
\vspace{-0.5cm}
\scriptsize
\setlength{\tabcolsep}{6pt}
\noindent\textit{Table~\ref{tab:edsc_ranks} (continued).}\par\vspace{2pt}
\resizebox{\linewidth}{!}{
\begin{tabular}{lcccccccc@{\hspace{0.75cm}}lcccccccc}
\toprule
\multicolumn{9}{c}{\textbf{Algorithm: MPA}\hfill\textbf{$dim=10$}\hfill \qquad\qquad} & \multicolumn{9}{c}{\textbf{$dim=30$}} \\
\cmidrule(lr){1-9} \cmidrule(lr){10-18}
Functions & C(000) & C(001) & C(010) & C(011) & C(100) & C(101) & C(110) & C(111) & Functions & C(000) & C(001) & C(010) & C(011) & C(100) & C(101) & C(110) & C(111)
\\
\midrule
F1 & 2.50 & 2.50 & 6.50 & 6.50 & 2.50 & 2.50 & 6.50 & 6.50 & F1 & 2.50 & 2.50 & 5.50 & 7.50 & 2.50 & 2.50 & 5.50 & 7.50
\\
F3 & 2.50 & 2.50 & 6.50 & 6.50 & 2.50 & 2.50 & 6.50 & 6.50 & F3 & 2.50 & 2.50 & 5.50 & 7.50 & 2.50 & 2.50 & 5.50 & 7.50
\\
F4 & 2.50 & 2.50 & 6.50 & 6.50 & 2.50 & 2.50 & 6.50 & 6.50 & F4 & 2.50 & 2.50 & 5.50 & 7.50 & 2.50 & 2.50 & 5.50 & 7.50
\\
F5 & 1.50 & 4.00 & 6.00 & 8.00 & 1.50 & 3.00 & 5.00 & 7.00 & F5 & 2.50 & 2.50 & 5.50 & 7.50 & 2.50 & 2.50 & 5.50 & 7.50
\\
F6 & 2.50 & 2.50 & 6.00 & 7.00 & 2.50 & 2.50 & 8.00 & 5.00 & F6 & 2.50 & 2.50 & 6.50 & 6.50 & 2.50 & 2.50 & 6.50 & 6.50
\\
F7 & 1.50 & 3.00 & 5.00 & 8.00 & 1.50 & 4.00 & 6.00 & 7.00 & F7 & 2.50 & 2.50 & 5.50 & 7.50 & 2.50 & 2.50 & 5.50 & 7.50
\\
F8 & 1.50 & 4.00 & 5.00 & 8.00 & 1.50 & 3.00 & 6.00 & 7.00 & F8 & 2.50 & 2.50 & 5.50 & 7.50 & 2.50 & 2.50 & 5.50 & 7.50
\\
F9 & 1.50 & 6.00 & 3.00 & 7.00 & 1.50 & 5.00 & 4.00 & 8.00 & F9 & 1.50 & 5.50 & 3.50 & 7.50 & 1.50 & 5.50 & 3.50 & 7.50
\\
F10 & 1.50 & 3.50 & 5.50 & 7.50 & 1.50 & 3.50 & 5.50 & 7.50 & F10 & 2.50 & 2.50 & 5.50 & 7.50 & 2.50 & 2.50 & 5.50 & 7.50
\\
F11 & 2.50 & 2.50 & 5.50 & 7.50 & 2.50 & 2.50 & 5.50 & 7.50 & F11 & 2.50 & 2.50 & 5.50 & 7.50 & 2.50 & 2.50 & 5.50 & 7.50
\\
F12 & 1.50 & 4.00 & 3.00 & 8.00 & 1.50 & 5.00 & 6.00 & 7.00 & F12 & 2.00 & 2.00 & 5.00 & 8.00 & 2.00 & 4.00 & 6.00 & 7.00
\\
F13 & 1.50 & 3.50 & 5.50 & 7.50 & 1.50 & 3.50 & 5.50 & 7.50 & F13 & 2.50 & 2.50 & 5.50 & 7.50 & 2.50 & 2.50 & 5.50 & 7.50
\\
F14 & 3.00 & 6.00 & 2.00 & 7.00 & 1.00 & 5.00 & 4.00 & 8.00 & F14 & 2.00 & 5.00 & 3.00 & 6.00 & 1.00 & 7.00 & 4.00 & 8.00
\\
F15 & 1.00 & 5.00 & 3.00 & 7.00 & 2.00 & 6.00 & 4.00 & 8.00 & F15 & 1.00 & 3.00 & 6.00 & 8.00 & 2.00 & 4.00 & 5.00 & 7.00
\\
F16 & 2.00 & 3.00 & 8.00 & 5.00 & 1.00 & 4.00 & 7.00 & 6.00 & F16 & 2.50 & 2.50 & 5.50 & 7.50 & 2.50 & 2.50 & 5.50 & 7.50
\\
F17 & 1.00 & 3.00 & 6.00 & 8.00 & 2.00 & 4.00 & 5.00 & 7.00 & F17 & 1.50 & 3.00 & 6.00 & 7.00 & 1.50 & 4.00 & 5.00 & 8.00
\\
F18 & 2.00 & 6.00 & 4.00 & 7.00 & 1.00 & 5.00 & 3.00 & 8.00 & F18 & 2.00 & 5.00 & 3.00 & 7.00 & 1.00 & 6.00 & 4.00 & 8.00
\\
F19 & 2.00 & 5.00 & 3.00 & 6.00 & 1.00 & 8.00 & 4.00 & 7.00 & F19 & 1.50 & 3.00 & 6.00 & 8.00 & 1.50 & 4.00 & 5.00 & 7.00
\\
F20 & 3.00 & 4.00 & 5.00 & 6.00 & 1.00 & 2.00 & 7.00 & 8.00 & F20 & 2.50 & 2.50 & 5.50 & 7.50 & 2.50 & 2.50 & 5.50 & 7.50
\\
F21 & 2.50 & 2.50 & 6.50 & 6.50 & 2.50 & 2.50 & 6.50 & 6.50 & F21 & 2.50 & 2.50 & 5.50 & 7.50 & 2.50 & 2.50 & 5.50 & 7.50
\\
F22 & 2.50 & 2.50 & 6.50 & 6.50 & 2.50 & 2.50 & 6.50 & 6.50 & F22 & 2.50 & 2.50 & 6.50 & 6.50 & 2.50 & 2.50 & 6.50 & 6.50
\\
F23 & 1.50 & 7.50 & 3.50 & 5.50 & 1.50 & 7.50 & 3.50 & 5.50 & F23 & 1.50 & 3.50 & 5.50 & 7.50 & 1.50 & 3.50 & 5.50 & 7.50
\\
F24 & 2.50 & 2.50 & 6.50 & 6.50 & 2.50 & 2.50 & 6.50 & 6.50 & F24 & 2.50 & 2.50 & 5.50 & 7.50 & 2.50 & 2.50 & 5.50 & 7.50
\\
F25 & 4.50 & 4.50 & 1.50 & 7.50 & 4.50 & 4.50 & 1.50 & 7.50 & F25 & 2.50 & 2.50 & 5.50 & 7.50 & 2.50 & 2.50 & 5.50 & 7.50
\\
F26 & 2.50 & 2.50 & 5.50 & 7.50 & 2.50 & 2.50 & 5.50 & 7.50 & F26 & 2.50 & 2.50 & 6.50 & 6.50 & 2.50 & 2.50 & 6.50 & 6.50
\\
F27 & 5.50 & 7.50 & 1.50 & 3.50 & 5.50 & 7.50 & 1.50 & 3.50 & F27 & 3.50 & 7.50 & 1.50 & 5.50 & 3.50 & 7.50 & 1.50 & 5.50
\\
F28 & 2.50 & 2.50 & 7.50 & 5.50 & 2.50 & 2.50 & 7.50 & 5.50 & F28 & 2.50 & 2.50 & 5.50 & 7.50 & 2.50 & 2.50 & 5.50 & 7.50
\\
F29 & 1.50 & 7.50 & 3.50 & 5.50 & 1.50 & 7.50 & 3.50 & 5.50 & F29 & 3.50 & 7.50 & 1.50 & 5.50 & 3.50 & 7.50 & 1.50 & 5.50
\\
F30 & 1.50 & 5.50 & 3.50 & 7.50 & 1.50 & 5.50 & 3.50 & 7.50 & F30 & 1.50 & 3.50 & 5.50 & 7.50 & 1.50 & 3.50 & 5.50 & 7.50
\\
\midrule
Mean rank & 2.21 & 4.05 & 4.86 & 6.76 & 2.03 & 4.09 & 5.21 & 6.79 & Mean rank & 2.29 & 3.22 & 5.12 & 7.22 & 2.26 & 3.50 & 5.12 & 7.26
\\
\bottomrule
\end{tabular}
}

\vspace{0.4cm}
\resizebox{\linewidth}{!}{
\begin{tabular}{lcccccccc@{\hspace{0.75cm}}lcccccccc}
\toprule
\multicolumn{9}{c}{\textbf{Algorithm: MPA}\hfill\textbf{$dim=50$}\hfill \qquad\qquad} & \multicolumn{9}{c}{\textbf{$dim=100$}} \\
\cmidrule(lr){1-9} \cmidrule(lr){10-18}
Functions & C(000) & C(001) & C(010) & C(011) & C(100) & C(101) & C(110) & C(111) & Functions & C(000) & C(001) & C(010) & C(011) & C(100) & C(101) & C(110) & C(111)
\\
\midrule
F1 & 2.50 & 2.50 & 5.50 & 7.50 & 2.50 & 2.50 & 5.50 & 7.50 & F1 & 2.50 & 2.50 & 5.50 & 7.50 & 2.50 & 2.50 & 5.50 & 7.50
\\
F3 & 2.50 & 2.50 & 5.50 & 7.50 & 2.50 & 2.50 & 5.50 & 7.50 & F3 & 2.50 & 2.50 & 5.50 & 7.50 & 2.50 & 2.50 & 5.50 & 7.50
\\
F4 & 2.50 & 2.50 & 5.50 & 7.50 & 2.50 & 2.50 & 5.50 & 7.50 & F4 & 2.50 & 2.50 & 7.50 & 5.50 & 2.50 & 2.50 & 7.50 & 5.50
\\
F5 & 2.50 & 2.50 & 5.50 & 7.50 & 2.50 & 2.50 & 5.50 & 7.50 & F5 & 2.50 & 2.50 & 7.50 & 5.50 & 2.50 & 2.50 & 7.50 & 5.50
\\
F6 & 2.50 & 2.50 & 6.50 & 6.50 & 2.50 & 2.50 & 6.50 & 6.50 & F6 & 2.50 & 2.50 & 6.50 & 6.50 & 2.50 & 2.50 & 6.50 & 6.50
\\
F7 & 2.50 & 2.50 & 5.50 & 7.50 & 2.50 & 2.50 & 5.50 & 7.50 & F7 & 2.50 & 2.50 & 5.50 & 7.50 & 2.50 & 2.50 & 5.50 & 7.50
\\
F8 & 2.50 & 2.50 & 5.50 & 7.50 & 2.50 & 2.50 & 5.50 & 7.50 & F8 & 2.50 & 2.50 & 5.50 & 7.50 & 2.50 & 2.50 & 5.50 & 7.50
\\
F9 & 1.50 & 3.50 & 5.50 & 7.50 & 1.50 & 3.50 & 5.50 & 7.50 & F9 & 1.50 & 3.50 & 7.50 & 5.50 & 1.50 & 3.50 & 7.50 & 5.50
\\
F10 & 2.50 & 2.50 & 7.50 & 5.50 & 2.50 & 2.50 & 7.50 & 5.50 & F10 & 2.50 & 2.50 & 7.50 & 5.50 & 2.50 & 2.50 & 7.50 & 5.50
\\
F11 & 2.50 & 2.50 & 7.50 & 5.50 & 2.50 & 2.50 & 7.50 & 5.50 & F11 & 2.50 & 2.50 & 7.50 & 5.50 & 2.50 & 2.50 & 7.50 & 5.50
\\
F12 & 2.50 & 2.50 & 5.50 & 7.50 & 2.50 & 2.50 & 5.50 & 7.50 & F12 & 2.50 & 2.50 & 5.50 & 7.50 & 2.50 & 2.50 & 5.50 & 7.50
\\
F13 & 2.50 & 2.50 & 5.50 & 7.50 & 2.50 & 2.50 & 5.50 & 7.50 & F13 & 2.50 & 2.50 & 5.50 & 7.50 & 2.50 & 2.50 & 5.50 & 7.50
\\
F14 & 1.50 & 5.50 & 3.50 & 7.50 & 1.50 & 5.50 & 3.50 & 7.50 & F14 & 1.50 & 5.50 & 3.50 & 7.50 & 1.50 & 5.50 & 3.50 & 7.50
\\
F15 & 2.50 & 2.50 & 5.50 & 7.50 & 2.50 & 2.50 & 5.50 & 7.50 & F15 & 2.50 & 2.50 & 5.50 & 7.50 & 2.50 & 2.50 & 5.50 & 7.50
\\
F16 & 4.00 & 1.00 & 8.00 & 5.00 & 3.00 & 2.00 & 7.00 & 6.00 & F16 & 2.00 & 3.00 & 7.00 & 5.00 & 4.00 & 1.00 & 8.00 & 6.00
\\
F17 & 2.00 & 2.00 & 8.00 & 6.00 & 2.00 & 4.00 & 7.00 & 5.00 & F17 & 2.50 & 2.50 & 7.50 & 5.50 & 2.50 & 2.50 & 7.50 & 5.50
\\
F18 & 1.50 & 5.50 & 3.50 & 7.50 & 1.50 & 5.50 & 3.50 & 7.50 & F18 & 1.50 & 5.50 & 3.50 & 7.50 & 1.50 & 5.50 & 3.50 & 7.50
\\
F19 & 2.50 & 2.50 & 5.50 & 7.50 & 2.50 & 2.50 & 5.50 & 7.50 & F19 & 2.50 & 2.50 & 6.00 & 8.00 & 2.50 & 2.50 & 7.00 & 5.00
\\
F20 & 2.50 & 2.50 & 5.50 & 7.50 & 2.50 & 2.50 & 5.50 & 7.50 & F20 & 2.50 & 2.50 & 7.50 & 5.50 & 2.50 & 2.50 & 7.50 & 5.50
\\
F21 & 2.50 & 2.50 & 5.50 & 7.50 & 2.50 & 2.50 & 5.50 & 7.50 & F21 & 1.50 & 4.00 & 5.00 & 8.00 & 1.50 & 3.00 & 6.00 & 7.00
\\
F22 & 2.50 & 2.50 & 5.50 & 7.50 & 2.50 & 2.50 & 5.50 & 7.50 & F22 & 2.50 & 2.50 & 5.50 & 7.50 & 2.50 & 2.50 & 5.50 & 7.50
\\
F23 & 1.50 & 3.50 & 5.50 & 7.50 & 1.50 & 3.50 & 5.50 & 7.50 & F23 & 1.50 & 4.00 & 5.00 & 8.00 & 1.50 & 3.00 & 6.00 & 7.00
\\
F24 & 2.50 & 2.50 & 5.50 & 7.50 & 2.50 & 2.50 & 5.50 & 7.50 & F24 & 2.50 & 2.50 & 5.50 & 7.50 & 2.50 & 2.50 & 5.50 & 7.50
\\
F25 & 2.50 & 2.50 & 5.50 & 7.50 & 2.50 & 2.50 & 5.50 & 7.50 & F25 & 2.50 & 2.50 & 7.50 & 5.50 & 2.50 & 2.50 & 7.50 & 5.50
\\
F26 & 2.50 & 2.50 & 7.50 & 5.50 & 2.50 & 2.50 & 7.50 & 5.50 & F26 & 2.50 & 2.50 & 5.50 & 7.50 & 2.50 & 2.50 & 5.50 & 7.50
\\
F27 & 7.00 & 5.00 & 1.00 & 4.00 & 6.00 & 8.00 & 2.00 & 3.00 & F27 & 5.00 & 7.00 & 1.00 & 3.00 & 6.00 & 8.00 & 2.00 & 4.00
\\
F28 & 2.50 & 2.50 & 5.50 & 7.50 & 2.50 & 2.50 & 5.50 & 7.50 & F28 & 4.00 & 1.00 & 8.00 & 5.00 & 3.00 & 2.00 & 7.00 & 6.00
\\
F29 & 5.50 & 7.50 & 1.50 & 3.50 & 5.50 & 7.50 & 1.50 & 3.50 & F29 & 5.50 & 7.50 & 3.50 & 1.50 & 5.50 & 7.50 & 3.50 & 1.50
\\
F30 & 2.00 & 4.00 & 6.00 & 7.00 & 3.00 & 1.00 & 5.00 & 8.00 & F30 & 4.00 & 2.00 & 6.00 & 7.00 & 3.00 & 1.00 & 5.00 & 8.00
\\
\midrule
Mean rank & 2.64 & 3.02 & 5.50 & 6.84 & 2.60 & 3.12 & 5.43 & 6.84 & Mean rank & 2.60 & 3.12 & 5.86 & 6.41 & 2.64 & 3.02 & 5.97 & 6.38
\\
\bottomrule
\end{tabular}
}
\end{table}
Both schemes allocate the four leading mean-rank positions to $C(000)$, $C(001)$, $C(100)$ and $C(101)$. This remains consistent for both algorithms across all tested dimensions. $C(011)$ and $C(111)$ have the two highest mean ranks. In DSC, $C(010)$ and $C(110)$ share the lowest rank of $1.5$ on $F27$ at $dim\in\{30, \ 50, \ 100\}$. The same tie occurs on $F29$ at $dim\in\{50, \ 100\}$. For both algorithms, the eDSC mean-rank range is smallest at $dim=100$. The ranges computed from means are $4.10$ for hMPA and $3.81$ for MPA.


\subsection{Exhaustive 3-configuration comparisons}
\label{sec7.2}

To assess whether the global differences persist in smaller comparison
sets, all $56$ three-configuration subsets were analyzed.
Tables~\ref{tab:triples_dsc} and~\ref{tab:triples_edsc} report the
corresponding Friedman-test $p$-values for DSC and eDSC(e.g., Eftimov and Koro{\v{s}}ec in \cite{TE2019}).

\clearpage

\begin{table}[h]
\vspace{-1.8cm}
\setlength{\heavyrulewidth}{0.3pt}
\setlength{\lightrulewidth}{0.3pt}
\setlength{\cmidrulewidth}{0.3pt}
\centering
\tiny
\renewcommand{\arraystretch}{0.86}
\setlength{\tabcolsep}{4pt}
\caption{Friedman-test $p$-values for the DSC 3-configuration comparisons.}
\label{tab:triples_dsc}
\resizebox{\linewidth}{!}{%
\begin{tabular}{lcccc}
\toprule
Triple & $dim=10$ & $dim=30$ & $dim=50$ & $dim=100$  \\
\midrule
\textbf{Algorithm: hMPA}\\
C(000) vs C(001) vs C(010) & 3.5273E-10 & 2.4439E-07 & 1.3745E-07 & 4.5830E-07 \\
C(000) vs C(001) vs C(011) & 3.1145E-12 & 2.1470E-09 & 1.8358E-09 & 7.0802E-09 \\
C(000) vs C(001) vs C(100) & 6.4572E-03 & 2.8856E-02 & 2.8232E-01 & 1.1518E-01 \\
C(000) vs C(001) vs C(101) & 8.7802E-05 & 1.3832E-03 & 1.3173E-01 & 7.4274E-02 \\
C(000) vs C(001) vs C(110) & 5.7865E-11 & 2.4439E-07 & 1.3745E-07 & 4.5830E-07 \\
C(000) vs C(001) vs C(111) & 3.1145E-12 & 2.1470E-09 & 1.8358E-09 & 7.0802E-09 \\
C(000) vs C(010) vs C(011) & 9.2055E-12 & 1.1412E-10 & 4.1684E-11 & 5.2626E-10 \\
C(000) vs C(010) vs C(100) & 8.7092E-13 & 2.0349E-11 & 1.0778E-09 & 2.2374E-08 \\
C(000) vs C(010) vs C(101) & 5.7146E-10 & 2.7426E-08 & 1.3860E-07 & 6.1665E-07 \\
C(000) vs C(010) vs C(110) & 7.6332E-13 & 1.5917E-11 & 5.5161E-10 & 1.7234E-08 \\
C(000) vs C(010) vs C(111) & 7.9041E-12 & 1.9640E-10 & 3.2726E-11 & 5.2626E-10 \\
C(000) vs C(011) vs C(100) & 8.7092E-13 & 2.0349E-11 & 3.4956E-11 & 9.4464E-10 \\
C(000) vs C(011) vs C(101) & 5.4757E-12 & 1.1446E-09 & 1.9770E-09 & 1.1064E-08 \\
C(000) vs C(011) vs C(110) & 1.4143E-11 & 1.1412E-10 & 7.9920E-11 & 5.2626E-10 \\
C(000) vs C(011) vs C(111) & 6.6138E-13 & 1.2099E-11 & 1.5917E-11 & 5.5161E-10 \\
C(000) vs C(100) vs C(101) & 8.8427E-05 & 5.5452E-05 & 1.3904E-01 & 6.2177E-02 \\
C(000) vs C(100) vs C(110) & 8.7092E-13 & 2.0349E-11 & 1.0778E-09 & 2.2374E-08 \\
C(000) vs C(100) vs C(111) & 8.7092E-13 & 2.0349E-11 & 3.4956E-11 & 9.4464E-10 \\
C(000) vs C(101) vs C(110) & 5.7146E-10 & 2.7426E-08 & 1.3860E-07 & 6.1665E-07 \\
C(000) vs C(101) vs C(111) & 5.4757E-12 & 1.1446E-09 & 1.9770E-09 & 1.1064E-08 \\
C(000) vs C(110) vs C(111) & 7.9041E-12 & 1.3177E-10 & 3.2726E-11 & 5.2626E-10 \\
C(001) vs C(010) vs C(011) & 5.6599E-10 & 4.2067E-07 & 1.5093E-08 & 1.1040E-08 \\
C(001) vs C(010) vs C(100) & 8.9876E-10 & 2.2603E-06 & 1.0623E-06 & 8.2300E-07 \\
C(001) vs C(010) vs C(101) & 3.7447E-07 & 5.0045E-04 & 5.8771E-06 & 5.8771E-06 \\
C(001) vs C(010) vs C(110) & 8.2447E-08 & 4.6114E-04 & 4.4697E-06 & 4.9631E-06 \\
C(001) vs C(010) vs C(111) & 3.5735E-10 & 7.2264E-07 & 1.1043E-08 & 1.1040E-08 \\
C(001) vs C(011) vs C(100) & 7.5169E-12 & 4.4073E-09 & 3.8019E-09 & 1.4170E-08 \\
C(001) vs C(011) vs C(101) & 1.6839E-12 & 1.9884E-08 & 2.1733E-08 & 2.1733E-08 \\
C(001) vs C(011) vs C(110) & 1.6071E-10 & 4.2067E-07 & 2.7950E-08 & 1.1040E-08 \\
C(001) vs C(011) vs C(111) & 6.6138E-13 & 1.1964E-08 & 1.4553E-08 & 1.4553E-08 \\
C(001) vs C(100) vs C(101) & 2.0294E-03 & 7.5456E-03 & 4.5943E-01 & 1.2246E-01 \\
C(001) vs C(100) vs C(110) & 1.4969E-10 & 2.2603E-06 & 1.0623E-06 & 8.2300E-07 \\
C(001) vs C(100) vs C(111) & 7.5169E-12 & 4.4073E-09 & 3.8019E-09 & 1.4170E-08 \\
C(001) vs C(101) vs C(110) & 6.9862E-08 & 5.0045E-04 & 5.8771E-06 & 5.8771E-06 \\
C(001) vs C(101) vs C(111) & 1.6839E-12 & 1.9884E-08 & 2.1733E-08 & 2.1733E-08 \\
C(001) vs C(110) vs C(111) & 8.0482E-11 & 5.6139E-07 & 1.1043E-08 & 1.1040E-08 \\
C(010) vs C(011) vs C(100) & 9.2055E-12 & 1.1412E-10 & 1.0448E-10 & 5.2626E-10 \\
C(010) vs C(011) vs C(101) & 1.0289E-08 & 2.2642E-07 & 2.8357E-08 & 1.1040E-08 \\
C(010) vs C(011) vs C(110) & 2.8874E-04 & 1.4279E-07 & 1.5959E-10 & 5.4768E-11 \\
C(010) vs C(011) vs C(111) & 3.9129E-05 & 9.7215E-08 & 2.5240E-11 & 4.2967E-11 \\
C(010) vs C(100) vs C(101) & 7.9880E-10 & 3.8216E-08 & 5.9161E-07 & 4.5830E-07 \\
C(010) vs C(100) vs C(110) & 7.6332E-13 & 1.5917E-11 & 1.4553E-08 & 1.7234E-08 \\
C(010) vs C(100) vs C(111) & 7.9041E-12 & 1.9640E-10 & 7.9920E-11 & 5.2626E-10 \\
C(010) vs C(101) vs C(110) & 2.5946E-06 & 5.2033E-05 & 5.2033E-05 & 4.9631E-06 \\
C(010) vs C(101) vs C(111) & 1.0684E-09 & 3.8427E-07 & 2.0267E-08 & 1.1040E-08 \\
C(010) vs C(110) vs C(111) & 3.3044E-06 & 6.2487E-07 & 1.9045E-11 & 5.4768E-11 \\
C(011) vs C(100) vs C(101) & 5.1615E-12 & 1.3275E-09 & 1.6661E-09 & 7.0802E-09 \\
C(011) vs C(100) vs C(110) & 1.4143E-11 & 1.1412E-10 & 2.0227E-10 & 5.2626E-10 \\
C(011) vs C(100) vs C(111) & 6.6138E-13 & 1.2099E-11 & 1.5917E-11 & 5.5161E-10 \\
C(011) vs C(101) vs C(110) & 8.0285E-10 & 2.2642E-07 & 2.7950E-08 & 1.1040E-08 \\
C(011) vs C(101) vs C(111) & 1.7951E-12 & 1.1964E-08 & 1.4553E-08 & 1.4553E-08 \\
C(011) vs C(110) vs C(111) & 1.0977E-05 & 2.4982E-07 & 4.8618E-10 & 4.2967E-11 \\
C(100) vs C(101) vs C(110) & 7.9880E-10 & 3.8216E-08 & 5.9161E-07 & 4.5830E-07 \\
C(100) vs C(101) vs C(111) & 5.1615E-12 & 1.3275E-09 & 1.6661E-09 & 7.0802E-09 \\
C(100) vs C(110) vs C(111) & 7.9041E-12 & 1.3177E-10 & 7.9920E-11 & 5.2626E-10 \\
C(101) vs C(110) vs C(111) & 1.0684E-09 & 2.9517E-07 & 1.1043E-08 & 1.1040E-08 \\
\\
\bottomrule
\end{tabular}
}
\end{table}
\newpage

\begin{table}[h]
\vspace{-1.8cm}
\setlength{\heavyrulewidth}{0.3pt}
\setlength{\lightrulewidth}{0.3pt}
\setlength{\cmidrulewidth}{0.3pt}
\centering
\tiny
\renewcommand{\arraystretch}{0.86}
\setlength{\tabcolsep}{4pt}
\noindent\textit{Table~\ref{tab:triples_dsc} (continued).}\par\vspace{4pt}
\resizebox{\linewidth}{!}{%
\begin{tabular}{lcccc}
\toprule
Triple & $dim=10$ & $dim=30$ & $dim=50$ & $dim=100$  \\
\midrule
\textbf{Algorithm: MPA}\\
C(000) vs C(001) vs C(010) & 1.8324E-09 & 4.9746E-08 & 4.2546E-07 & 3.9710E-08
\\
C(000) vs C(001) vs C(011) & 2.2670E-11 & 1.4275E-09 & 2.0825E-09 & 4.5991E-10
\\
C(000) vs C(001) vs C(100) & 3.8845E-04 & 4.1324E-03 & 2.5149E-02 & 1.3963E-01
\\
C(000) vs C(001) vs C(101) & 4.9477E-03 & 2.7103E-03 & 6.6716E-02 & 2.6598E-01
\\
C(000) vs C(001) vs C(110) & 3.0522E-10 & 1.6525E-08 & 4.2546E-07 & 3.9710E-08
\\
C(000) vs C(001) vs C(111) & 2.2670E-11 & 1.4275E-09 & 2.0825E-09 & 4.5991E-10
\\
C(000) vs C(010) vs C(011) & 5.6734E-12 & 1.1412E-10 & 3.2726E-11 & 6.3634E-11
\\
C(000) vs C(010) vs C(100) & 6.7600E-12 & 3.6527E-11 & 8.8588E-10 & 8.8588E-10
\\
C(000) vs C(010) vs C(101) & 6.7258E-09 & 3.0344E-08 & 8.2317E-07 & 6.6519E-08
\\
C(000) vs C(010) vs C(110) & 1.2168E-12 & 2.0349E-11 & 4.3672E-10 & 4.3672E-10
\\
C(000) vs C(010) vs C(111) & 8.4622E-12 & 1.1412E-10 & 7.9920E-11 & 6.3634E-11
\\
C(000) vs C(011) vs C(100) & 6.7600E-12 & 3.6527E-11 & 2.7658E-11 & 2.7658E-11
\\
C(000) vs C(011) vs C(101) & 1.4094E-10 & 3.0689E-09 & 4.1381E-09 & 7.5826E-10
\\
C(000) vs C(011) vs C(110) & 3.6403E-11 & 2.2358E-10 & 4.1684E-11 & 4.9985E-11
\\
C(000) vs C(011) vs C(111) & 4.6662E-13 & 1.2099E-11 & 2.0349E-11 & 1.5917E-11
\\
C(000) vs C(100) vs C(101) & 6.3584E-03 & 1.5305E-03 & 1.7377E-01 & 7.9787E-01
\\
C(000) vs C(100) vs C(110) & 6.7600E-12 & 3.6527E-11 & 8.8588E-10 & 8.8588E-10
\\
C(000) vs C(100) vs C(111) & 6.7600E-12 & 3.6527E-11 & 2.7658E-11 & 2.7658E-11
\\
C(000) vs C(101) vs C(110) & 4.4580E-09 & 3.0344E-08 & 8.2317E-07 & 6.6519E-08
\\
C(000) vs C(101) vs C(111) & 1.4094E-10 & 3.0689E-09 & 4.1381E-09 & 7.5826E-10
\\
C(000) vs C(110) vs C(111) & 1.4966E-11 & 1.9640E-10 & 7.9920E-11 & 6.3634E-11
\\
C(001) vs C(010) vs C(011) & 3.4656E-09 & 2.2642E-07 & 2.0267E-08 & 2.0072E-09
\\
C(001) vs C(010) vs C(100) & 7.9488E-10 & 1.0710E-07 & 5.3553E-07 & 5.4126E-08
\\
C(001) vs C(010) vs C(101) & 6.5848E-07 & 6.9692E-05 & 6.9692E-05 & 4.0836E-07
\\
C(001) vs C(010) vs C(110) & 1.6133E-06 & 5.6204E-05 & 4.6993E-05 & 2.4873E-07
\\
C(001) vs C(010) vs C(111) & 2.3931E-08 & 2.2642E-07 & 5.3158E-08 & 2.0072E-09
\\
C(001) vs C(011) vs C(100) & 2.8179E-11 & 3.2182E-09 & 3.2004E-09 & 6.4295E-10
\\
C(001) vs C(011) vs C(101) & 5.6448E-11 & 2.5527E-08 & 2.5527E-08 & 8.8588E-10
\\
C(001) vs C(011) vs C(110) & 2.5607E-08 & 2.3876E-07 & 2.8357E-08 & 1.5030E-09
\\
C(001) vs C(011) vs C(111) & 1.4730E-10 & 1.1964E-08 & 1.7234E-08 & 5.5161E-10
\\
C(001) vs C(100) vs C(101) & 6.8530E-05 & 5.8942E-04 & 4.9787E-02 & 2.4980E-01
\\
C(001) vs C(100) vs C(110) & 7.9488E-10 & 3.6041E-08 & 5.3553E-07 & 5.4126E-08
\\
C(001) vs C(100) vs C(111) & 9.2504E-11 & 3.2182E-09 & 3.2004E-09 & 6.4295E-10
\\
C(001) vs C(101) vs C(110) & 4.0707E-07 & 6.4913E-06 & 6.9692E-05 & 4.0836E-07
\\
C(001) vs C(101) vs C(111) & 2.8873E-10 & 2.5527E-08 & 2.5527E-08 & 8.8588E-10
\\
C(001) vs C(110) vs C(111) & 5.8957E-08 & 1.9049E-07 & 5.3158E-08 & 2.0072E-09
\\
C(010) vs C(011) vs C(100) & 5.6734E-12 & 1.1412E-10 & 3.2726E-11 & 6.3634E-11
\\
C(010) vs C(011) vs C(101) & 2.2534E-08 & 1.1354E-07 & 2.0267E-08 & 2.0072E-09
\\
C(010) vs C(011) vs C(110) & 7.9208E-04 & 2.9280E-06 & 1.9045E-11 & 1.1593E-10
\\
C(010) vs C(011) vs C(111) & 4.7305E-05 & 2.0823E-07 & 2.5240E-11 & 3.1261E-10
\\
C(010) vs C(100) vs C(101) & 2.1117E-10 & 1.1891E-08 & 4.4509E-07 & 3.2678E-08
\\
C(010) vs C(100) vs C(110) & 1.2168E-12 & 2.0349E-11 & 4.3672E-10 & 4.3672E-10
\\
C(010) vs C(100) vs C(111) & 8.4622E-12 & 1.1412E-10 & 7.9920E-11 & 6.3634E-11
\\
C(010) vs C(101) vs C(110) & 1.0110E-06 & 4.9631E-06 & 4.6993E-05 & 2.4873E-07
\\
C(010) vs C(101) vs C(111) & 4.2044E-09 & 1.1354E-07 & 5.3158E-08 & 2.0072E-09
\\
C(010) vs C(110) vs C(111) & 1.2115E-02 & 2.4862E-06 & 6.4614E-10 & 2.3917E-10
\\
C(011) vs C(100) vs C(101) & 1.3888E-11 & 1.2440E-09 & 1.9770E-09 & 3.0729E-10
\\
C(011) vs C(100) vs C(110) & 3.6403E-11 & 2.2358E-10 & 4.1684E-11 & 4.9985E-11
\\
C(011) vs C(100) vs C(111) & 4.6662E-13 & 1.2099E-11 & 2.0349E-11 & 1.5917E-11
\\
C(011) vs C(101) vs C(110) & 1.7618E-08 & 2.3876E-07 & 2.8357E-08 & 1.5030E-09
\\
C(011) vs C(101) vs C(111) & 1.4730E-10 & 1.1964E-08 & 1.7234E-08 & 5.5161E-10
\\
C(011) vs C(110) vs C(111) & 2.7739E-03 & 1.8148E-06 & 3.2368E-11 & 1.4730E-10
\\
C(100) vs C(101) vs C(110) & 1.3937E-10 & 1.1891E-08 & 4.4509E-07 & 3.2678E-08
\\
C(100) vs C(101) vs C(111) & 1.3888E-11 & 1.2440E-09 & 1.9770E-09 & 3.0729E-10
\\
C(100) vs C(110) vs C(111) & 1.4966E-11 & 1.9640E-10 & 7.9920E-11 & 6.3634E-11
\\
C(101) vs C(110) vs C(111) & 4.0987E-08 & 1.9049E-07 & 5.3158E-08 & 2.0072E-09
\\
\bottomrule
\end{tabular}
}
\end{table}
\newpage

\clearpage

\begin{table}[h!]
\setlength{\heavyrulewidth}{0.3pt}
\setlength{\lightrulewidth}{0.3pt}
\setlength{\cmidrulewidth}{0.3pt}
\centering
\tiny
\renewcommand{\arraystretch}{0.86}
\setlength{\tabcolsep}{4pt}
\caption{Friedman-test $p$-values for the eDSC 3-configuration comparisons.}
\label{tab:triples_edsc}
\resizebox{\linewidth}{!}{%
\begin{tabular}{lcccc}
\toprule
Triple & $dim=10$ & $dim=30$ & $dim=50$ & $dim=100$  \\
\midrule
\textbf{Algorithm: hMPA}\\
C(000) vs C(001) vs C(010) & 1.1836E-07 & 1.8929E-07 & 9.9048E-08 & 1.9503E-08 \\
C(000) vs C(001) vs C(011) & 2.5493E-08 & 3.6908E-10 & 6.0286E-11 & 5.0316E-09 \\
C(000) vs C(001) vs C(100) & 1.6095E-03 & 1.2958E-04 & 2.5980E-01 & 3.1602E-02 \\
C(000) vs C(001) vs C(101) & 8.0169E-05 & 1.2958E-04 & 4.3372E-02 & 1.5784E-01 \\
C(000) vs C(001) vs C(110) & 5.9608E-06 & 6.9635E-08 & 9.9048E-08 & 5.0316E-09 \\
C(000) vs C(001) vs C(111) & 2.1252E-09 & 3.6908E-10 & 6.0286E-11 & 5.0316E-09 \\
C(000) vs C(010) vs C(011) & 1.0198E-09 & 2.5350E-11 & 2.2058E-10 & 6.0696E-08 \\
C(000) vs C(010) vs C(100) & 8.1053E-10 & 6.7767E-10 & 6.7767E-10 & 4.3672E-10 \\
C(000) vs C(010) vs C(101) & 1.8947E-06 & 1.5814E-07 & 3.8056E-08 & 7.0802E-09 \\
C(000) vs C(010) vs C(110) & 9.4464E-10 & 4.3672E-10 & 5.5161E-10 & 4.3672E-10 \\
C(000) vs C(010) vs C(111) & 1.2086E-10 & 6.3634E-11 & 2.2058E-10 & 4.7579E-08 \\
C(000) vs C(011) vs C(100) & 8.1053E-10 & 4.6662E-13 & 6.7767E-10 & 4.3672E-10 \\
C(000) vs C(011) vs C(101) & 2.6014E-08 & 2.3898E-10 & 4.1178E-10 & 7.0802E-09 \\
C(000) vs C(011) vs C(110) & 3.5924E-09 & 2.5350E-11 & 9.5723E-11 & 3.4791E-08 \\
C(000) vs C(011) vs C(111) & 2.0349E-11 & 1.5917E-11 & 1.2099E-11 & 4.3672E-10 \\
C(000) vs C(100) vs C(101) & 5.6103E-06 & 1.4428E-04 & 2.4980E-01 & 2.8116E-02 \\
C(000) vs C(100) vs C(110) & 8.1053E-10 & 6.7767E-10 & 6.7767E-10 & 4.3672E-10 \\
C(000) vs C(100) vs C(111) & 2.5305E-11 & 4.6662E-13 & 2.0349E-11 & 4.3672E-10 \\
C(000) vs C(101) vs C(110) & 1.6233E-07 & 1.5814E-07 & 3.8056E-08 & 7.0802E-09 \\
C(000) vs C(101) vs C(111) & 3.4669E-09 & 2.3898E-10 & 4.1178E-10 & 7.0802E-09 \\
C(000) vs C(110) vs C(111) & 1.7235E-10 & 5.2368E-11 & 9.5723E-11 & 3.5353E-08 \\
C(001) vs C(010) vs C(011) & 1.3966E-06 & 2.8597E-09 & 2.8434E-09 & 2.1454E-06 \\
C(001) vs C(010) vs C(100) & 2.1699E-07 & 1.8929E-07 & 1.3342E-07 & 2.8341E-08 \\
C(001) vs C(010) vs C(101) & 2.6989E-03 & 4.6993E-05 & 2.9231E-07 & 3.5006E-07 \\
C(001) vs C(010) vs C(110) & 5.0376E-04 & 3.9245E-06 & 4.4697E-06 & 1.1964E-08 \\
C(001) vs C(010) vs C(111) & 3.2920E-07 & 7.9825E-09 & 2.8434E-09 & 1.6243E-06 \\
C(001) vs C(011) vs C(100) & 3.9760E-08 & 5.8617E-11 & 4.0912E-10 & 7.5233E-09 \\
C(001) vs C(011) vs C(101) & 1.4702E-06 & 4.3672E-10 & 5.5161E-10 & 1.8026E-08 \\
C(001) vs C(011) vs C(110) & 6.8232E-05 & 1.5392E-09 & 1.1117E-09 & 4.2569E-07 \\
C(001) vs C(011) vs C(111) & 1.9884E-08 & 5.5161E-10 & 1.2099E-11 & 1.1964E-08 \\
C(001) vs C(100) vs C(101) & 7.3111E-03 & 1.4236E-04 & 1.1390E-01 & 1.8316E-02 \\
C(001) vs C(100) vs C(110) & 5.6103E-06 & 6.9635E-08 & 1.3342E-07 & 7.5233E-09 \\
C(001) vs C(100) vs C(111) & 3.3983E-09 & 5.8617E-11 & 4.0912E-10 & 7.5233E-09 \\
C(001) vs C(101) vs C(110) & 2.0615E-02 & 4.6993E-05 & 4.4697E-06 & 1.8026E-08 \\
C(001) vs C(101) vs C(111) & 2.2354E-06 & 4.3672E-10 & 5.5161E-10 & 1.8026E-08 \\
C(001) vs C(110) vs C(111) & 1.0702E-06 & 2.0372E-08 & 1.1117E-09 & 4.4217E-07 \\
C(010) vs C(011) vs C(100) & 1.0198E-09 & 3.8152E-12 & 1.2340E-09 & 6.0696E-08 \\
C(010) vs C(011) vs C(101) & 1.9825E-05 & 2.8597E-09 & 6.7840E-09 & 7.4265E-07 \\
C(010) vs C(011) vs C(110) & 1.3235E-06 & 5.1091E-12 & 1.4653E-09 & 1.0170E-01 \\
C(010) vs C(011) vs C(111) & 2.4570E-08 & 3.1261E-10 & 3.1111E-08 & 1.0170E-01 \\
C(010) vs C(100) vs C(101) & 6.3304E-07 & 1.7797E-07 & 3.8056E-08 & 7.0802E-09 \\
C(010) vs C(100) vs C(110) & 9.4464E-10 & 4.3672E-10 & 5.5161E-10 & 4.3672E-10 \\
C(010) vs C(100) vs C(111) & 1.2086E-10 & 9.9218E-12 & 1.2340E-09 & 4.7579E-08 \\
C(010) vs C(101) vs C(110) & 3.9426E-02 & 4.6993E-05 & 2.9231E-07 & 1.1964E-08 \\
C(010) vs C(101) vs C(111) & 2.8051E-05 & 7.9825E-09 & 6.7840E-09 & 5.8216E-07 \\
C(010) vs C(110) vs C(111) & 3.2129E-08 & 5.3665E-10 & 3.7681E-08 & 5.4701E-02 \\
C(011) vs C(100) vs C(101) & 4.3613E-08 & 3.6211E-11 & 2.2660E-09 & 7.0802E-09 \\
C(011) vs C(100) vs C(110) & 3.5924E-09 & 3.8152E-12 & 5.5445E-10 & 3.4791E-08 \\
C(011) vs C(100) vs C(111) & 6.7767E-10 & 3.4966E-13 & 4.3672E-10 & 4.3672E-10 \\
C(011) vs C(101) vs C(110) & 6.1343E-05 & 2.8597E-09 & 2.8434E-09 & 4.2569E-07 \\
C(011) vs C(101) vs C(111) & 7.7960E-06 & 5.5161E-10 & 4.3672E-10 & 1.1964E-08 \\
C(011) vs C(110) vs C(111) & 9.9861E-07 & 1.8162E-11 & 1.1640E-09 & 6.1327E-03 \\
C(100) vs C(101) vs C(110) & 3.8682E-07 & 1.7797E-07 & 3.8056E-08 & 7.0802E-09 \\
C(100) vs C(101) vs C(111) & 5.6028E-09 & 3.6211E-11 & 2.2660E-09 & 7.0802E-09 \\
C(100) vs C(110) vs C(111) & 1.7235E-10 & 7.9041E-12 & 5.5445E-10 & 3.5353E-08 \\
C(101) vs C(110) vs C(111) & 4.2675E-06 & 3.9202E-08 & 2.8434E-09 & 4.4217E-07 \\
\bottomrule
\end{tabular}%
}
\end{table}

\newpage

\clearpage

\begin{table}[h!]
\setlength{\heavyrulewidth}{0.3pt}
\setlength{\lightrulewidth}{0.3pt}
\setlength{\cmidrulewidth}{0.3pt}
\centering
\tiny
\renewcommand{\arraystretch}{0.86}
\setlength{\tabcolsep}{4pt}
\noindent\textit{Table~\ref{tab:triples_edsc} (continued).}\par\vspace{4pt}
\resizebox{\linewidth}{!}{%
\begin{tabular}{lcccc}
\toprule
Triple & $dim=10$ & $dim=30$ & $dim=50$ & $dim=100$  \\
\midrule
\textbf{Algorithm: MPA}\\
C(000) vs C(001) vs C(010) & 2.5638E-05 & 9.0629E-08 & 2.6355E-08 & 2.7633E-08
\\
C(000) vs C(001) vs C(011) & 8.5193E-08 & 3.6095E-10 & 1.4769E-09 & 1.6661E-09
\\
C(000) vs C(001) vs C(100) & 9.0419E-06 & 7.6436E-04 & 2.2313E-01 & 6.0810E-02
\\
C(000) vs C(001) vs C(101) & 2.6009E-06 & 1.4649E-04 & 2.3588E-01 & 2.6598E-01
\\
C(000) vs C(001) vs C(110) & 7.5353E-06 & 9.0629E-08 & 2.6355E-08 & 2.7633E-08
\\
C(000) vs C(001) vs C(111) & 1.9770E-09 & 3.6095E-10 & 1.4769E-09 & 1.6661E-09
\\
C(000) vs C(010) vs C(011) & 2.8585E-08 & 3.2726E-11 & 2.5617E-09 & 6.0696E-08
\\
C(000) vs C(010) vs C(100) & 3.8401E-09 & 4.3672E-10 & 5.5161E-10 & 6.7767E-10
\\
C(000) vs C(010) vs C(101) & 6.7052E-06 & 4.9746E-08 & 5.0988E-08 & 7.1829E-08
\\
C(000) vs C(010) vs C(110) & 5.5161E-10 & 4.3672E-10 & 4.3672E-10 & 4.3672E-10
\\
C(000) vs C(010) vs C(111) & 7.4356E-09 & 2.5350E-11 & 4.9606E-09 & 6.0696E-08
\\
C(000) vs C(011) vs C(100) & 4.5697E-11 & 2.5437E-13 & 5.5161E-10 & 6.7767E-10
\\
C(000) vs C(011) vs C(101) & 2.8826E-07 & 8.7470E-10 & 2.9990E-09 & 4.4762E-09
\\
C(000) vs C(011) vs C(110) & 1.0032E-08 & 2.5350E-11 & 2.5617E-09 & 6.0696E-08
\\
C(000) vs C(011) vs C(111) & 5.5161E-10 & 1.2099E-11 & 4.3672E-10 & 4.3672E-10
\\
C(000) vs C(100) vs C(101) & 4.9721E-05 & 7.6436E-04 & 2.2313E-01 & 3.7769E-01
\\
C(000) vs C(100) vs C(110) & 1.4226E-09 & 4.3672E-10 & 5.5161E-10 & 6.7767E-10
\\
C(000) vs C(100) vs C(111) & 2.5305E-11 & 2.5437E-13 & 5.5161E-10 & 6.7767E-10
\\
C(000) vs C(101) vs C(110) & 2.6402E-06 & 4.9746E-08 & 5.0988E-08 & 7.1829E-08
\\
C(000) vs C(101) vs C(111) & 5.0796E-09 & 2.3693E-10 & 2.9990E-09 & 4.4762E-09
\\
C(000) vs C(110) vs C(111) & 7.7270E-10 & 1.2017E-10 & 2.5617E-09 & 4.6809E-08
\\
C(001) vs C(010) vs C(011) & 1.7722E-04 & 1.1043E-08 & 1.5098E-08 & 5.8216E-07
\\
C(001) vs C(010) vs C(100) & 6.5233E-07 & 6.0236E-08 & 2.6355E-08 & 3.8056E-08
\\
C(001) vs C(010) vs C(101) & 1.0325E-01 & 3.9245E-06 & 2.9231E-07 & 2.4873E-07
\\
C(001) vs C(010) vs C(110) & 1.8604E-01 & 3.9245E-06 & 2.4873E-07 & 2.4873E-07
\\
C(001) vs C(010) vs C(111) & 2.1862E-05 & 1.5392E-09 & 3.1343E-08 & 5.8216E-07
\\
C(001) vs C(011) vs C(100) & 7.2478E-09 & 3.6211E-11 & 1.4769E-09 & 2.2660E-09
\\
C(001) vs C(011) vs C(101) & 9.4675E-05 & 1.1964E-08 & 5.5161E-10 & 4.3672E-10
\\
C(001) vs C(011) vs C(110) & 2.3583E-04 & 1.5392E-09 & 1.5098E-08 & 5.8216E-07
\\
C(001) vs C(011) vs C(111) & 3.8728E-06 & 4.3672E-10 & 4.3672E-10 & 4.3672E-10
\\
C(001) vs C(100) vs C(101) & 2.8444E-07 & 7.6436E-04 & 2.4527E-01 & 4.8204E-01
\\
C(001) vs C(100) vs C(110) & 5.8823E-07 & 6.0236E-08 & 2.6355E-08 & 3.8056E-08
\\
C(001) vs C(100) vs C(111) & 1.8588E-10 & 3.6211E-11 & 1.4769E-09 & 2.2660E-09
\\
C(001) vs C(101) vs C(110) & 6.2491E-02 & 3.9245E-06 & 2.9231E-07 & 2.4873E-07
\\
C(001) vs C(101) vs C(111) & 5.1977E-07 & 4.3672E-10 & 5.5161E-10 & 4.3672E-10
\\
C(001) vs C(110) vs C(111) & 1.5544E-04 & 7.8193E-09 & 1.5098E-08 & 4.4217E-07
\\
C(010) vs C(011) vs C(100) & 3.9418E-10 & 4.9298E-12 & 2.5617E-09 & 6.0696E-08
\\
C(010) vs C(011) vs C(101) & 3.3231E-04 & 1.1043E-08 & 1.5098E-08 & 5.8216E-07
\\
C(010) vs C(011) vs C(110) & 6.9360E-06 & 7.0110E-12 & 6.2487E-07 & 2.7645E-01
\\
C(010) vs C(011) vs C(111) & 3.3500E-06 & 7.0110E-12 & 6.2487E-07 & 2.7645E-01
\\
C(010) vs C(100) vs C(101) & 3.6529E-06 & 8.4568E-08 & 3.8056E-08 & 6.6519E-08
\\
C(010) vs C(100) vs C(110) & 4.3672E-10 & 4.3672E-10 & 4.3672E-10 & 4.3672E-10
\\
C(010) vs C(100) vs C(111) & 7.1703E-11 & 3.8152E-12 & 4.9606E-09 & 6.0696E-08
\\
C(010) vs C(101) vs C(110) & 1.8604E-01 & 3.9245E-06 & 2.4873E-07 & 2.4873E-07
\\
C(010) vs C(101) vs C(111) & 2.1036E-04 & 1.5392E-09 & 3.1343E-08 & 5.8216E-07
\\
C(010) vs C(110) vs C(111) & 7.6655E-05 & 1.1593E-10 & 6.2487E-07 & 2.7738E-01
\\
C(011) vs C(100) vs C(101) & 2.6014E-08 & 2.3693E-10 & 2.2660E-09 & 3.8657E-09
\\
C(011) vs C(100) vs C(110) & 7.7270E-10 & 3.8152E-12 & 2.5617E-09 & 6.0696E-08
\\
C(011) vs C(100) vs C(111) & 2.5437E-13 & 2.5437E-13 & 4.3672E-10 & 4.3672E-10
\\
C(011) vs C(101) vs C(110) & 4.4102E-04 & 7.9825E-09 & 1.5098E-08 & 5.8216E-07
\\
C(011) vs C(101) vs C(111) & 1.5607E-06 & 1.1964E-08 & 4.3672E-10 & 4.3672E-10
\\
C(011) vs C(110) vs C(111) & 1.5710E-04 & 1.1593E-10 & 6.2487E-07 & 3.9479E-01
\\
C(100) vs C(101) vs C(110) & 1.1150E-05 & 8.4568E-08 & 3.8056E-08 & 6.6519E-08
\\
C(100) vs C(101) vs C(111) & 5.7146E-10 & 5.9146E-11 & 2.2660E-09 & 3.8657E-09
\\
C(100) vs C(110) vs C(111) & 1.3918E-10 & 1.9382E-11 & 2.5617E-09 & 4.6809E-08
\\
C(101) vs C(110) vs C(111) & 2.9640E-04 & 7.8193E-09 & 1.5098E-08 & 4.4217E-07
\\
\bottomrule
\end{tabular}%
}
\end{table}


For DSC, all 56 triples have $p<0.05$ at $dim=10$ and $30$ for both algorithms (see Table~\ref{tab:triples_dsc}). At $dim=50$, the counts are 52 for hMPA and 54 for MPA. At $dim=100$, both counts are 52. All nonsignificant DSC triples contain only configurations from: $C(000)$, $C(001)$, $C(100)$, $C(101)$. For eDSC, all 56 hMPA triples have $p<0.05$ at $dim=10$ and $30$ (see Table~\ref{tab:triples_edsc}). The counts are 53 at $dim=50$ and 52 at $dim=100$. The corresponding MPA counts are 52, 56, 52 and 48.

\subsection{Control comparisons}
\label{sec7.3}

The control protocol compares $C(000)$ with each remaining
configuration. Tables~\ref{tab:control_dsc} and~\ref{tab:control_edsc}
report the unadjusted and Bonferroni-adjusted Wilcoxon $p$-values (e.g., Derrac in \cite{JD2011}) for
DSC and eDSC (e.g., Eftimov and Koro{\v{s}}ec in \cite{TE2019}), respectively. The final row of each table gives the
overall control $p$-value.

\begin{table}[ht]
\setlength{\heavyrulewidth}{0.3pt}
\setlength{\lightrulewidth}{0.3pt}
\setlength{\cmidrulewidth}{0.3pt}
\centering
\scriptsize
\setlength{\tabcolsep}{4pt}
\caption{Wilcoxon and Bonferroni-adjusted $p$-values for the DSC
control comparisons.}
\label{tab:control_dsc}
\resizebox{\linewidth}{!}{%
\begin{tabular}{lcc@{\hspace{0.35cm}}cc@{\hspace{0.35cm}}cc@{\hspace{0.35cm}}cc}
\toprule
 & \multicolumn{2}{c}{$dim=10$} & \multicolumn{2}{c}{$dim=30$} & \multicolumn{2}{c}{$dim=50$} & \multicolumn{2}{c}{$dim=100$} \\
\cmidrule(lr){2-3} \cmidrule(lr){4-5} \cmidrule(lr){6-7} \cmidrule(lr){8-9}
Comparison & Wilcoxon & Bonferroni & Wilcoxon & Bonferroni & Wilcoxon & Bonferroni & Wilcoxon & Bonferroni \\
\midrule
\textbf{Algorithm: hMPA}\\
C(000) vs C(001) &  1.4E-02 & 9.8E-02 & 2.4E-02 & 1.6E-01 & 1.1E-01 & 7.7E-01 & 2.3E-02 & 1.6E-01
\\
C(000) vs C(010) & 7.8E-08 & 5.4E-07 & 5.7E-07 & 4.0E-06 & 3.7E-06 & 2.6E-05 & 2.1E-05 & 1.4E-04
\\
C(000) vs C(011) & 7.8E-08 & 5.4E-07 & 5.7E-07 & 4.0E-06 & 5.7E-07 & 4.0E-06 & 3.7E-06 & 2.6E-05
\\
C(000) vs C(100) & 4.2E-01 & 1.0E+00 & 7.2E-02 & 5.0E-01 & 4.2E-01 & 1.0E+00 & 3.7E-02 & 2.6E-01
\\
C(000) vs C(101) & 8.9E-04 & 6.2E-03 & 5.2E-04 & 3.6E-03 & 1.5E-01 & 1.0E+00 & 3.9E-02 & 2.8E-01
\\
C(000) vs C(110) & 7.8E-08 & 5.4E-07 & 5.7E-07 & 4.0E-06 & 3.7E-06 & 2.6E-05 & 2.1E-05 & 1.4E-04
\\
C(000) vs C(111) & 7.8E-08 & 5.4E-07 & 5.7E-07 & 4.0E-06 & 5.7E-07 & 4.0E-06 & 3.7E-06 & 2.6E-05
\\
\midrule
\textbf{Overall control $p$}  & 4.3E-01 & - & 9.4E-02 & - & 5.6E-01 & - & 9.7E-02 & -
\\
\bottomrule
\end{tabular}
}
\vspace{0.4cm}
\resizebox{\linewidth}{!}{%
\begin{tabular}{lcc@{\hspace{0.35cm}}cc@{\hspace{0.35cm}}cc@{\hspace{0.35cm}}cc}
\toprule
 & \multicolumn{2}{c}{$dim=10$} & \multicolumn{2}{c}{$dim=30$} & \multicolumn{2}{c}{$dim=50$} & \multicolumn{2}{c}{$dim=100$} \\
\cmidrule(lr){2-3} \cmidrule(lr){4-5} \cmidrule(lr){6-7} \cmidrule(lr){8-9}
Comparison & Wilcoxon & Bonferroni & Wilcoxon & Bonferroni & Wilcoxon & Bonferroni & Wilcoxon & Bonferroni \\
\midrule
\textbf{Algorithm: MPA}\\
C(000) vs C(001) & 1.8E-03 & 1.2E-02 & 5.0E-03 & 3.5E-02 & 2.4E-02 & 1.6E-01 & 1.1E-01 & 7.7E-01
\\
C(000) vs C(010) & 7.8E-08 & 5.4E-07 & 5.7E-07 & 4.0E-06 & 3.7E-06 & 2.6E-05 & 3.7E-06 & 2.6E-05
\\
C(000) vs C(011) & 7.8E-08 & 5.4E-07 & 5.7E-07 & 4.0E-06 & 5.7E-07 & 4.0E-06 & 5.7E-07 & 4.0E-06
\\
C(000) vs C(100) & 3.6E-02 & 2.5E-01 & 1.0E+00 & 1.0E+00 & 1.0E+00 & 1.0E+00 & 7.7E-01 & 1.0E+00
\\
C(000) vs C(101) & 5.2E-02 & 3.6E-01 & 8.3E-03 & 5.8E-02 & 2.7E-01 & 1.0E+00 & 5.7E-01 & 1.0E+00
\\
C(000) vs C(110) & 7.8E-08 & 5.4E-07 & 5.7E-07 & 4.0E-06 & 3.7E-06 & 2.6E-05 & 3.7E-06 & 2.6E-05
\\
C(000) vs C(111) & 7.8E-08 & 5.4E-07 & 5.7E-07 & 4.0E-06 & 5.7E-07 & 4.0E-06 & 5.7E-07 & 4.0E-06
\\
\midrule
\textbf{Overall control $p$} & 8.7E-02 & -- & 1.0E+00 & -- & 1.0E+00 & -- & 9.1E-01 & --
\\
\bottomrule
\end{tabular}
}
\end{table}

\begin{table}[ht]
\setlength{\heavyrulewidth}{0.3pt}
\setlength{\lightrulewidth}{0.3pt}
\setlength{\cmidrulewidth}{0.3pt}
\centering
\vspace{-0.8cm}
\scriptsize
\setlength{\tabcolsep}{4pt}
\caption{Wilcoxon and Bonferroni-adjusted $p$-values for the eDSC
control comparisons.}
\label{tab:control_edsc}
\resizebox{\linewidth}{!}{%
\begin{tabular}{lcc@{\hspace{0.35cm}}cc@{\hspace{0.35cm}}cc@{\hspace{0.35cm}}cc}
\toprule
 & \multicolumn{2}{c}{$dim=10$} & \multicolumn{2}{c}{$dim=30$} & \multicolumn{2}{c}{$dim=50$} & \multicolumn{2}{c}{$dim=100$} \\
\cmidrule(lr){2-3} \cmidrule(lr){4-5} \cmidrule(lr){6-7} \cmidrule(lr){8-9}
Comparison & Wilcoxon & Bonferroni & Wilcoxon & Bonferroni & Wilcoxon & Bonferroni & Wilcoxon & Bonferroni \\
\midrule
\textbf{Algorithm: hMPA}\\
C(000) vs C(001) & 4.5E-03 & 3.1E-02 & 2.6E-03 & 1.8E-02 & 4.1E-02 & 2.9E-01 & 2.0E-02 & 1.4E-01
\\
C(000) vs C(010) & 3.7E-06 & 2.6E-05 & 3.7E-06 & 2.6E-05 & 3.7E-06 & 2.6E-05 & 3.7E-06 & 2.6E-05
\\
C(000) vs C(011) & 3.7E-06 & 2.6E-05 & 5.7E-07 & 4.0E-06 & 5.7E-07 & 4.0E-06 & 3.7E-06 & 2.6E-05
\\
C(000) vs C(100) & 7.2E-02 & 5.0E-01 & 2.3E-01 & 1.0E+00 & 3.5E-01 & 1.0E+00 & 1.0E+00 & 1.0E+00
\\
C(000) vs C(101) & 1.2E-04 & 8.6E-04 & 6.3E-04 & 4.4E-03 & 6.6E-02 & 4.6E-01 & 1.8E-01 & 1.0E+00
\\
C(000) vs C(110) & 3.7E-06 & 2.6E-05 & 3.7E-06 & 2.6E-05 & 3.7E-06 & 2.6E-05 & 3.7E-06 & 2.6E-05
\\
C(000) vs C(111) & 5.7E-07 & 4.0E-06 & 5.7E-07 & 4.0E-06 & 5.7E-07 & 4.0E-06 & 3.7E-06 & 2.6E-05
\\
\midrule
\textbf{Overall control $p$} & 7.6E-02 & - & 2.4E-01 & - & 4.1E-01 & - & 1.0E+00 & -
\\
\bottomrule
\end{tabular}
}
\resizebox{\linewidth}{!}{%
\begin{tabular}{lcc@{\hspace{0.35cm}}cc@{\hspace{0.35cm}}cc@{\hspace{0.35cm}}cc}
\toprule
 & \multicolumn{2}{c}{$dim=10$} & \multicolumn{2}{c}{$dim=30$} & \multicolumn{2}{c}{$dim=50$} & \multicolumn{2}{c}{$dim=100$} \\
\cmidrule(lr){2-3} \cmidrule(lr){4-5} \cmidrule(lr){6-7} \cmidrule(lr){8-9}
Comparison & Wilcoxon & Bonferroni & Wilcoxon & Bonferroni & Wilcoxon & Bonferroni & Wilcoxon & Bonferroni \\
\midrule
\textbf{Algorithm: MPA}\\
C(000) vs C(001) & 8.4E-03 & 5.9E-02 & 1.3E-02 & 9.4E-02 & 1.3E-01 & 9.1E-01 & 3.5E-01 & 1.0E+00
\\
C(000) vs C(010) & 2.1E-05 & 1.4E-04 & 3.7E-06 & 2.6E-05 & 3.7E-06 & 2.6E-05 & 3.7E-06 & 2.6E-05
\\
C(000) vs C(011) & 6.1E-06 & 4.3E-05 & 5.7E-07 & 4.0E-06 & 3.7E-06 & 2.6E-05 & 3.7E-06 & 2.6E-05
\\
C(000) vs C(100) & 1.5E-01 & 1.0E+00 & 1.0E+00 & 1.0E+00 & 1.0E+00 & 1.0E+00 & 3.5E-01 & 1.0E+00
\\
C(000) vs C(101) & 3.0E-04 & 2.1E-03 & 6.3E-04 & 4.4E-03 & 2.3E-01 & 1.0E+00 & 3.9E-01 & 1.0E+00
\\
C(000) vs C(110) & 6.1E-06 & 4.3E-05 & 3.7E-06 & 2.6E-05 & 3.7E-06 & 2.6E-05 & 3.7E-06 & 2.6E-05
\\
C(000) vs C(111) & 5.7E-07 & 4.0E-06 & 5.7E-07 & 4.0E-06 & 3.7E-06 & 2.6E-05 & 3.7E-06 & 2.6E-05
\\
\midrule
\textbf{Overall control $p$} & 1.6E-01 & -- & 1.0E+00 & -- & 1.0E+00 & -- & 7.4E-01 & --
\\
\bottomrule
\end{tabular}
}
\end{table}
\clearpage

\Bem{The symbol ''-'' in  denotes a non-applicable Bonferroni-adjusted value for the aggregated overall control $p$- value.}

\NOWE{In Tables:~\ref{tab:control_dsc} and~\ref{tab:control_edsc}, the symbol ``--'' indicates that the Bonferroni-adjusted aggregated overall control p-value is not applicable.}



In both schemes, the comparisons between $C(000)$ and $C(010)$, $C(011)$, $C(110)$ and $C(111)$ have a Bonferroni-adjusted $p$-value (e.g., Derrac in \cite{JD2011}) of less than 0.05. This holds for both algorithms at every dimension. However, comparisons with $C(100)$ are non-significant throughout.

For DSC (see Table~\ref{tab:control_dsc}), five comparisons are significant at $dim=10$ and $dim=30$ for both algorithms. The additional comparison involves $C(101)$ for hMPA and $C(001)$ for MPA. For eDSC (see Table~\ref{tab:control_edsc}), the counts at these dimensions are six for hMPA and five for MPA. The additional comparisons involve $C(001)$ and $C(101)$ for hMPA, and $C(101)$ for MPA. At $dim=50$ and $dim=100$, four comparisons are significant for both algorithms in both schemes. All reported overall control $p$-values exceed 0.05.


\subsection{Convergence Analysis}
\label{Sec7.4}
The convergence plots in Figure~1 complement the DSC and eDSC results. They illustrate the emergence of differences among the 8 configurations of the \texttt{parameterised CEC 2017 benchmark}. $F1$, $F9$, $F16$, and $F23$ represent the four function classes and capture distinct convergence patterns. The same functions are used across all dimensions (i.e., $dim\in\{10, \ 30, \ 50, \ 100\}$) to ensure comparability.
\vspace{0.4cm}

\begin{figure}[ht]
\centering
\noindent\textit{Figure~1(continued).}\par\vspace{2pt}
\vspace{0.3cm}
\begin{subfigure}
{0.49\textwidth}
    \centering
    \includegraphics[
        width=\linewidth,
        keepaspectratio
    ]{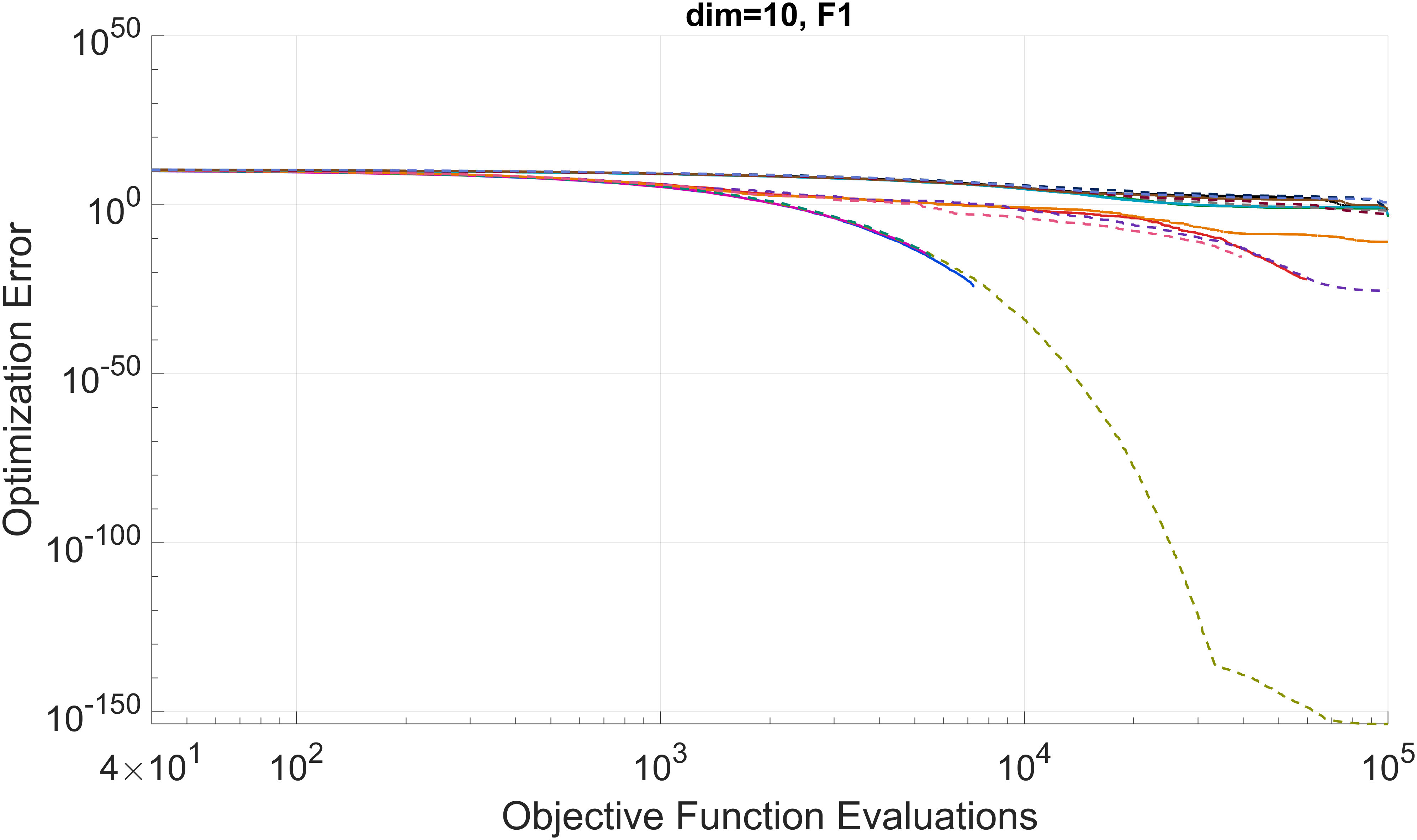}
\end{subfigure}
\hfill
\begin{subfigure}{0.49\textwidth}
    \centering
    \includegraphics[
        width=\linewidth,
        keepaspectratio
    ]{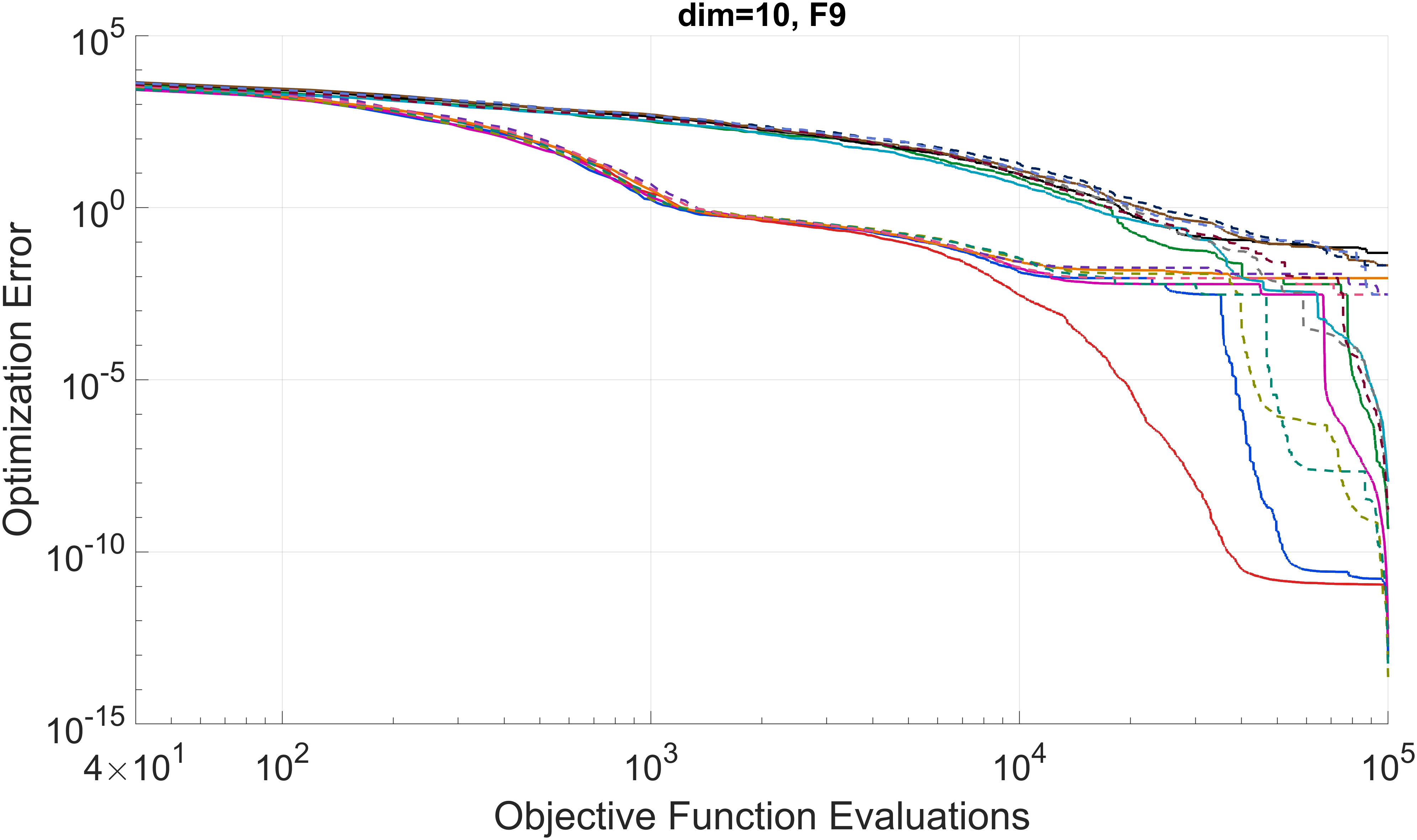}
\end{subfigure}

\end{figure}

\begin{figure}[ht]
\centering

\begin{subfigure}{0.49\textwidth}
    \centering
    \includegraphics[
        width=\linewidth,
        keepaspectratio
    ]{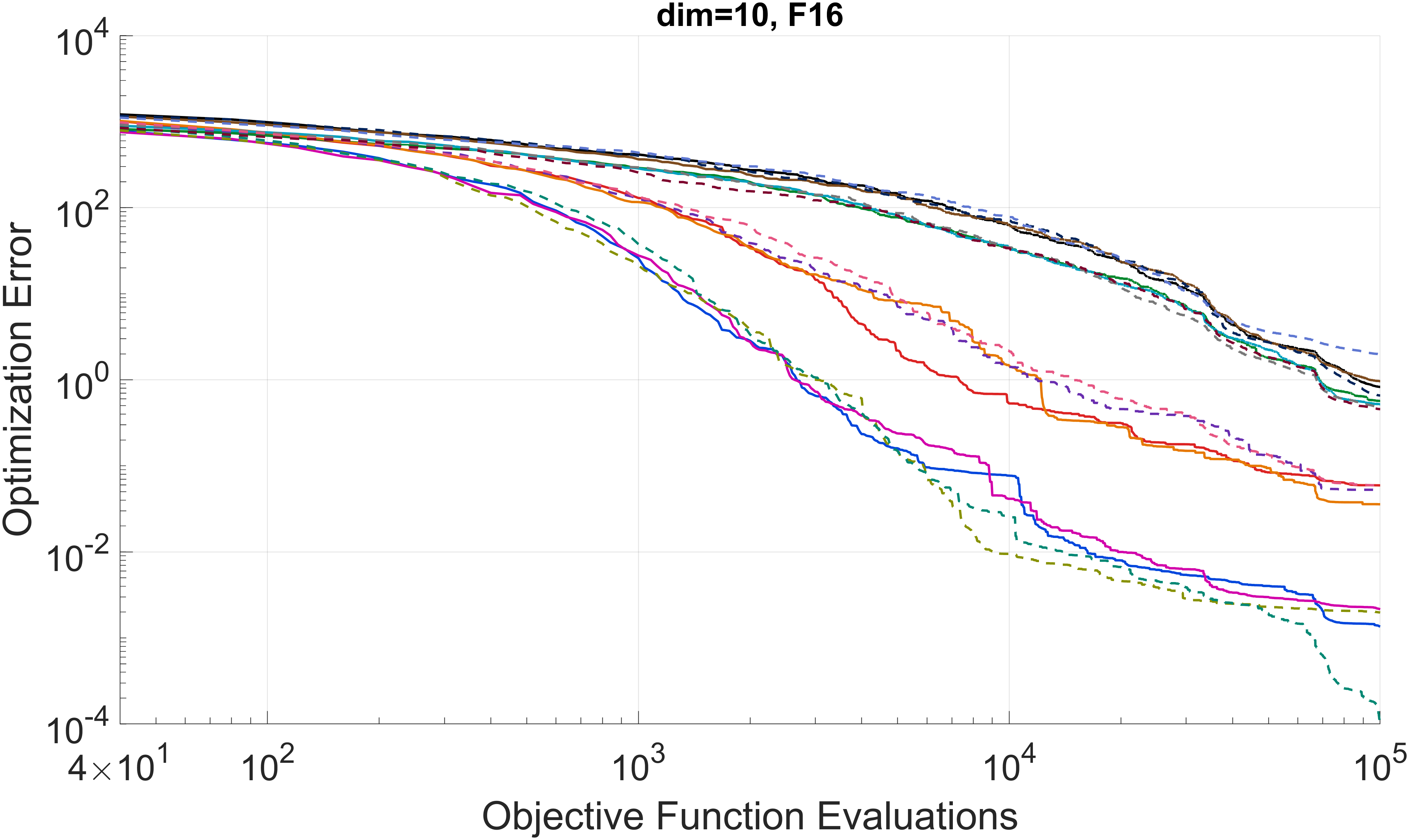}
\end{subfigure}
\hfill
\begin{subfigure}{0.49\textwidth}
    \centering
    \includegraphics[
        width=\linewidth,
        keepaspectratio
    ]{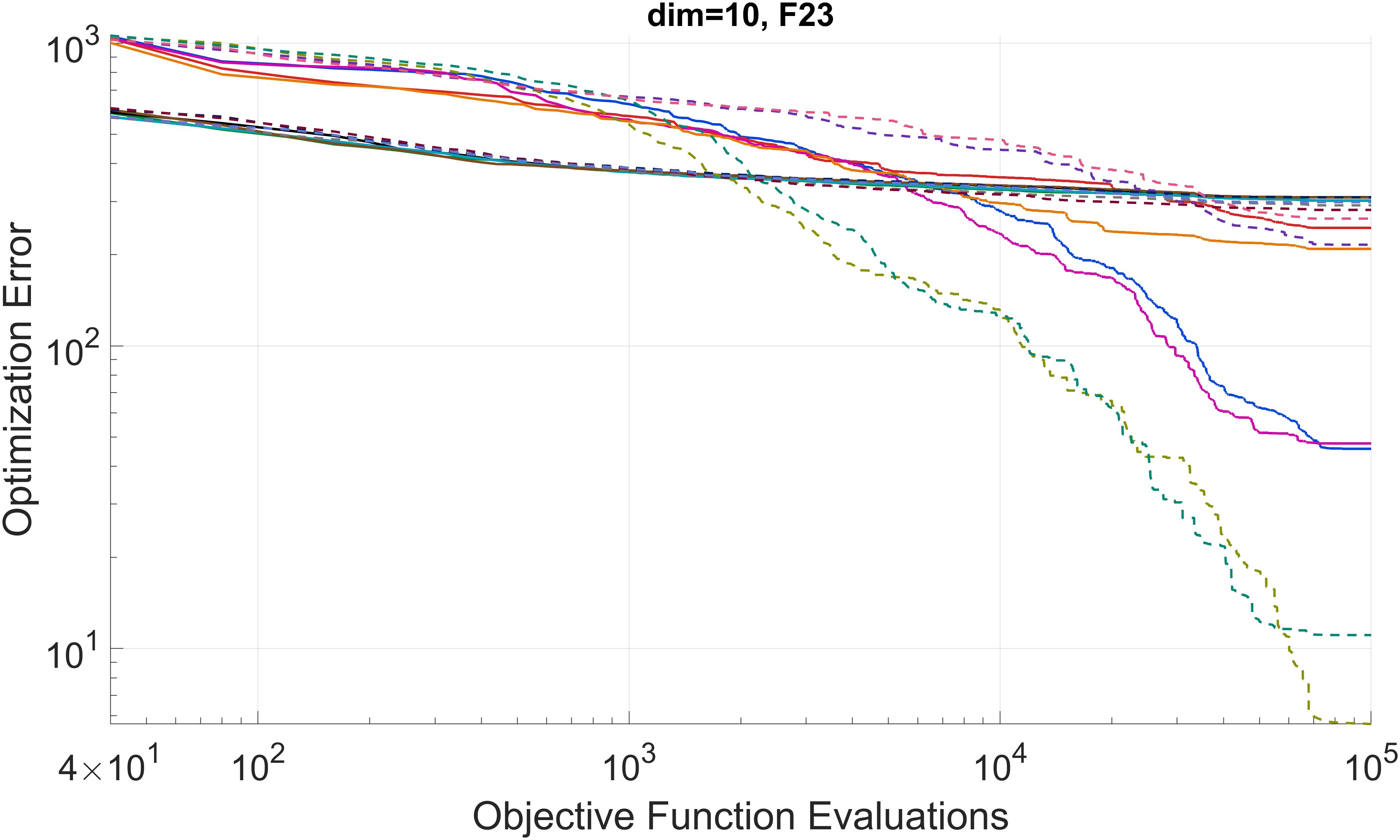}
\end{subfigure}
\end{figure}


\begin{figure}[ht]
\centering
\noindent\textit{Figure~1(continued).}\par\vspace{2pt}
\vspace{0.3cm}
\begin{subfigure}{0.49\textwidth}
    \centering
    \includegraphics[
        width=\linewidth,
        keepaspectratio
    ]{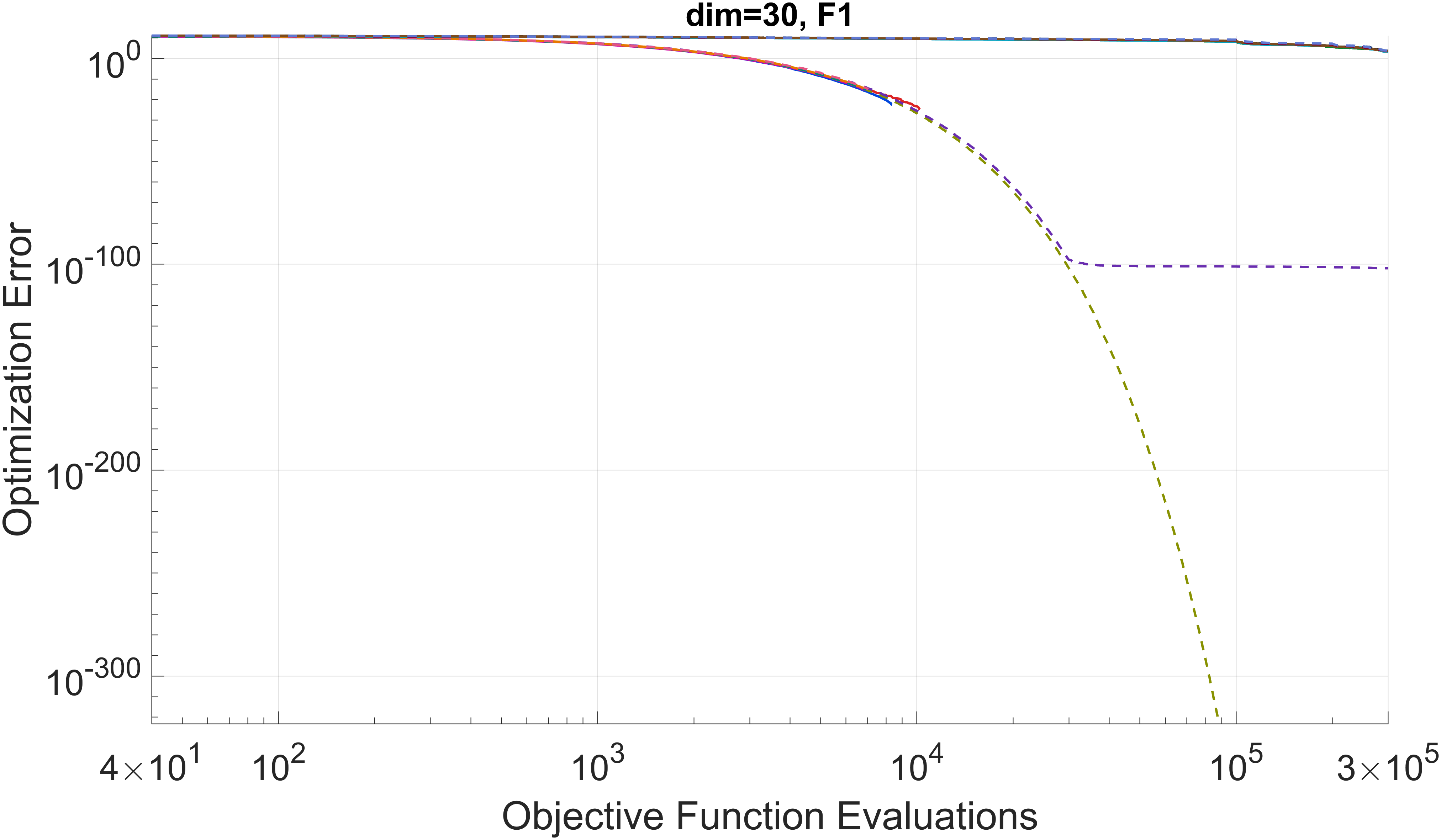}
\end{subfigure}
\hfill
\begin{subfigure}{0.49\textwidth}
    \centering
    \includegraphics[
        width=\linewidth,
        keepaspectratio
    ]{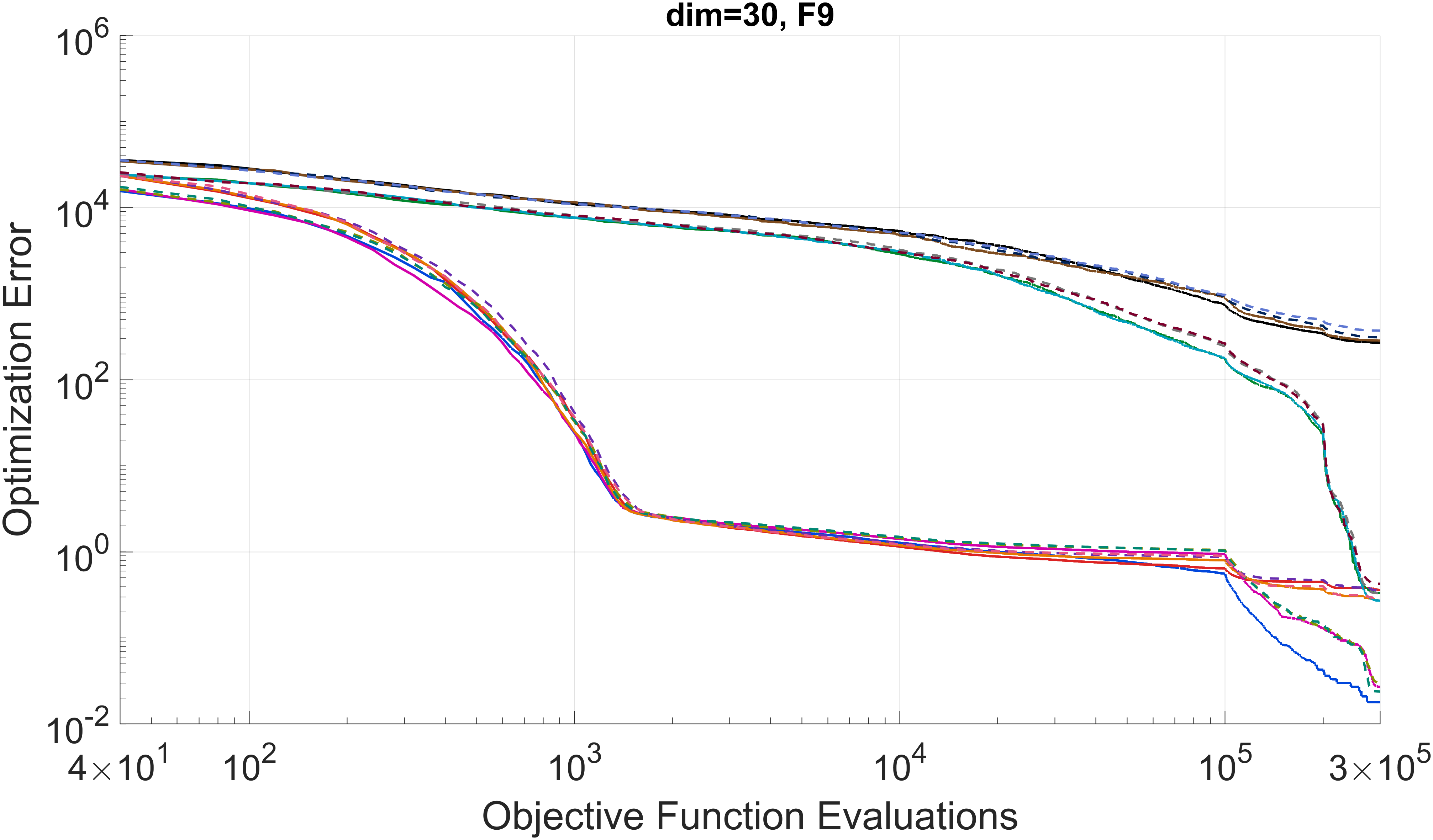}
\end{subfigure}

\end{figure}

\begin{figure}[ht]
\centering

\begin{subfigure}{0.49\textwidth}
    \centering
    \includegraphics[
        width=\linewidth,
        keepaspectratio
    ]{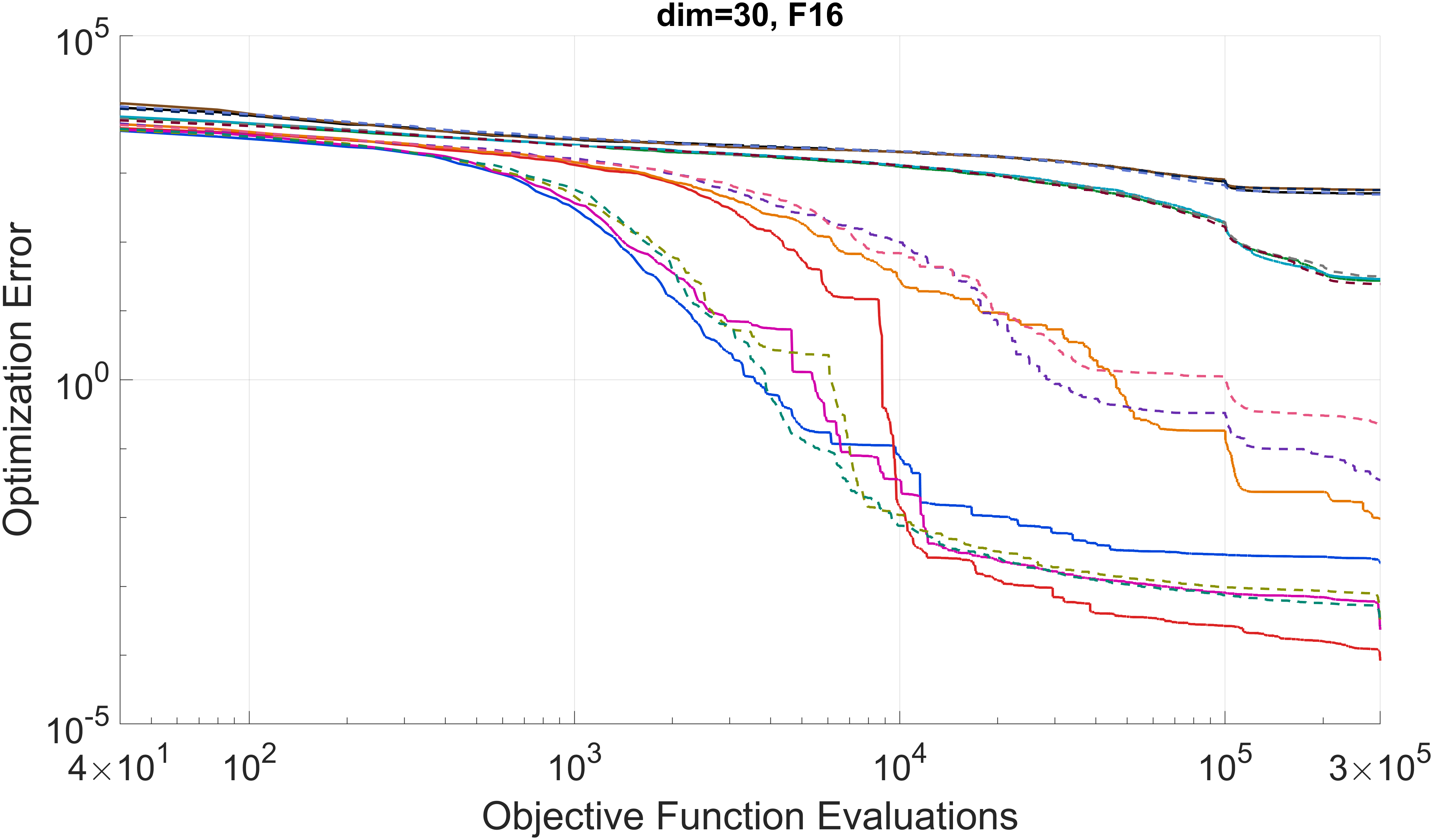}
\end{subfigure}
\hfill
\begin{subfigure}{0.49\textwidth}
    \centering
    \includegraphics[
        width=\linewidth,
        keepaspectratio
    ]{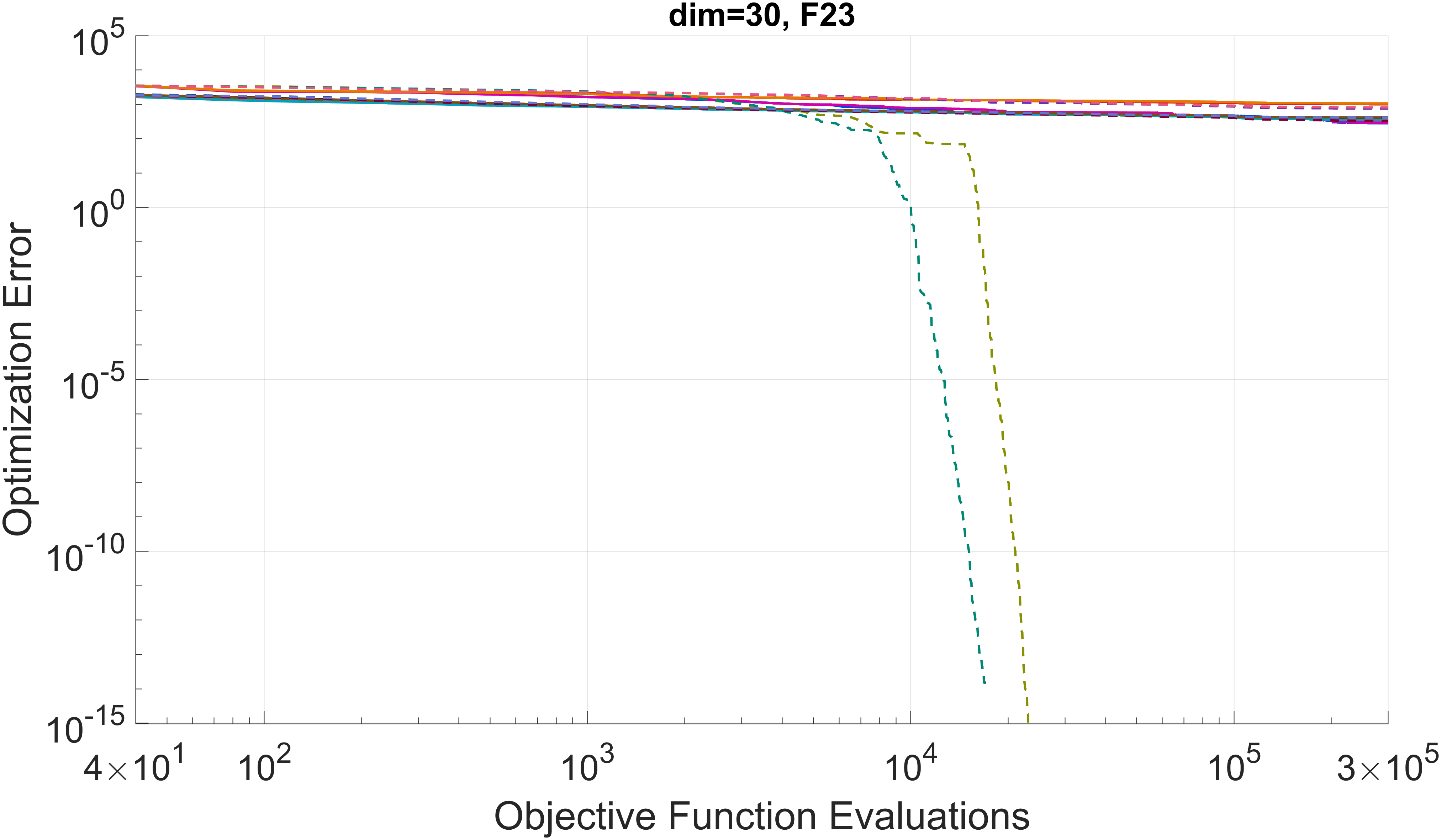}
\end{subfigure}

\end{figure}


\begin{figure}[ht]
\centering

\begin{subfigure}{0.49\textwidth}
    \centering
    \includegraphics[
        width=\linewidth,
        keepaspectratio
    ]{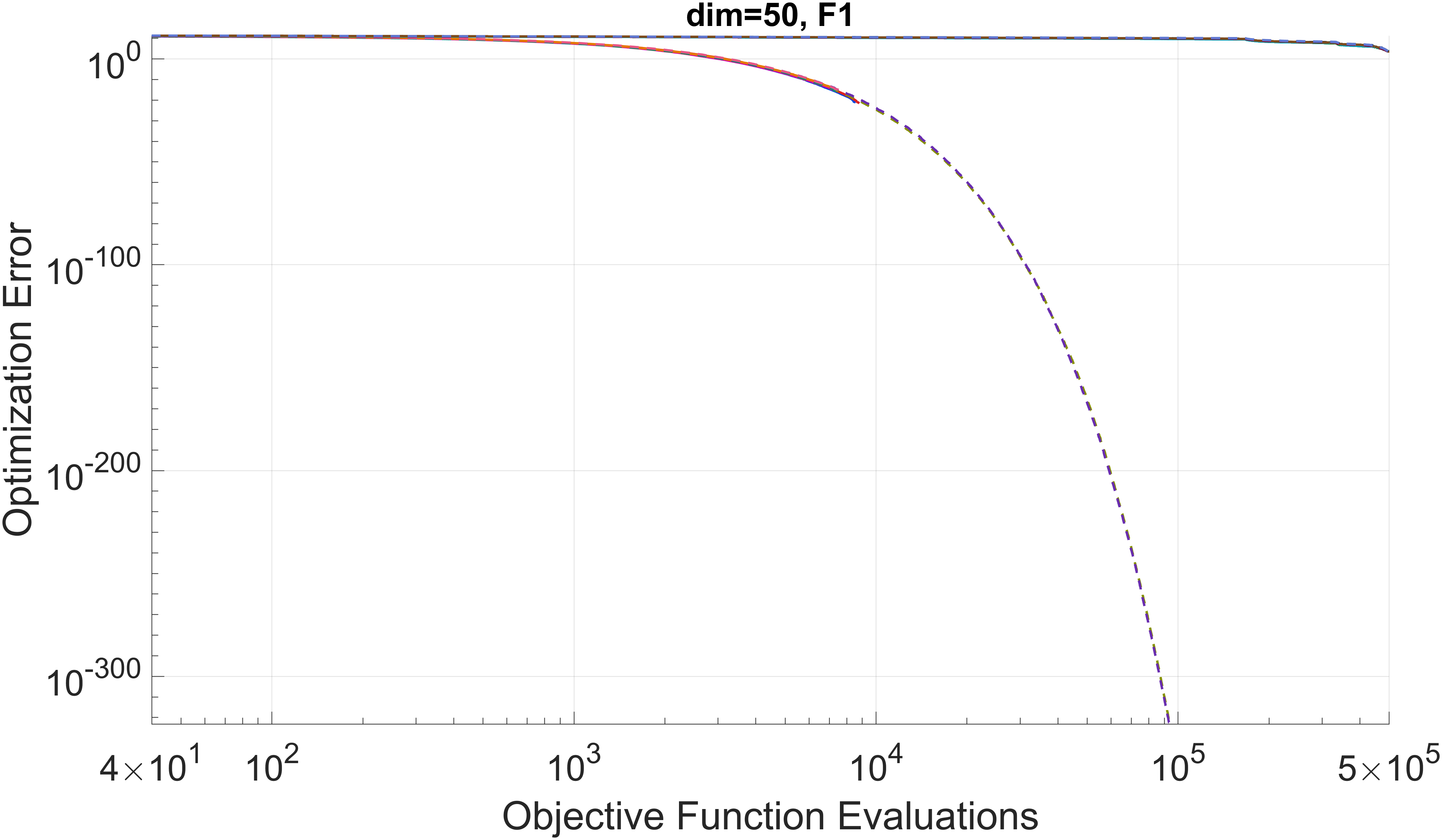}
\end{subfigure}
\hfill
\begin{subfigure}{0.49\textwidth}
    \centering
    \includegraphics[
        width=\linewidth,
        keepaspectratio
    ]{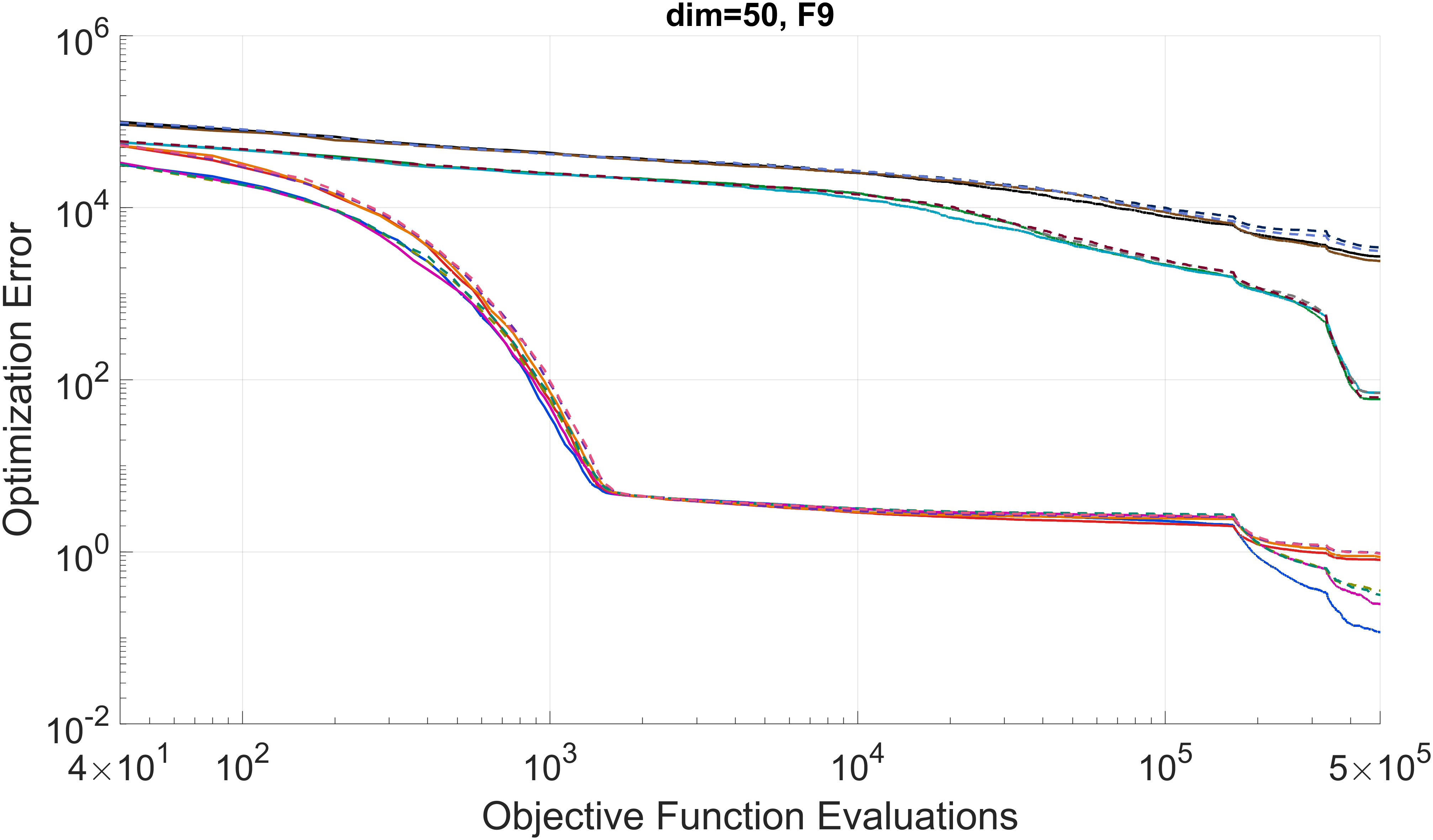}
\end{subfigure}

\end{figure}

\begin{figure}[ht]
\centering

\begin{subfigure}{0.49\textwidth}
    \centering
    \includegraphics[
        width=\linewidth,
        keepaspectratio
    ]{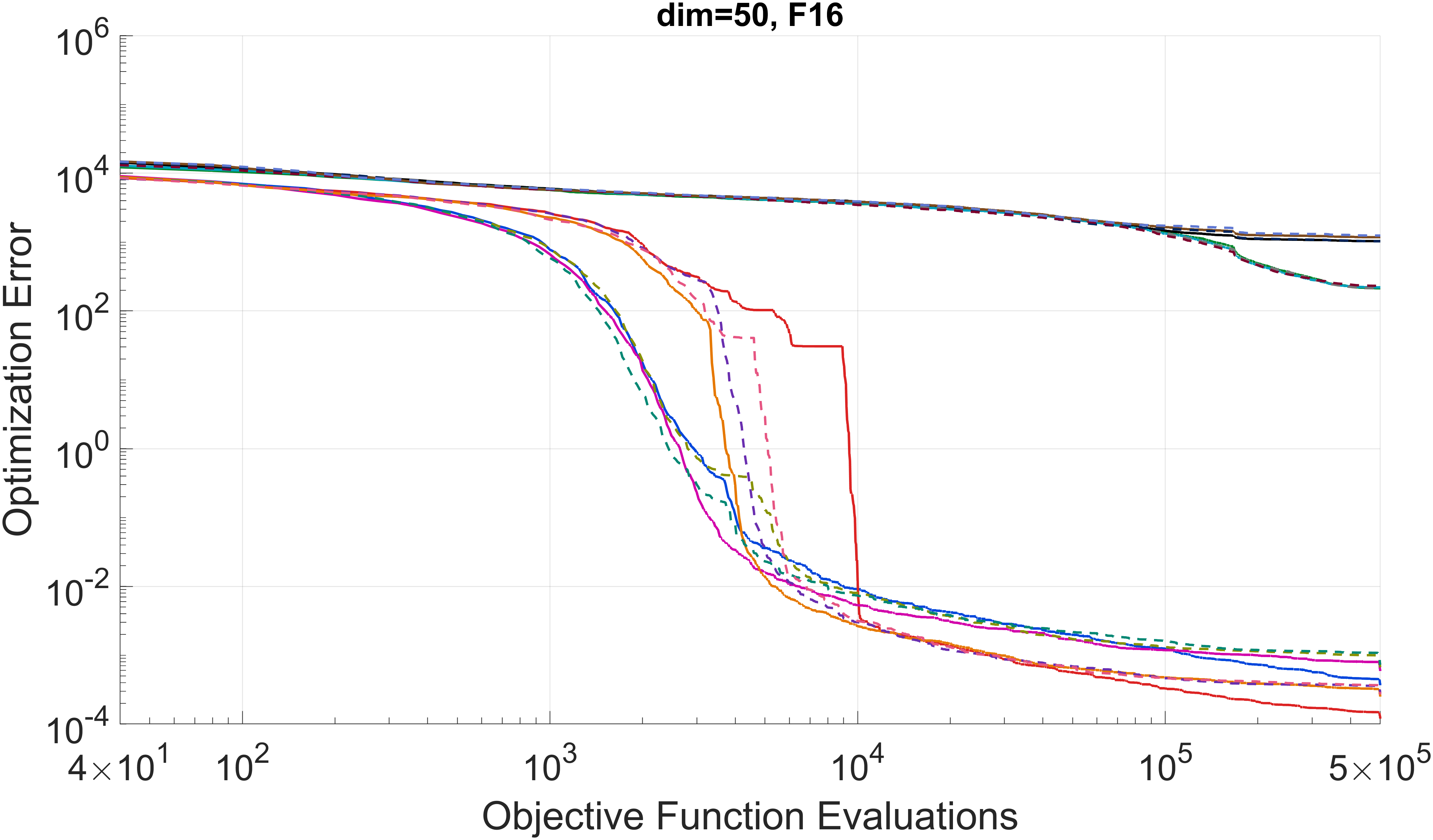}
\end{subfigure}
\hfill
\begin{subfigure}{0.49\textwidth}
    \centering
    \includegraphics[
        width=\linewidth,
        keepaspectratio
    ]{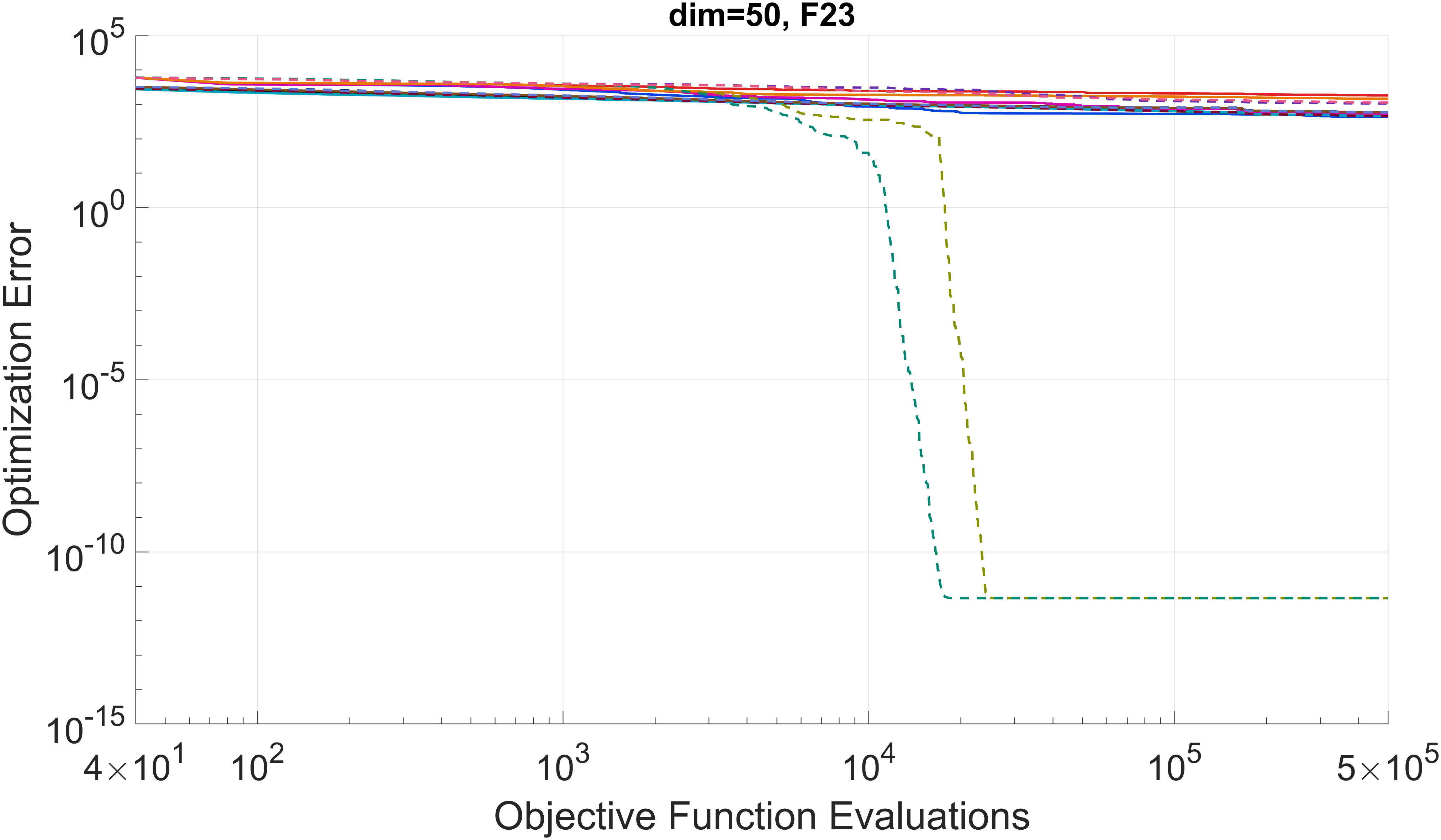}
\end{subfigure}

\end{figure}


\begin{figure}[ht]
\centering

\begin{subfigure}{0.49\textwidth}
    \centering
    \includegraphics[
        width=\linewidth,
        keepaspectratio
    ]{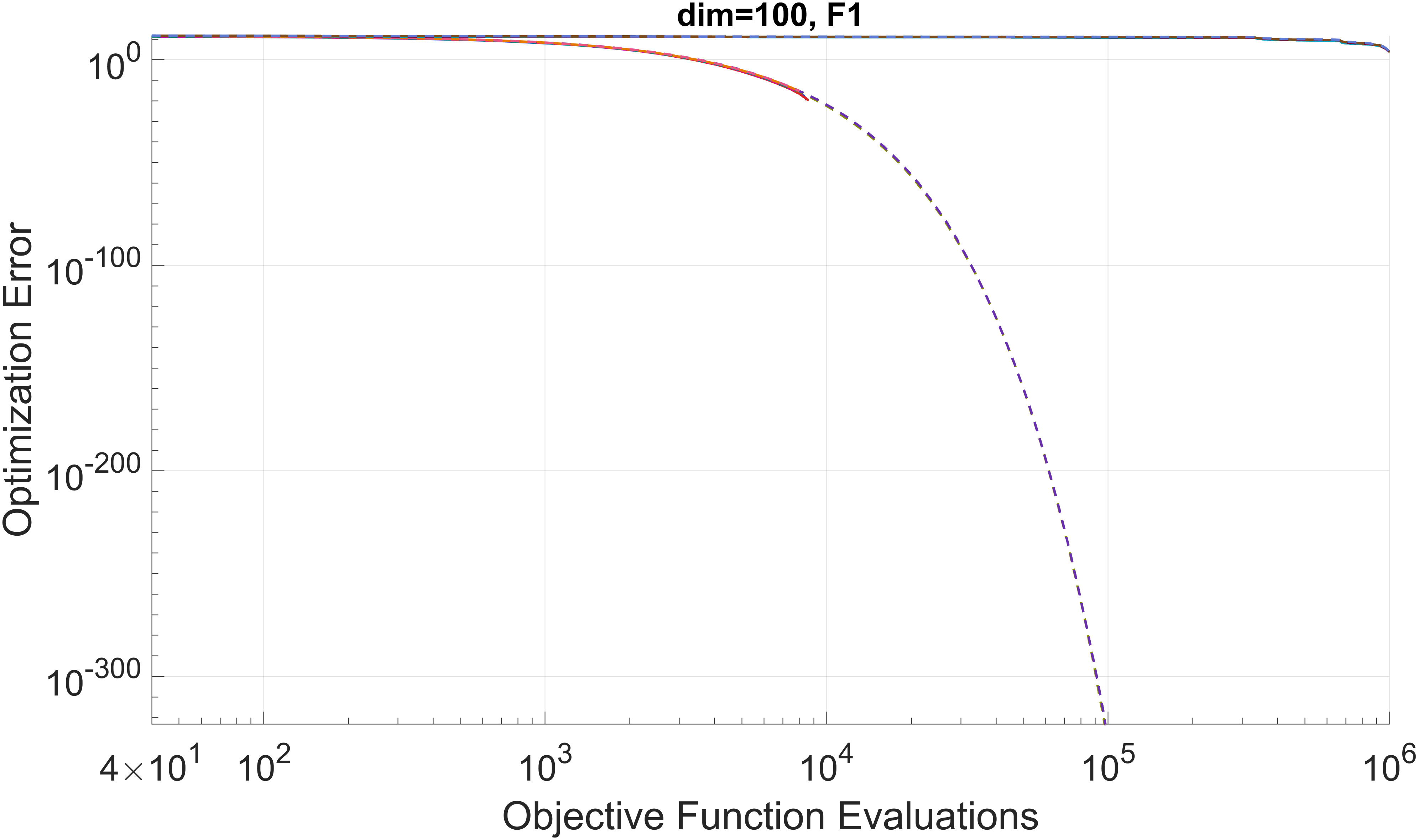}
\end{subfigure}
\hfill
\begin{subfigure}{0.49\textwidth}
    \centering
    \includegraphics[
        width=\linewidth,
        keepaspectratio
    ]{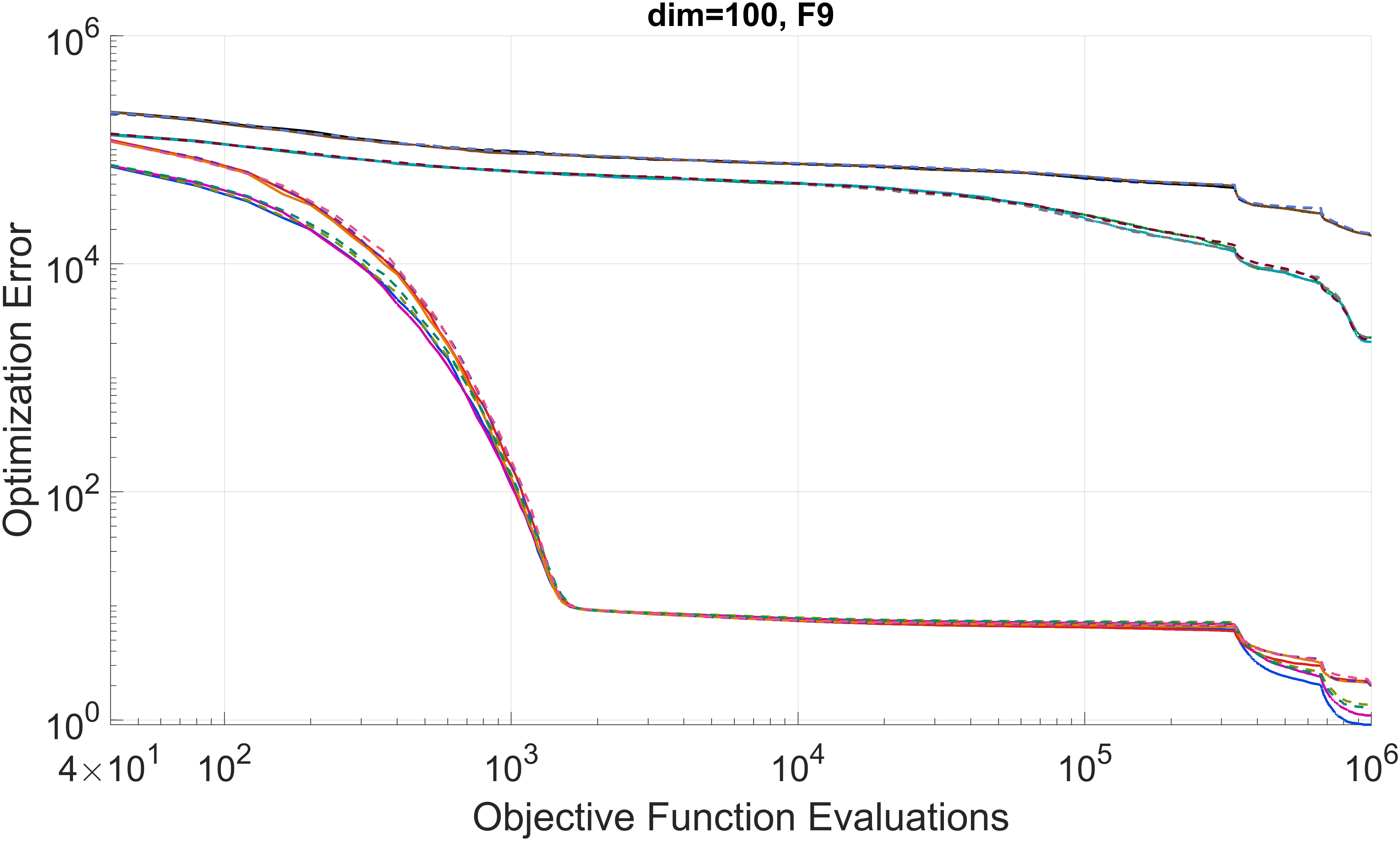}
\end{subfigure}

\end{figure}
\clearpage
\begin{figure}[ht]
\centering
\vspace{-1cm}
\begin{subfigure}{0.49\textwidth}
    \centering
    \includegraphics[
        width=\linewidth,
        keepaspectratio
    ]{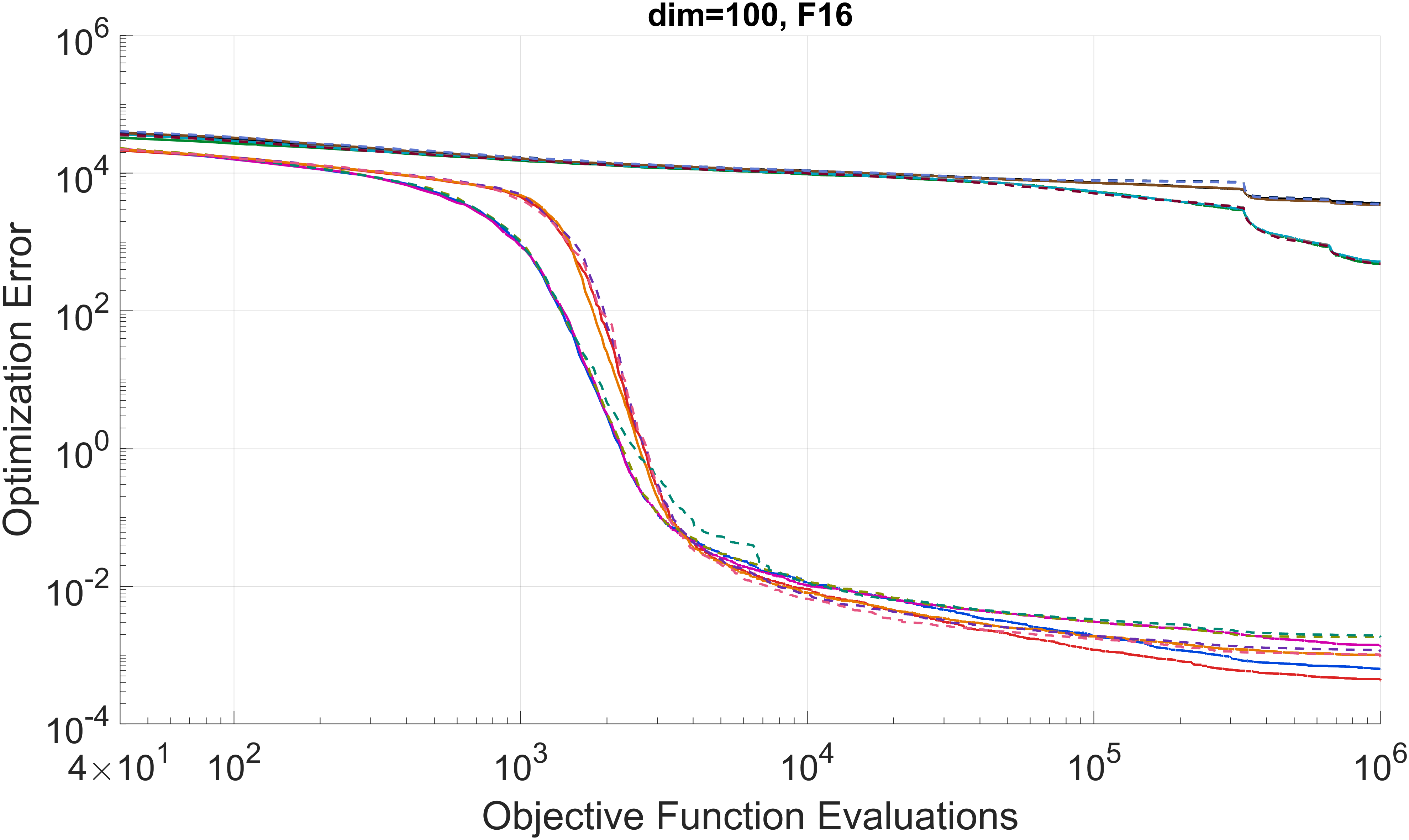}
\end{subfigure}
\hfill
\begin{subfigure}{0.49\textwidth}
    \centering
    \includegraphics[
        width=\linewidth,
        keepaspectratio
    ]{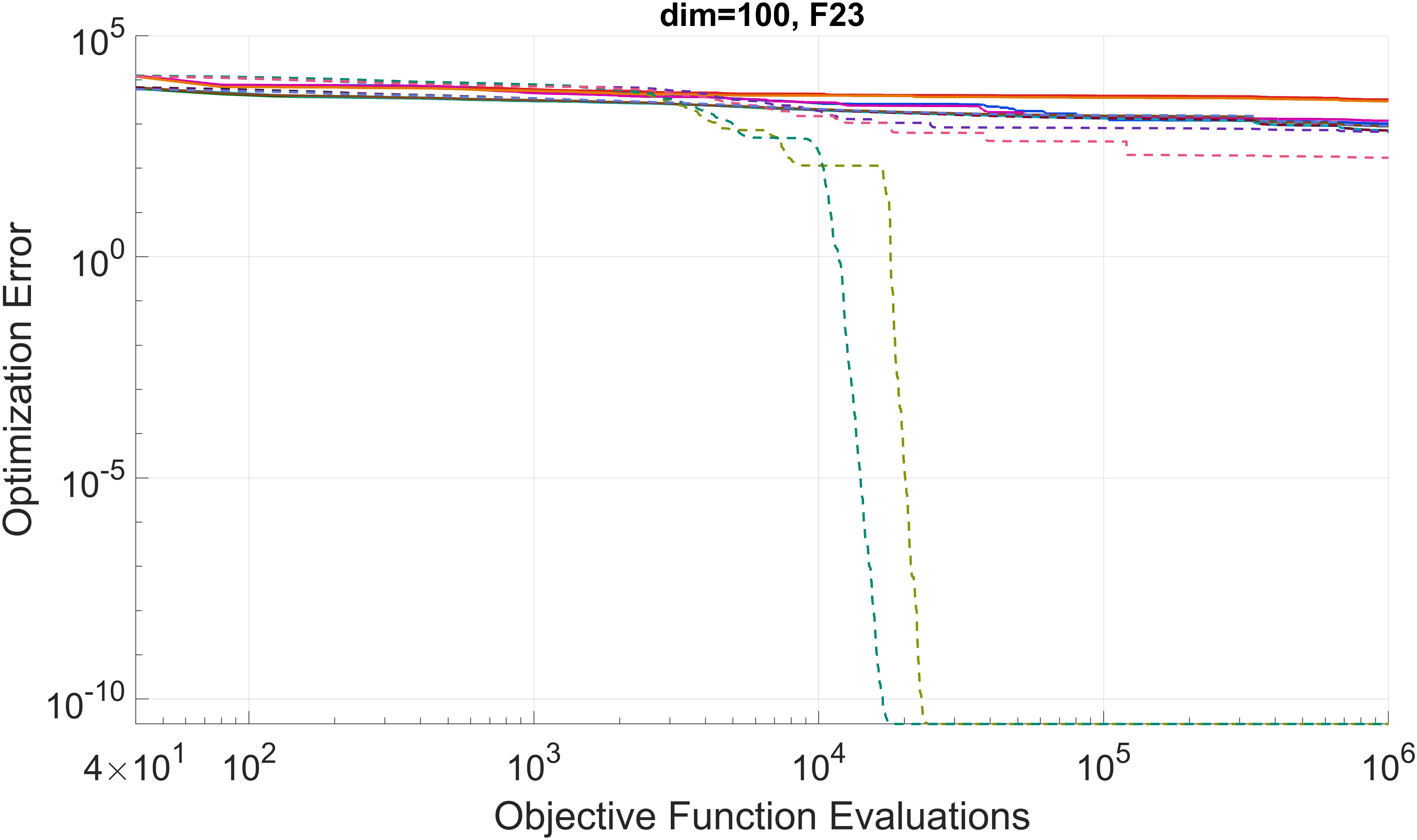}
\end{subfigure}

\vspace{0.6cm}

\includegraphics[
    width=0.7\textwidth,
    keepaspectratio
]{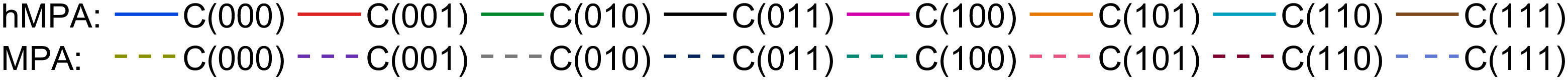}\\
\vspace{0.5cm}

Figure 1. Convergence profiles for $F1$, $F9$, $F16$, and $F23$ across the 8 parameterized 
CEC~2017 configurations at 
$dim\in\{10, \ 30, \ 50, \ 100\}.$
\end{figure}
The $FE$ axis (horizontal axis) starts at 40 across dimensions, which is the first logged point after two population evaluations of 20 members each. Both algorithms (i.e., hMPA and MPA), demonstrate significant error reduction with $C(000)$ and $C(001)$ for $F1$. For $F9$, hMPA achieves its lowest final errors with $C(000)$ at $dim\in\{30, \ 50\}$ and $dim=100$. MPA performs best with $C(000)$ at $dim = 10$ and $C(100)$ at higher dimensions (i.e., $dim\in\{30, \ 50, \ 100\}$). For $F16$, hMPA performs best with $C(000)$ at $dim = 10$. In higher dimensions, $C(001)$ yields the smallest final error. For $F23$, $C(000)$ is among hMPA’s best configurations at 
$dim\in\{10, \ 30,  \ 50\}$. By contrast, $C(010)$ and $C(110)$ lead at $dim=100$. The MPA achieves lower error levels than the hMPA with the configurations $C(000)$ and $C(100)$. The configuration $C(000)$ yields the lowest final error at $dim=10$. In particular, $C(100)$ reaches low error levels earlier at higher dimensions. For both algorithms, $C(111)$ trails the leaders in final error across $F1$, $F9$, $F16$ and $F23$.
These observations are discussed in Section~~8.

\subsection{Boxplot Analysis}
\label{sec7.5}
The boxplots supplement the convergence analysis (see Section \ref{Sec7.4}). They summarise the final optimisation errors for MPA and hMPA across the different runs. The same benchmark configurations, functions ($F1$, $F9$, $F16$, $F23$) and dimensions ($dim \in\{10, \ 30, \ 50, \ 100\}$) are used. Figure 2 shows medians, means, data variability, and outliers.

\begin{figure}[ht]
\centering
\noindent\textit{Figure~2(continued).}\par\vspace{2pt}
\vspace{0.35cm}
\begin{subfigure}{0.49\textwidth}
    \centering
    \includegraphics[
        width=\linewidth,
        keepaspectratio
    ]{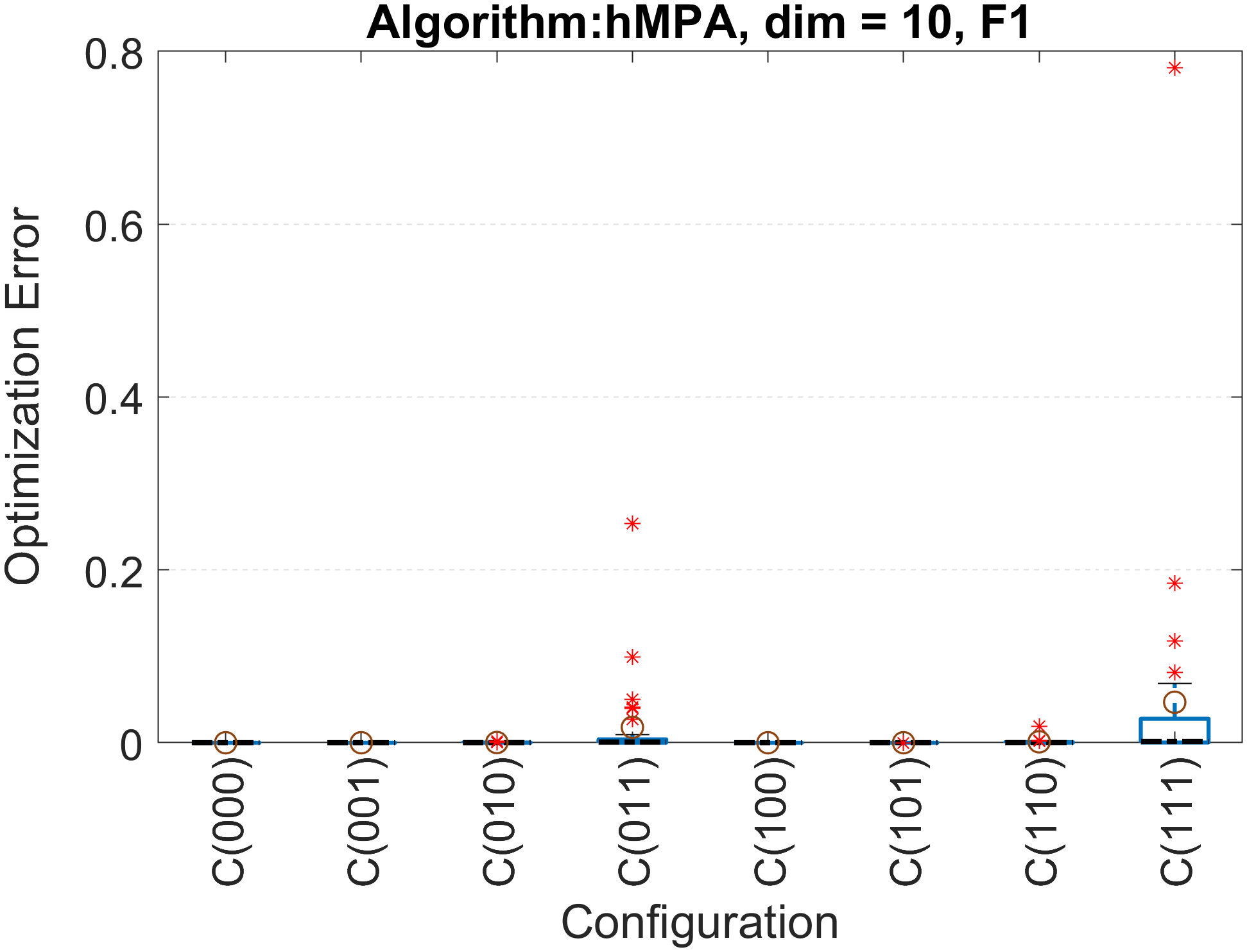}
\end{subfigure}
\hfill
\begin{subfigure}{0.49\textwidth}
    \centering
    \includegraphics[
        width=\linewidth,
        keepaspectratio
    ]{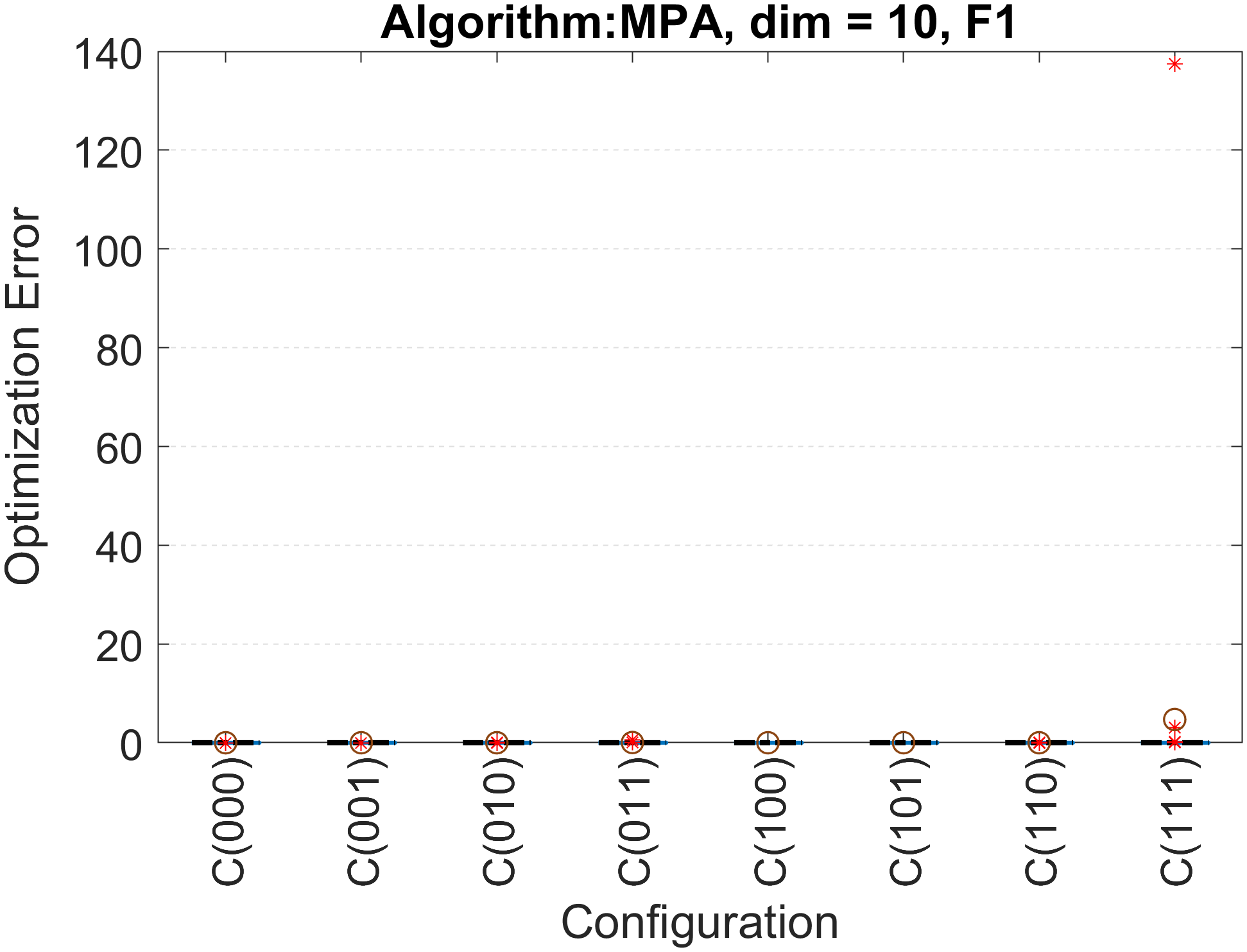}
\end{subfigure}

\end{figure}

\begin{figure}[ht]
\centering
\noindent\textit{Figure~2(continued).}\par\vspace{2pt}
\vspace{0.25cm}
\begin{subfigure}{0.49\textwidth}
    \centering
    \includegraphics[
        width=\linewidth,
        keepaspectratio
    ]{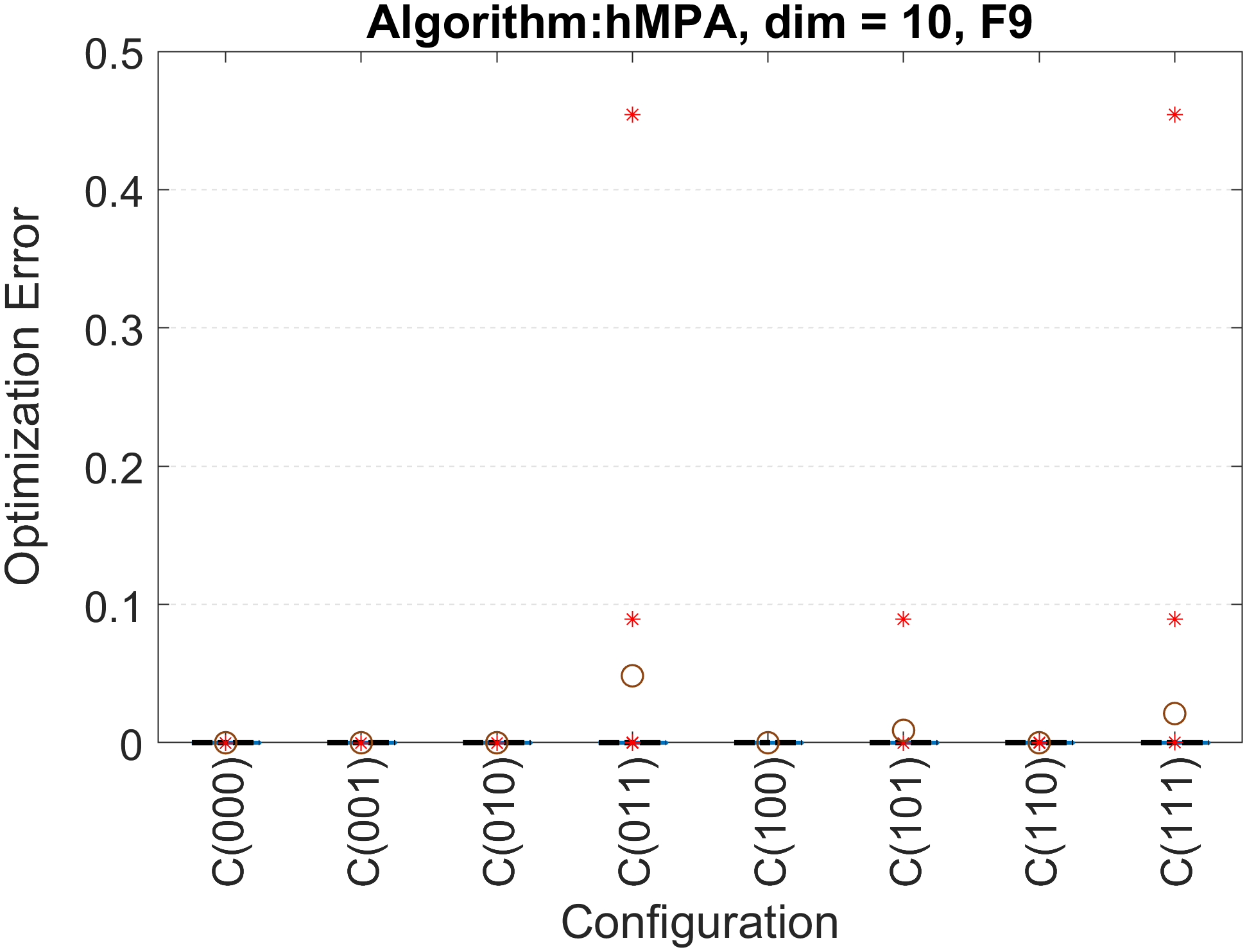}
\end{subfigure}
\hfill
\begin{subfigure}{0.49\textwidth}
    \centering
    \includegraphics[
        width=\linewidth,
        keepaspectratio
    ]{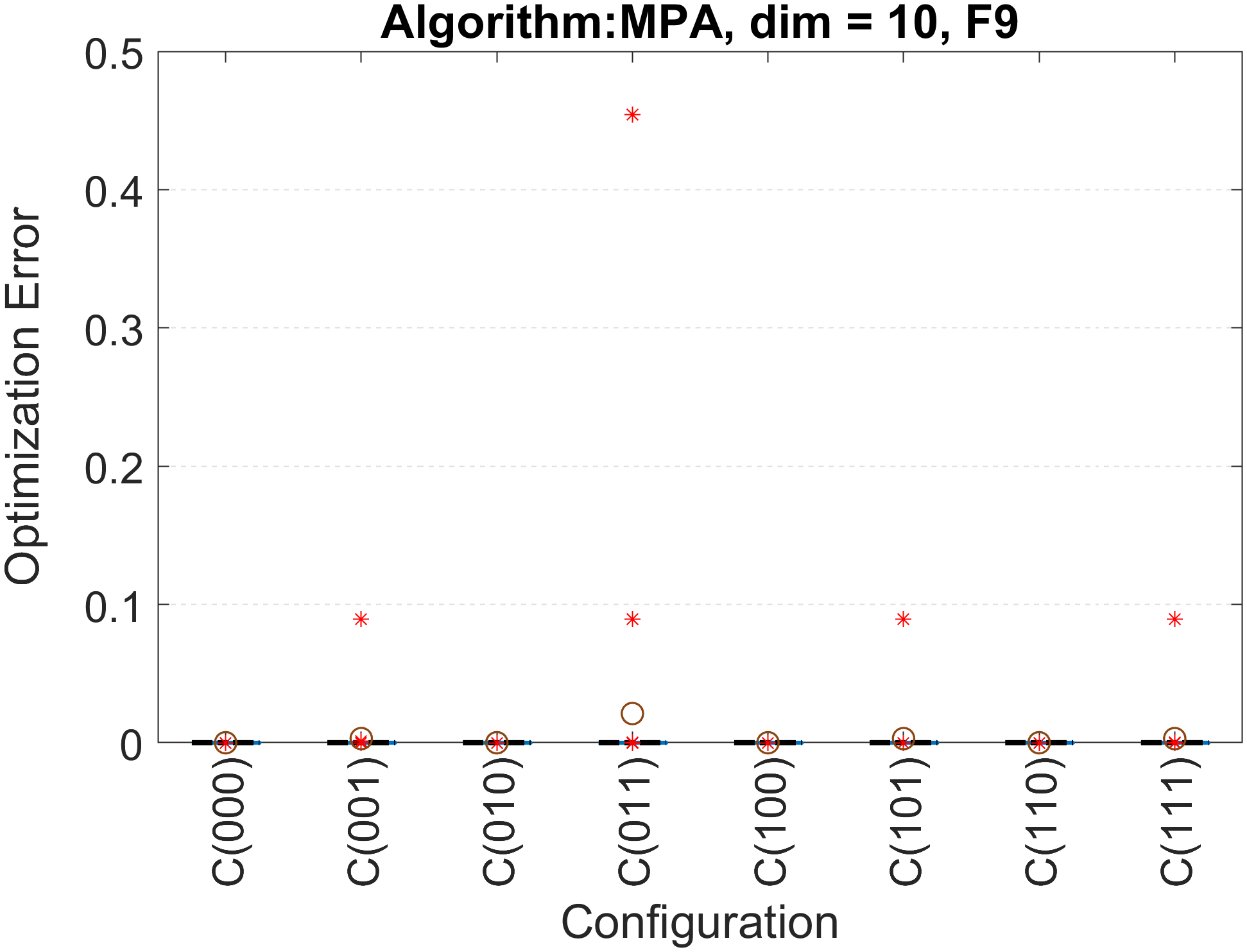}
\end{subfigure}

\end{figure}

\begin{figure}[ht]
\centering

\begin{subfigure}{0.49\textwidth}
    \centering
    \includegraphics[
        width=\linewidth,
        keepaspectratio
    ]{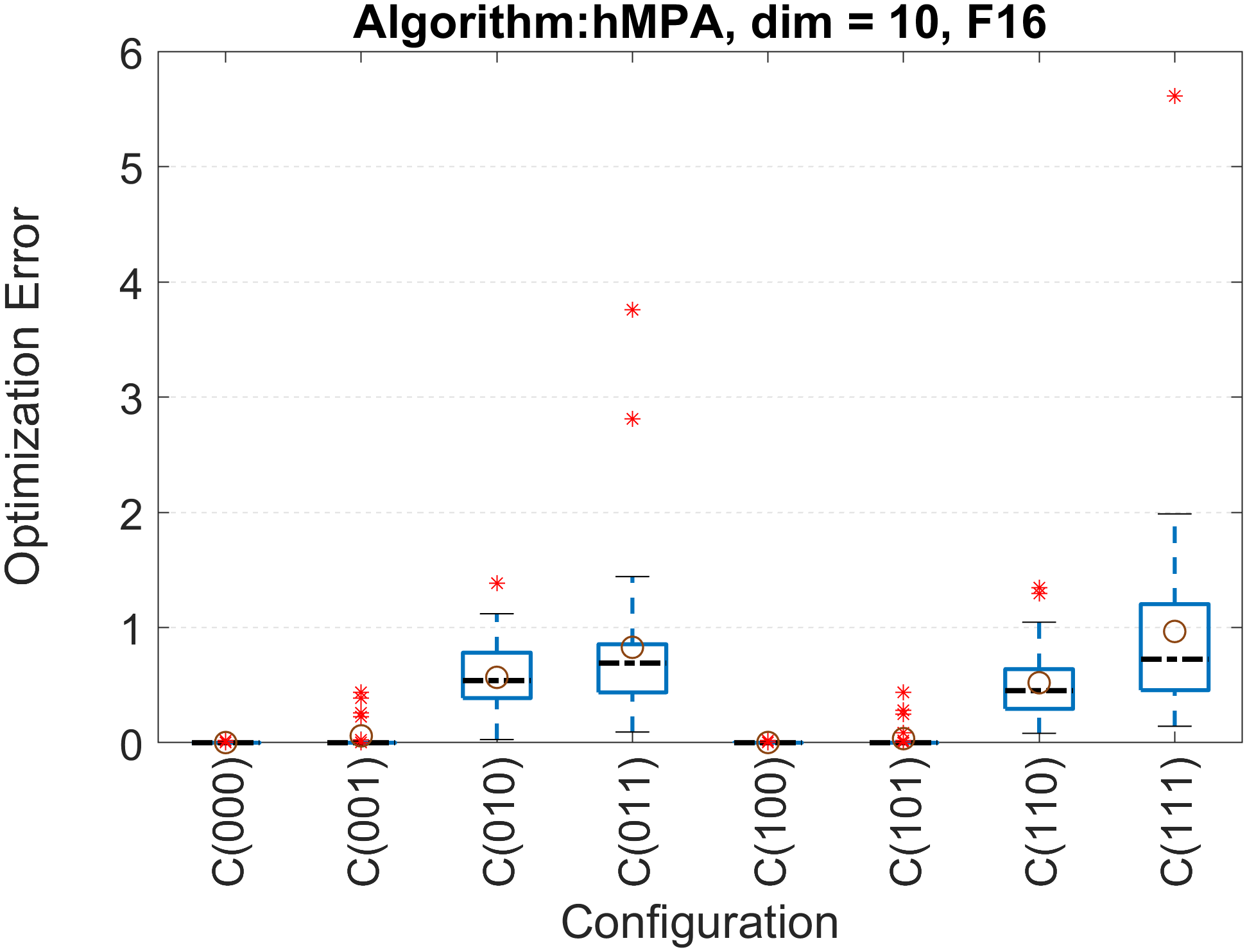}
\end{subfigure}
\hfill
\begin{subfigure}{0.49\textwidth}
    \centering
    \includegraphics[
        width=\linewidth,
        keepaspectratio
    ]{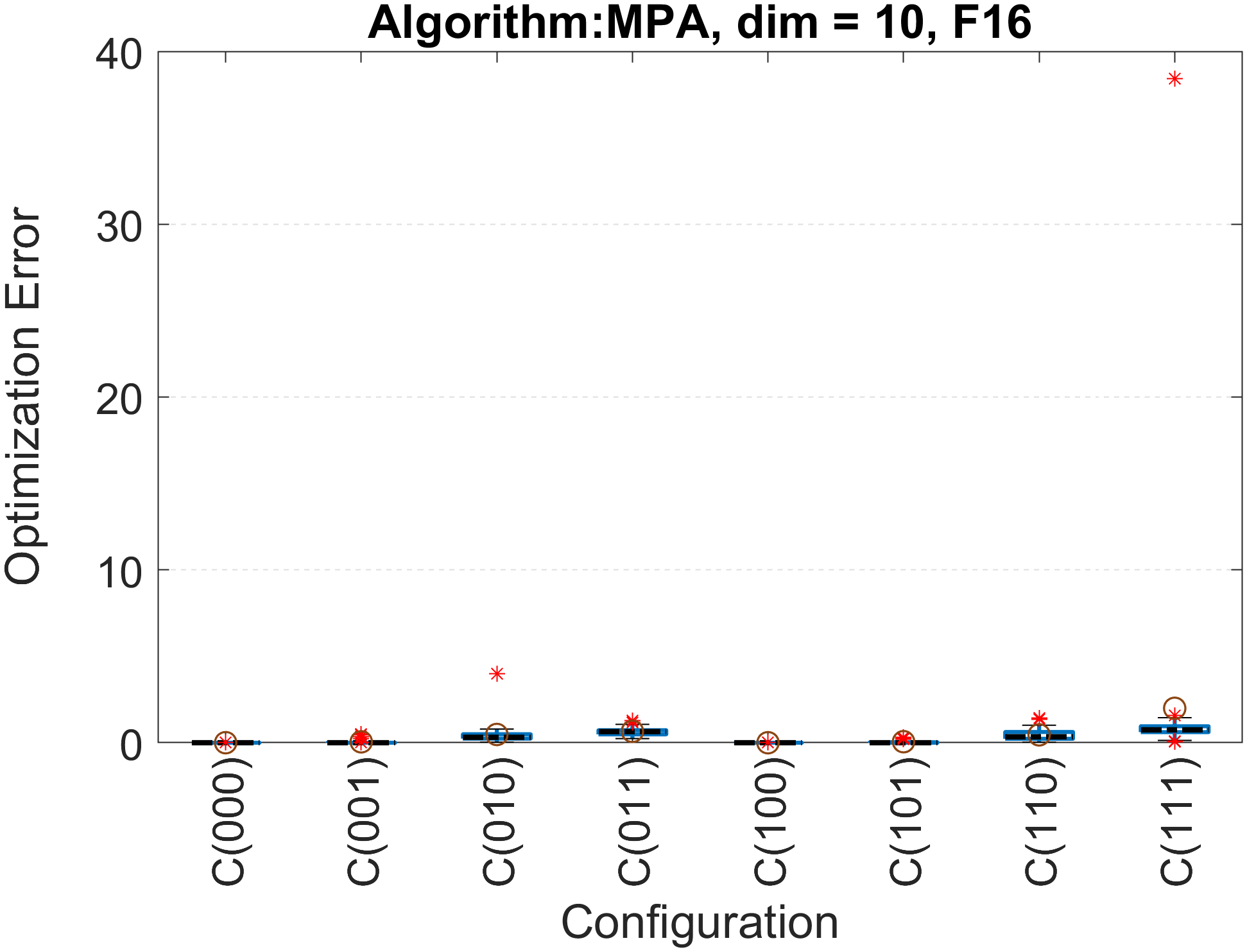}
\end{subfigure}

\end{figure}

\begin{figure}[ht]
\centering

\begin{subfigure}{0.49\textwidth}
    \centering
    \includegraphics[
        width=\linewidth,
        keepaspectratio
    ]{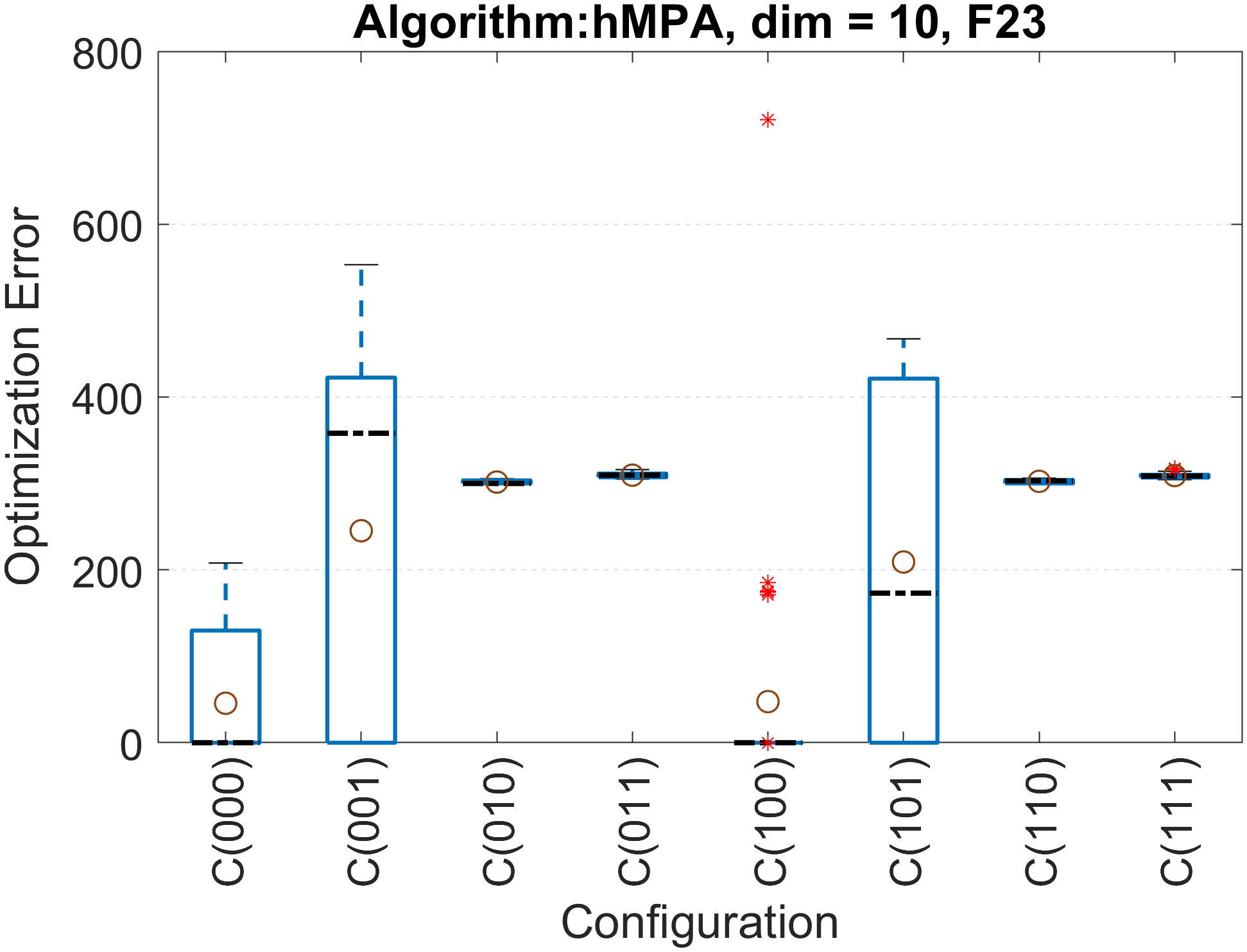}
\end{subfigure}
\hfill
\begin{subfigure}{0.49\textwidth}
    \centering
    \includegraphics[
        width=\linewidth,
        keepaspectratio
    ]{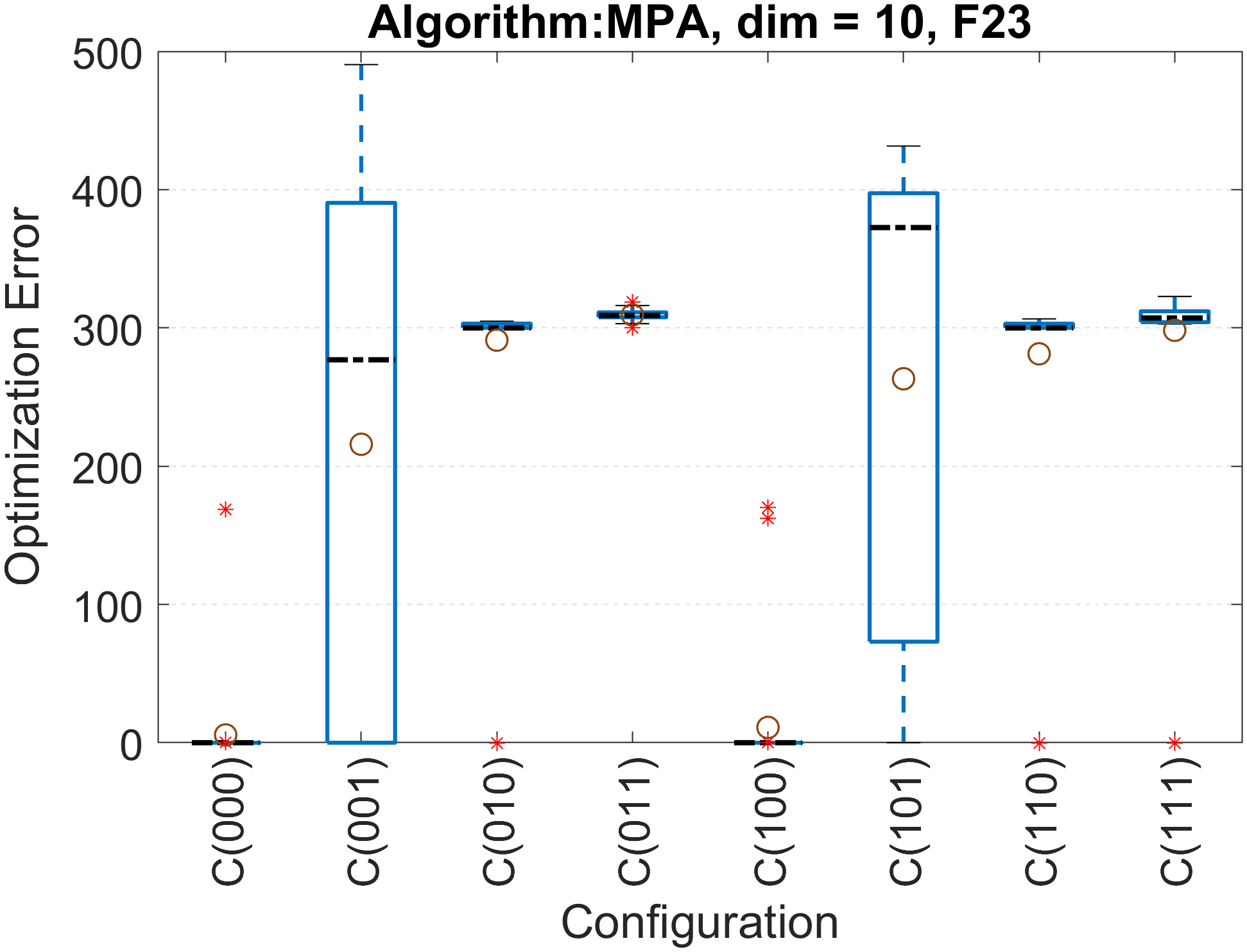}
\end{subfigure}

\end{figure}


\begin{figure}[ht]
\centering

\begin{subfigure}{0.49\textwidth}
    \centering
    \includegraphics[
        width=\linewidth,
        keepaspectratio
    ]{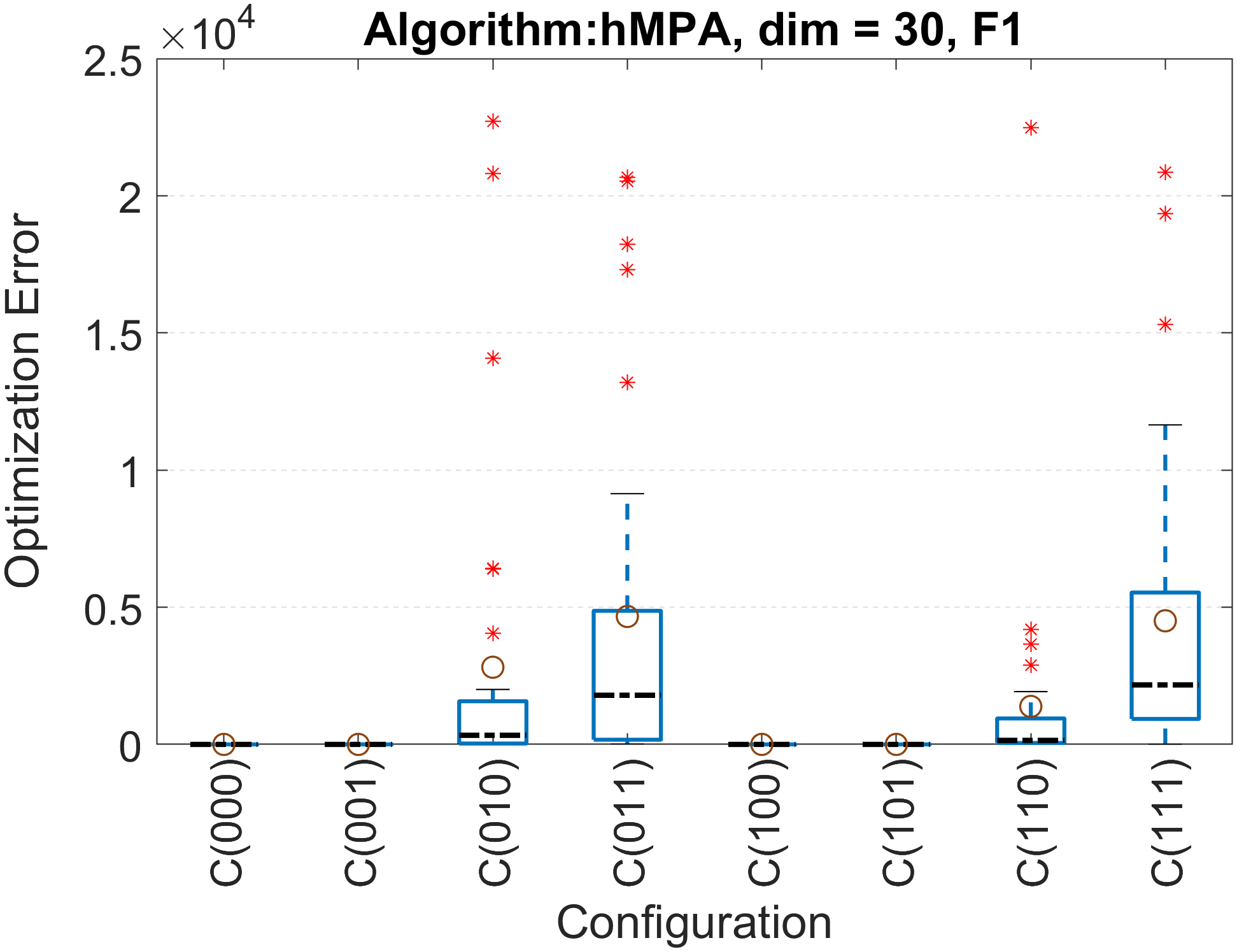}
\end{subfigure}
\hfill
\begin{subfigure}{0.49\textwidth}
    \centering
    \includegraphics[
        width=\linewidth,
        keepaspectratio
    ]{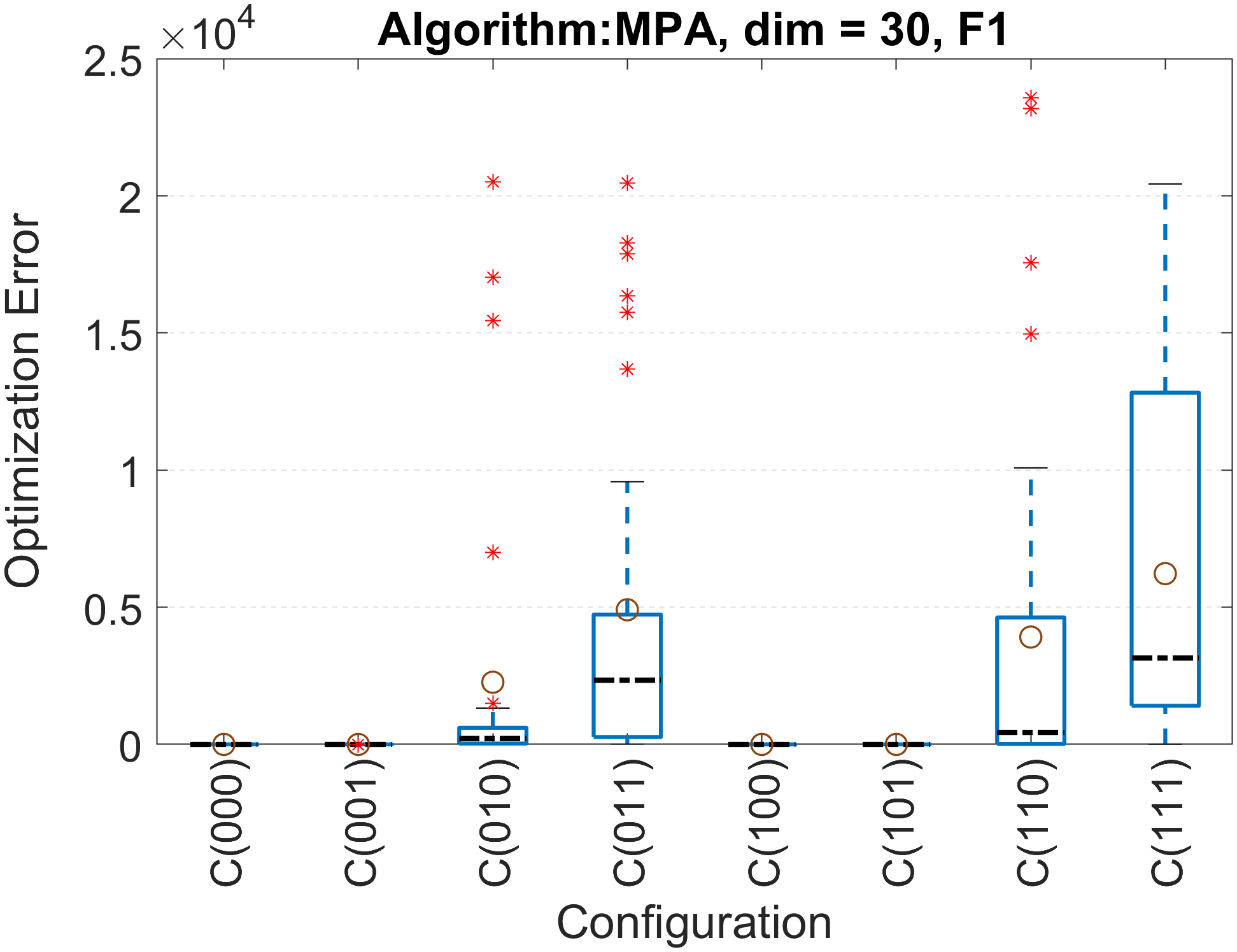}
\end{subfigure}

\end{figure}

\begin{figure}[ht]
\centering
\noindent\textit{Figure~2(continued).}\par\vspace{2pt}
\vspace{0.2cm}
\begin{subfigure}{0.49\textwidth}
    \centering
    \includegraphics[
        width=\linewidth,
        keepaspectratio
    ]{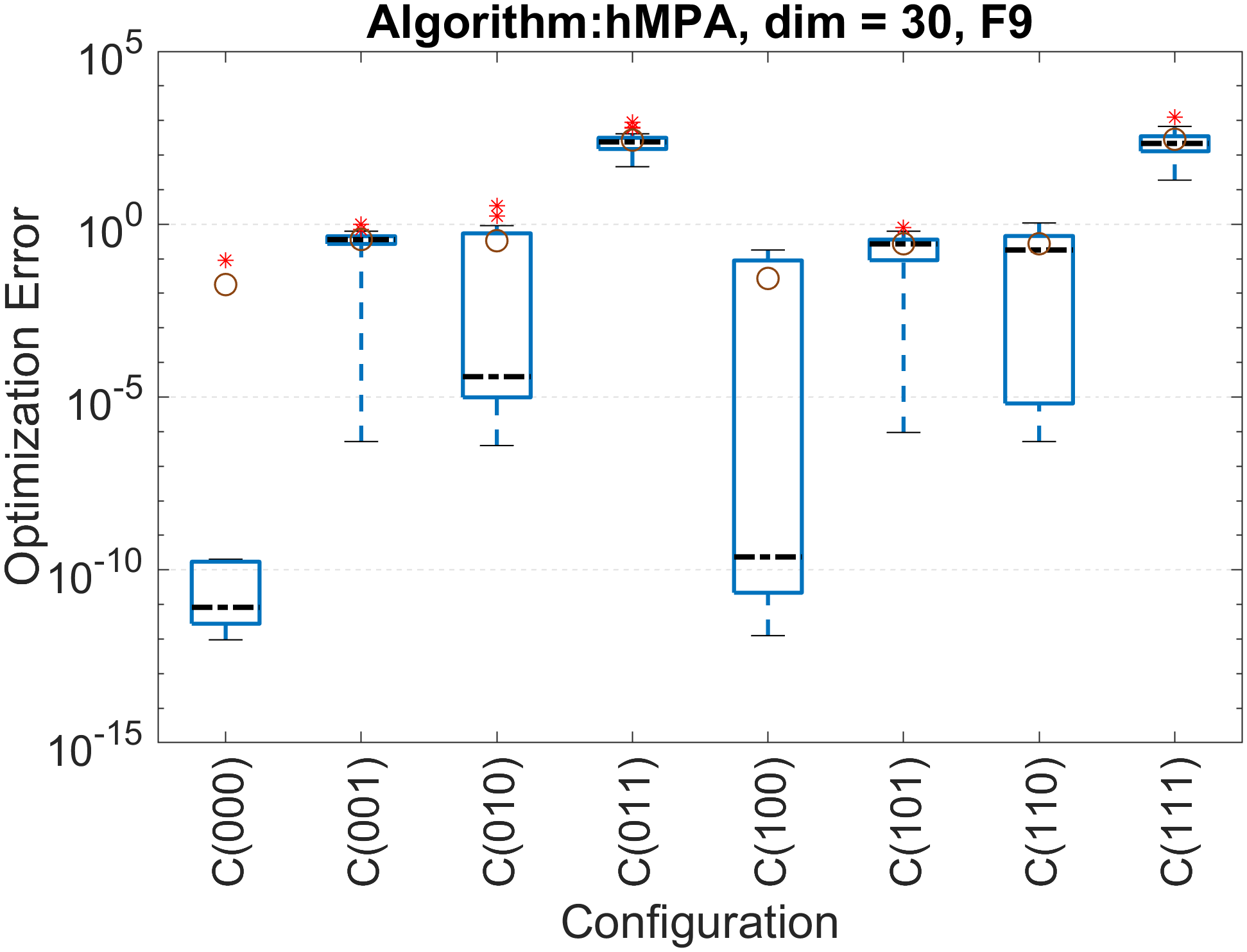}
\end{subfigure}
\hfill
\begin{subfigure}{0.49\textwidth}
    \centering
    \includegraphics[
        width=\linewidth,
        keepaspectratio
    ]{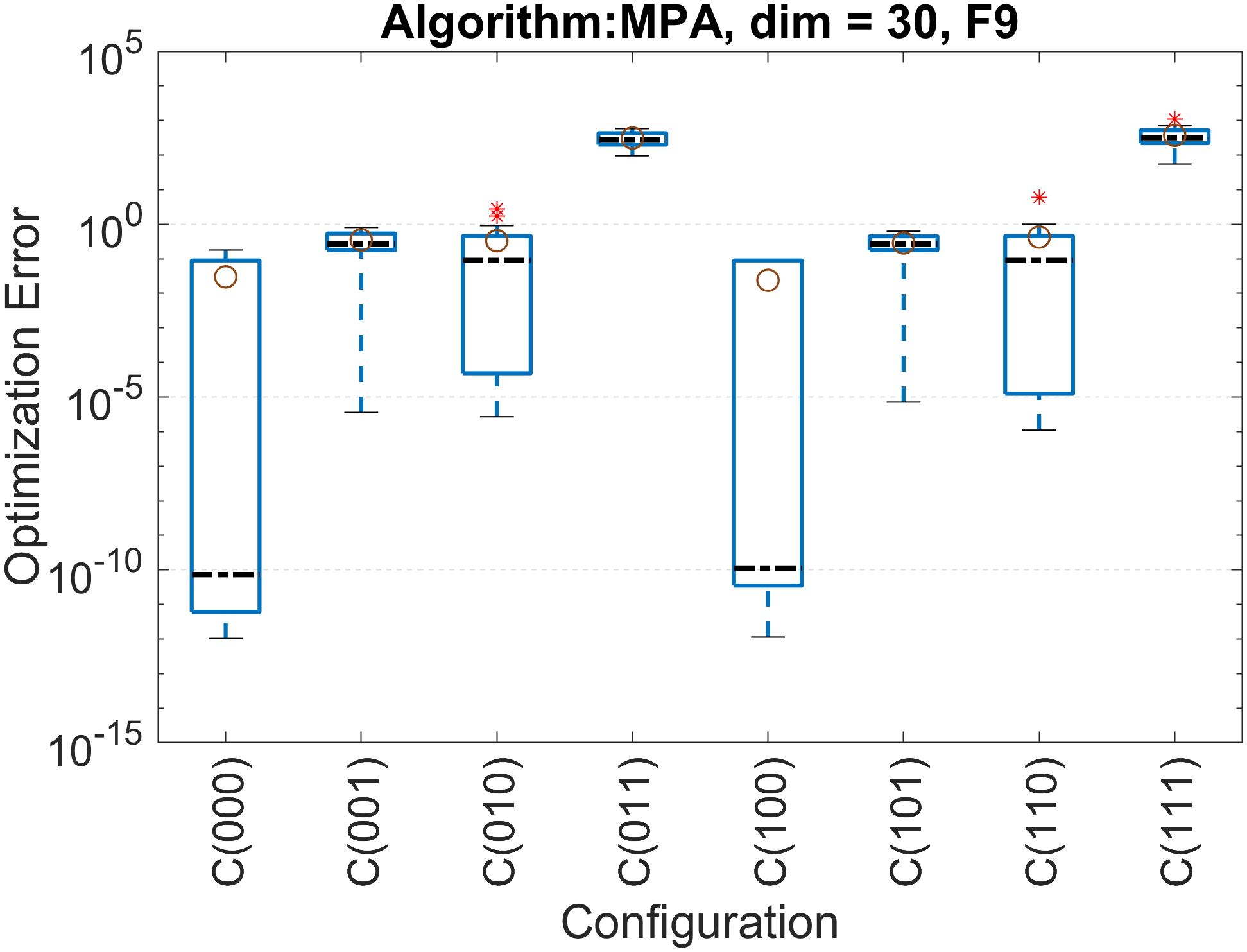}
\end{subfigure}

\end{figure}

\begin{figure}[ht]
\centering

\begin{subfigure}{0.49\textwidth}
    \centering
    \includegraphics[
        width=\linewidth,
        keepaspectratio
    ]{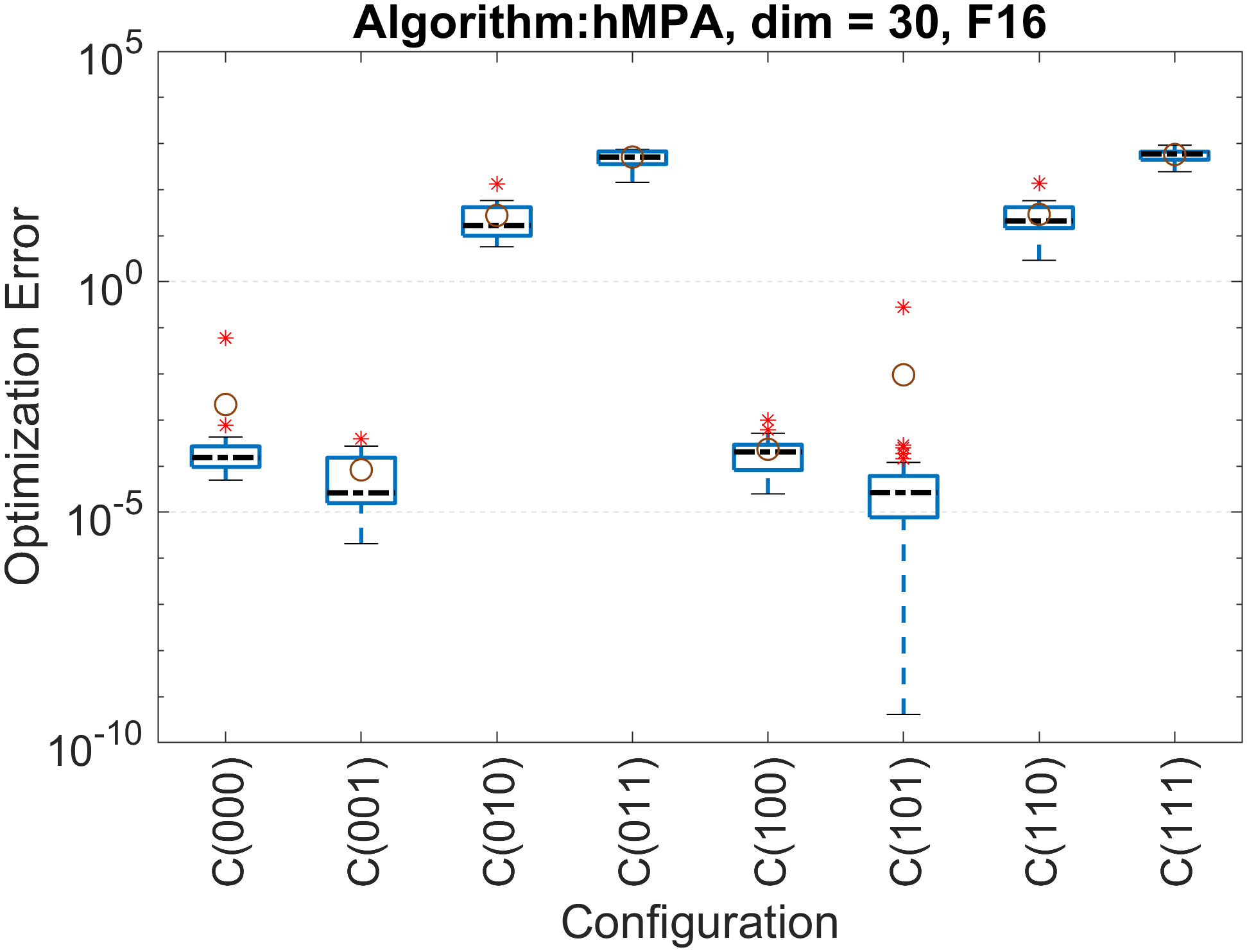}
\end{subfigure}
\hfill
\begin{subfigure}{0.49\textwidth}
    \centering
    \includegraphics[
        width=\linewidth,
        keepaspectratio
    ]{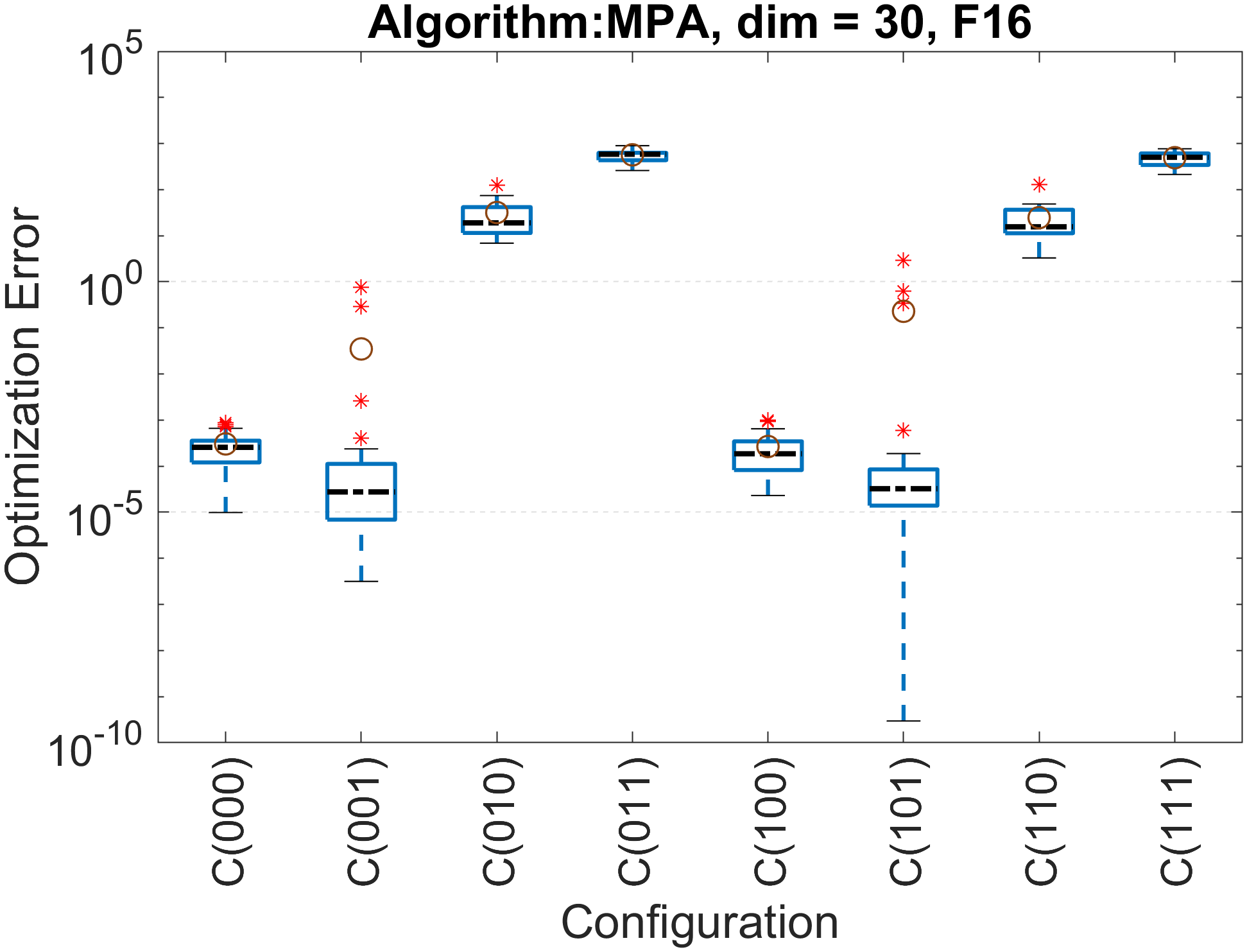}
\end{subfigure}

\end{figure}

\begin{figure}[ht]
\centering

\begin{subfigure}{0.49\textwidth}
    \centering
    \includegraphics[
        width=\linewidth,
        keepaspectratio
    ]{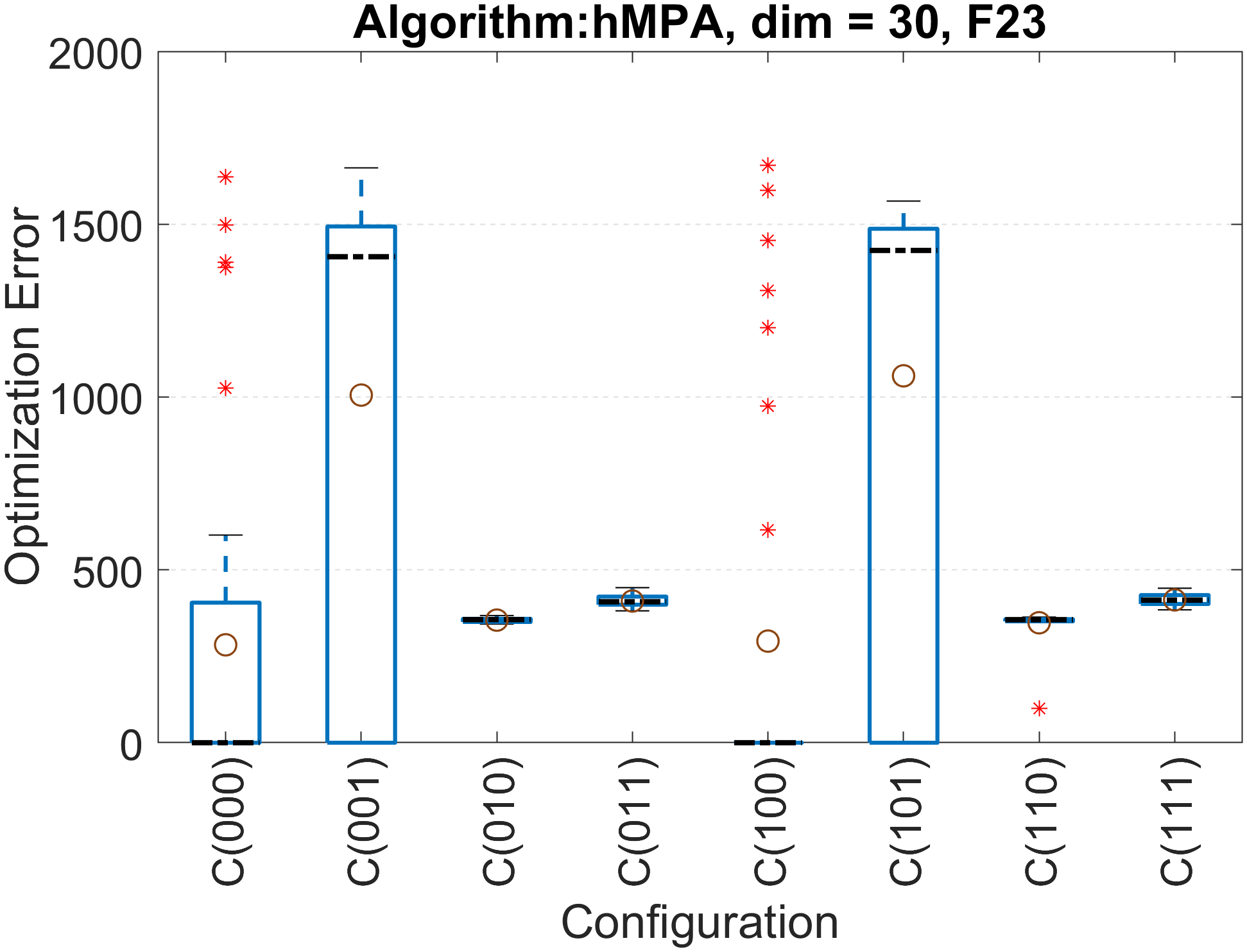}
\end{subfigure}
\hfill
\begin{subfigure}{0.49\textwidth}
    \centering
    \includegraphics[
        width=\linewidth,
        keepaspectratio
    ]{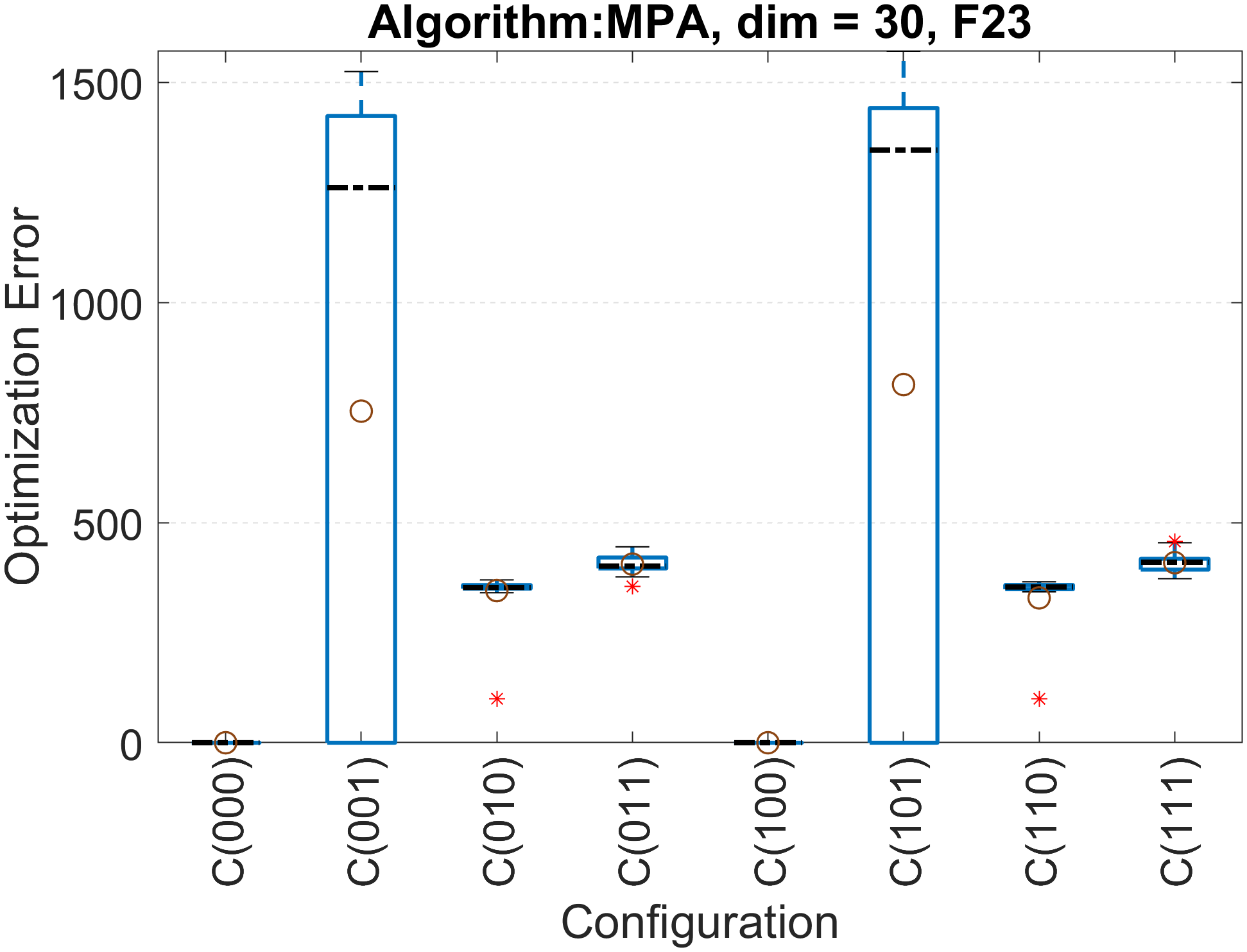}
\end{subfigure}

\end{figure}


\begin{figure}[ht]
\centering

\begin{subfigure}{0.49\textwidth}
    \centering
    \includegraphics[
        width=\linewidth,
        keepaspectratio
    ]{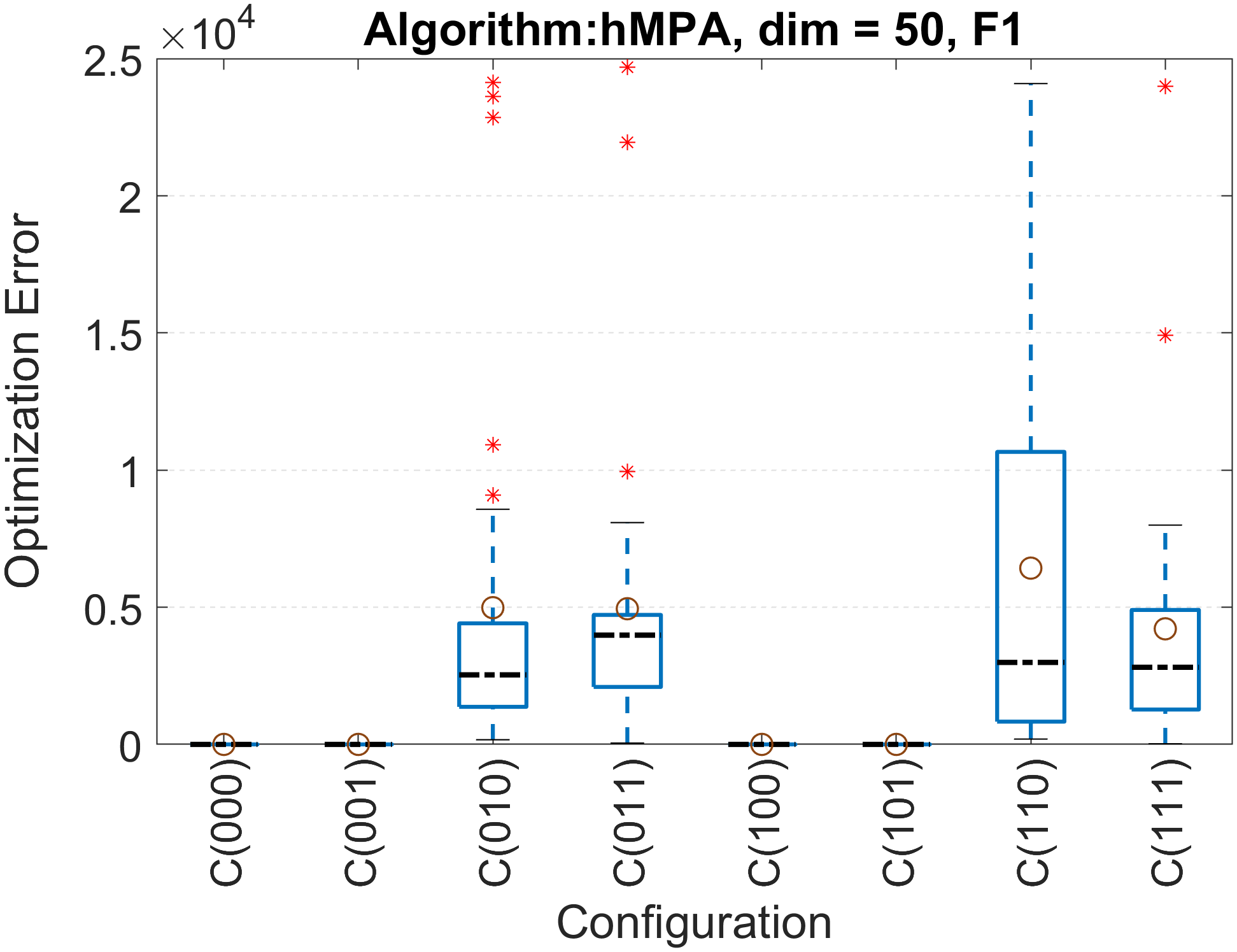}
\end{subfigure}
\hfill
\begin{subfigure}{0.49\textwidth}
    \centering
    \includegraphics[
        width=\linewidth,
        keepaspectratio
    ]{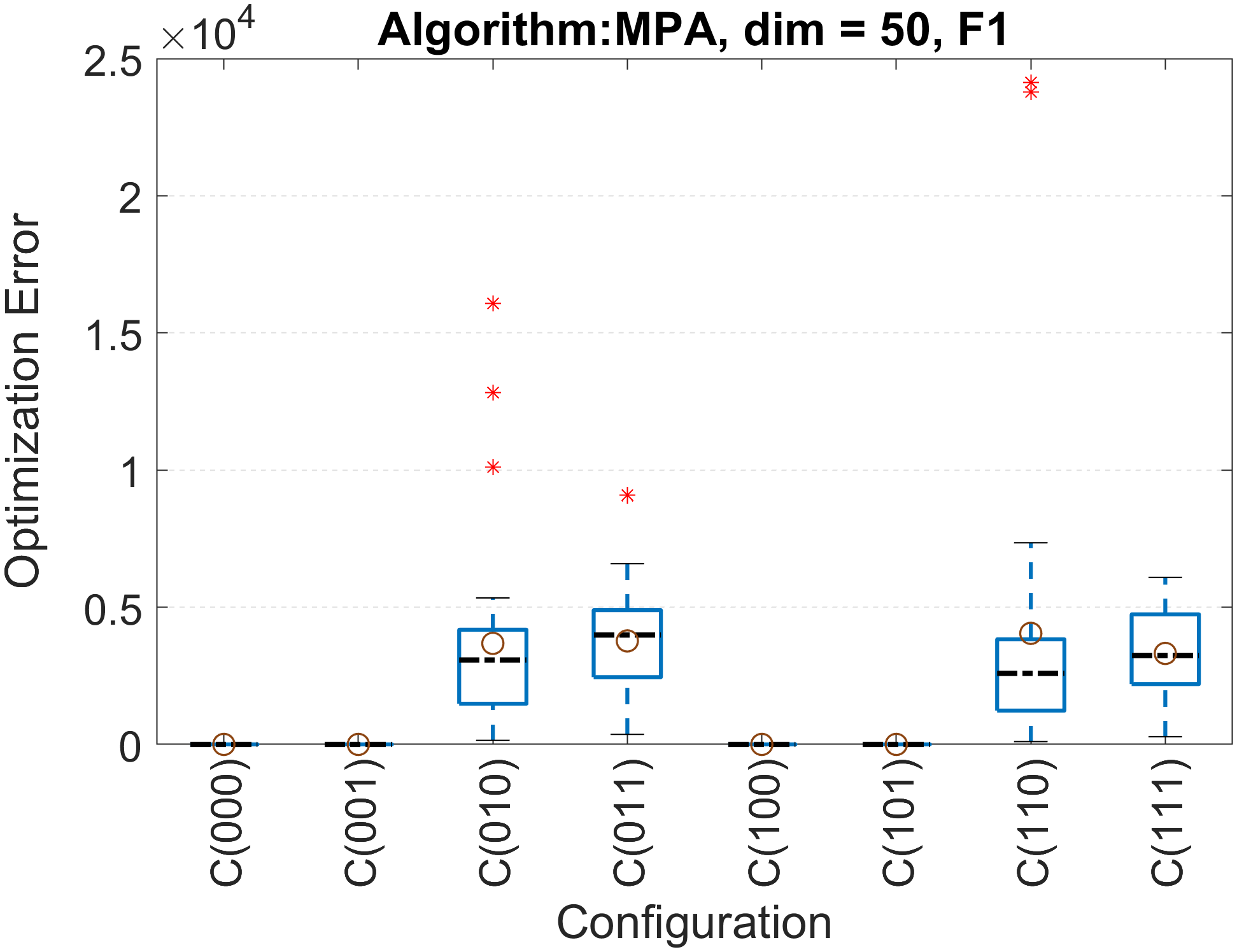}
\end{subfigure}

\end{figure}

\begin{figure}[ht]
\centering
\noindent\textit{Figure~2(continued).}\par\vspace{2pt}
\vspace{0.2cm}
\begin{subfigure}{0.49\textwidth}
    \centering
    \includegraphics[
        width=\linewidth,
        keepaspectratio
    ]{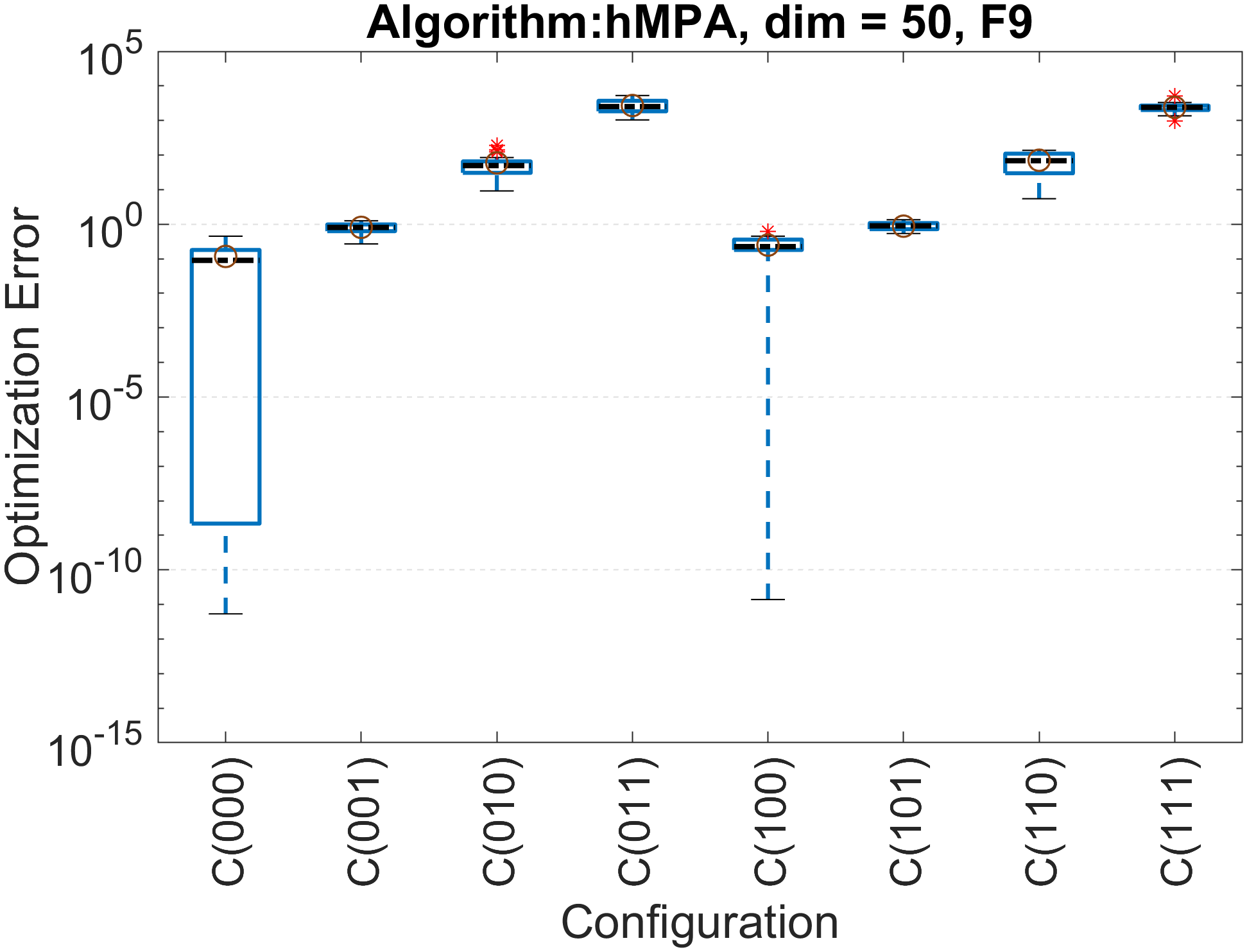}
\end{subfigure}
\hfill
\begin{subfigure}{0.49\textwidth}
    \centering
    \includegraphics[
        width=\linewidth,
        keepaspectratio
    ]{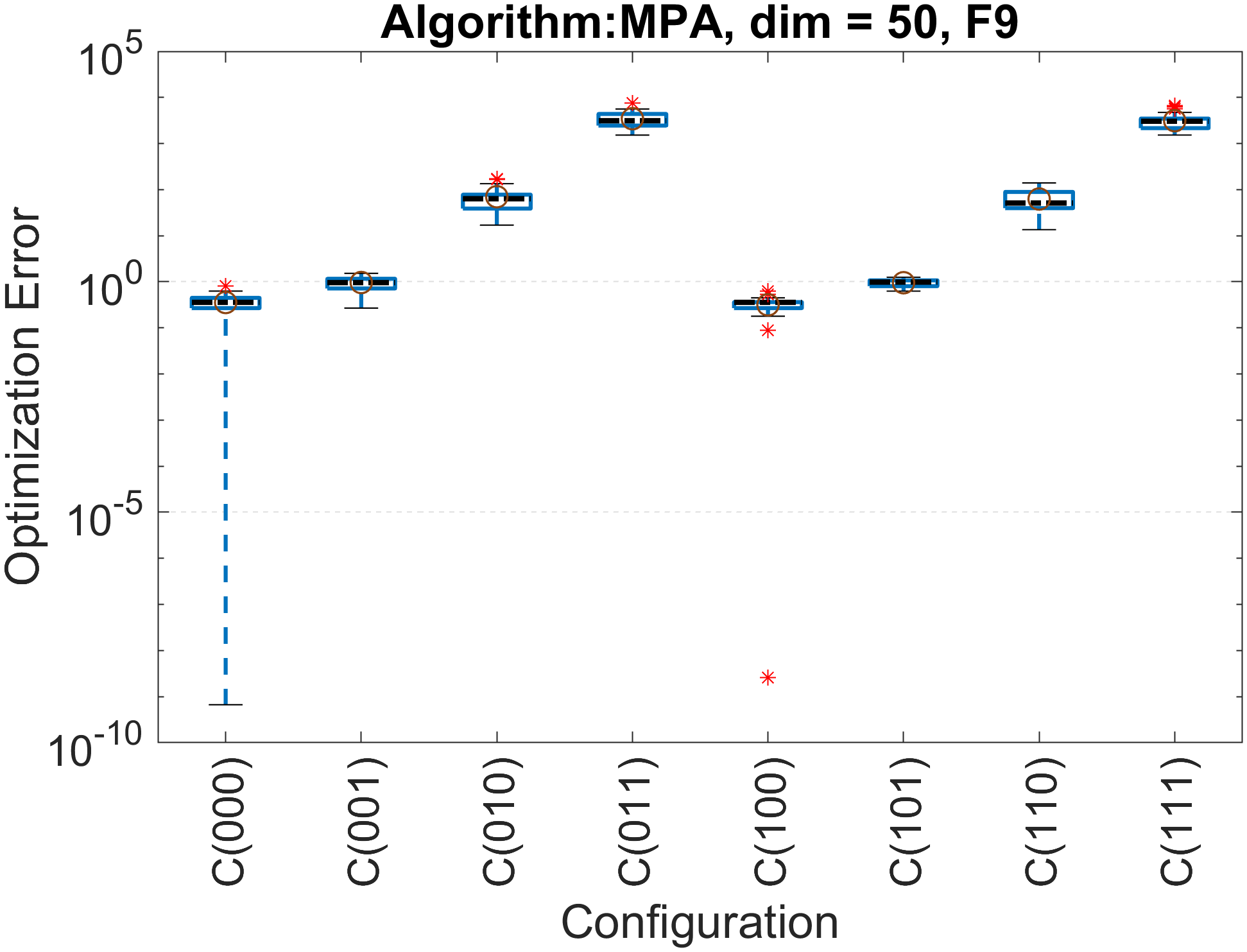}
\end{subfigure}

\end{figure}

\begin{figure}[ht]
\centering

\begin{subfigure}{0.49\textwidth}
    \centering
    \includegraphics[
        width=\linewidth,
        keepaspectratio
    ]{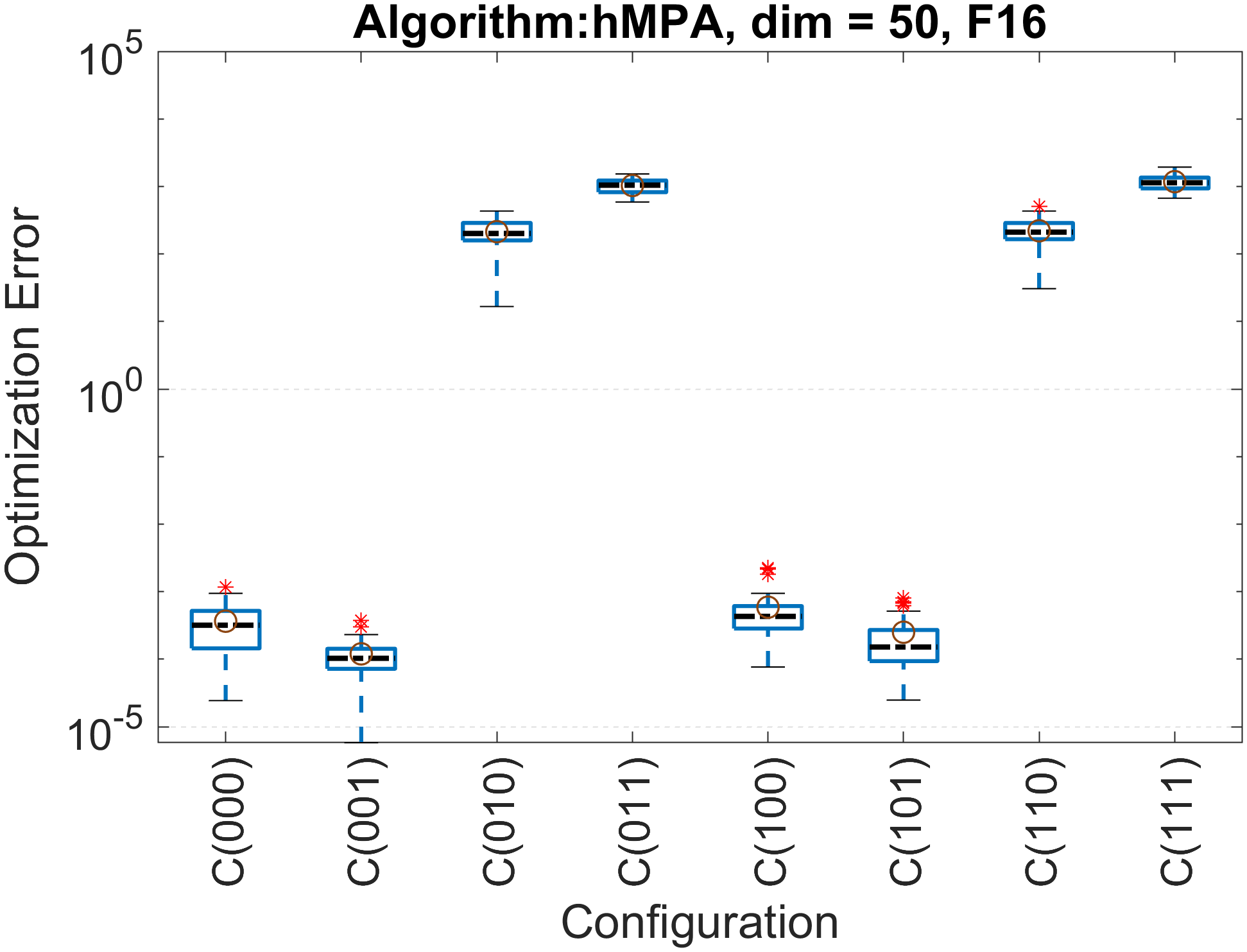}
\end{subfigure}
\hfill
\begin{subfigure}{0.49\textwidth}
    \centering
    \includegraphics[
        width=\linewidth,
        keepaspectratio
    ]{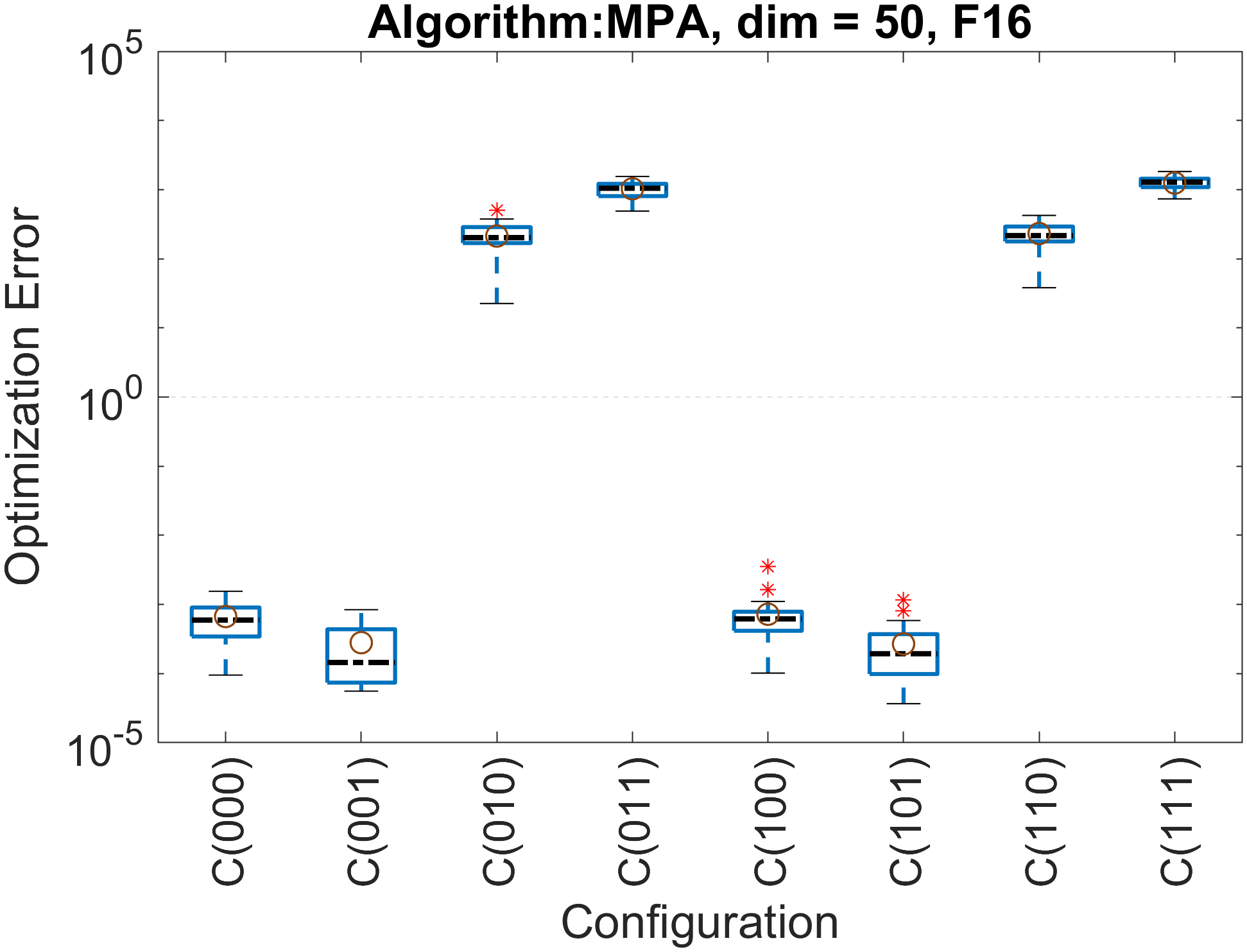}
\end{subfigure}

\end{figure}

\begin{figure}[ht]
\centering

\begin{subfigure}{0.49\textwidth}
    \centering
    \includegraphics[
        width=\linewidth,
        keepaspectratio
    ]{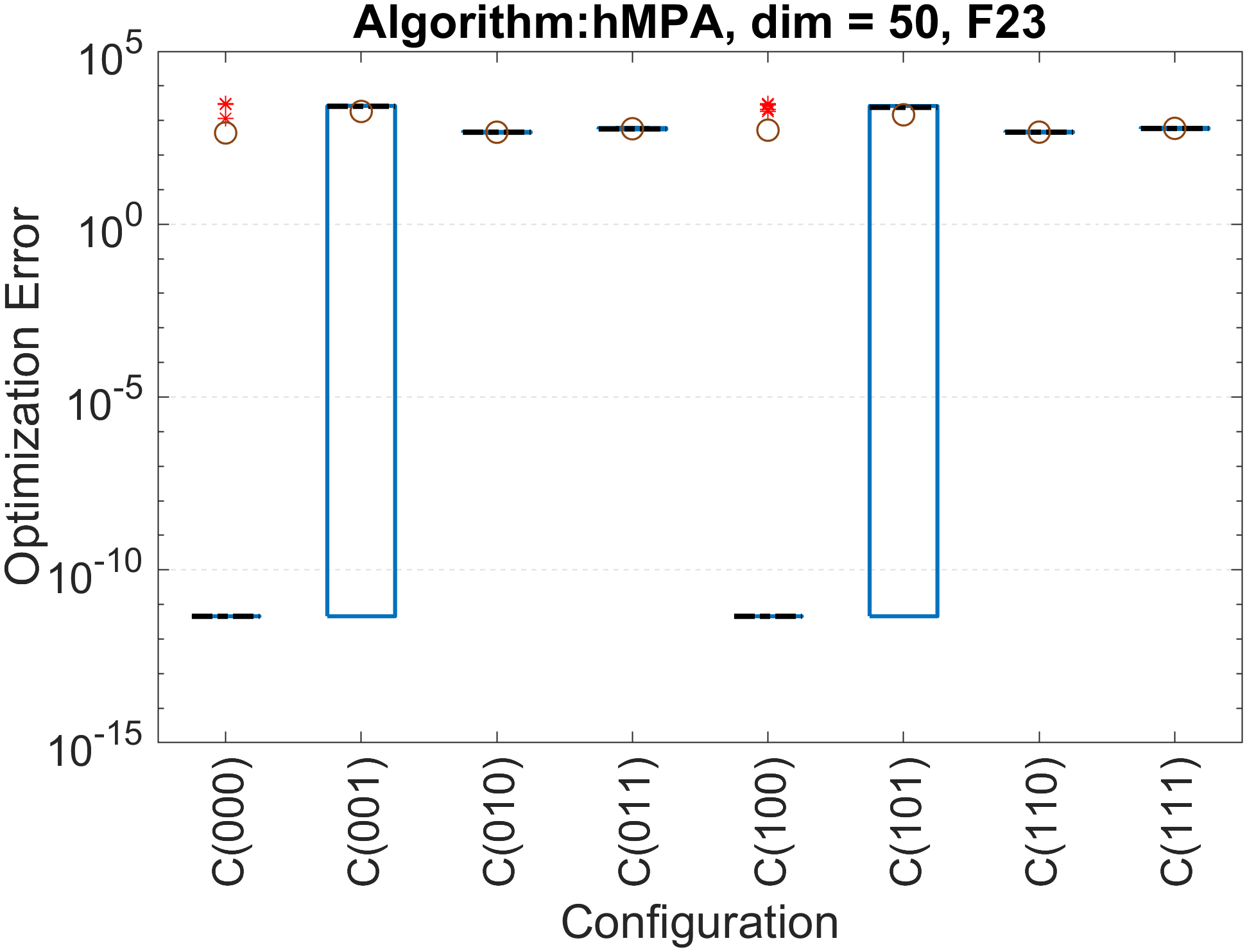}
\end{subfigure}
\hfill
\begin{subfigure}{0.49\textwidth}
    \centering
    \includegraphics[
        width=\linewidth,
        keepaspectratio
    ]{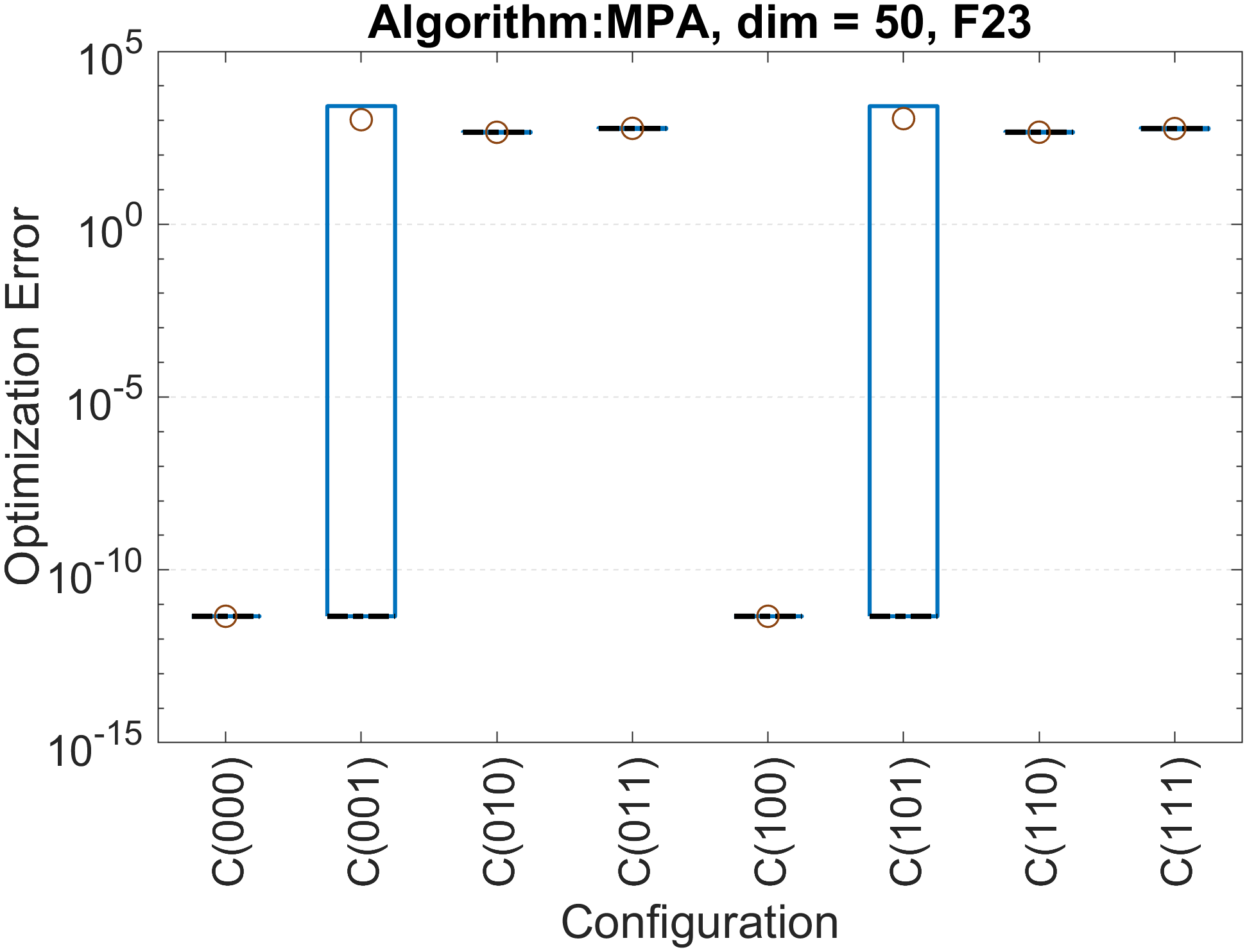}
\end{subfigure}

\end{figure}


\begin{figure}[ht]
\centering

\begin{subfigure}{0.49\textwidth}
    \centering
    \includegraphics[
        width=\linewidth,
        keepaspectratio
    ]{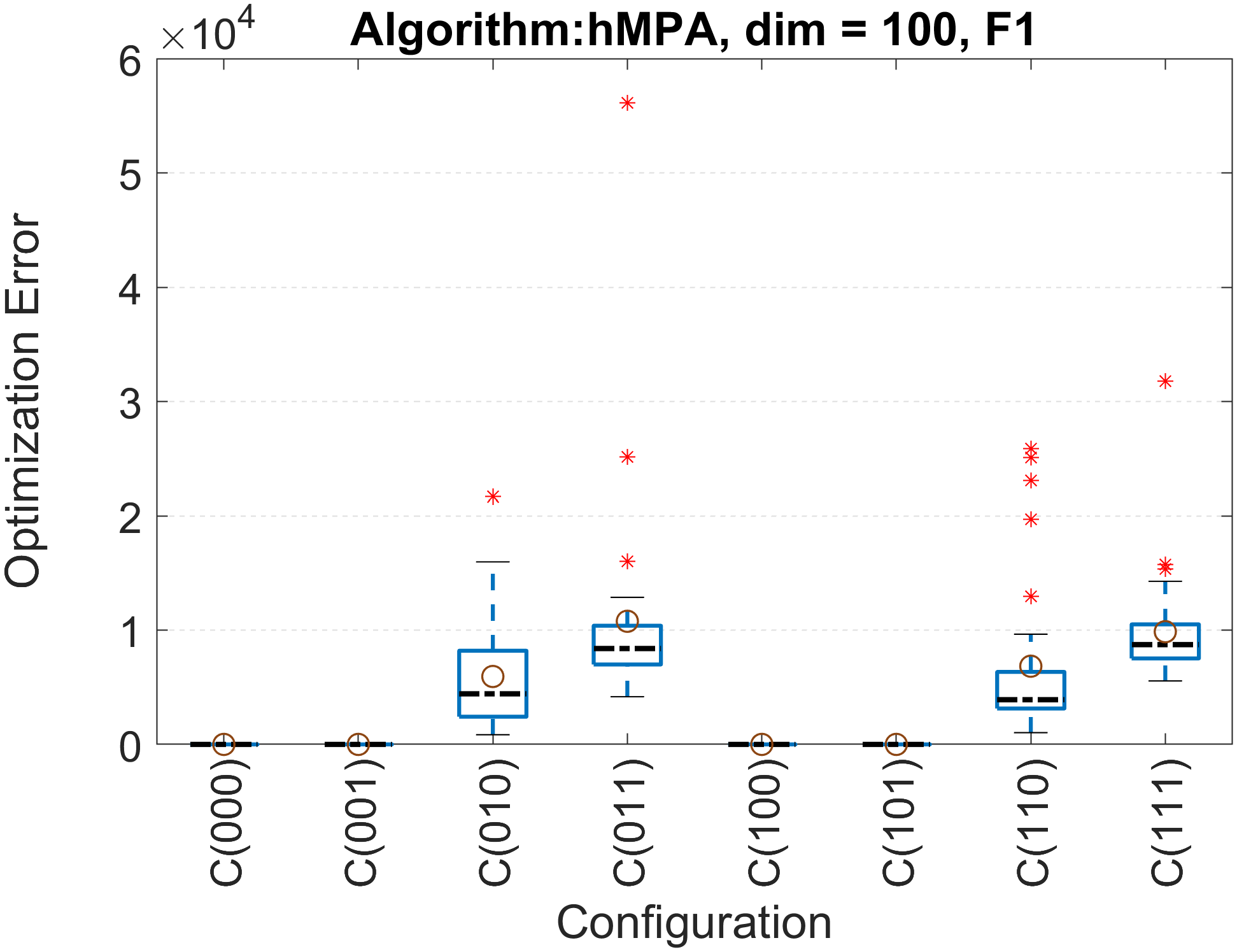}
\end{subfigure}
\hfill
\begin{subfigure}{0.49\textwidth}
    \centering
    \includegraphics[
        width=\linewidth,
        keepaspectratio
    ]{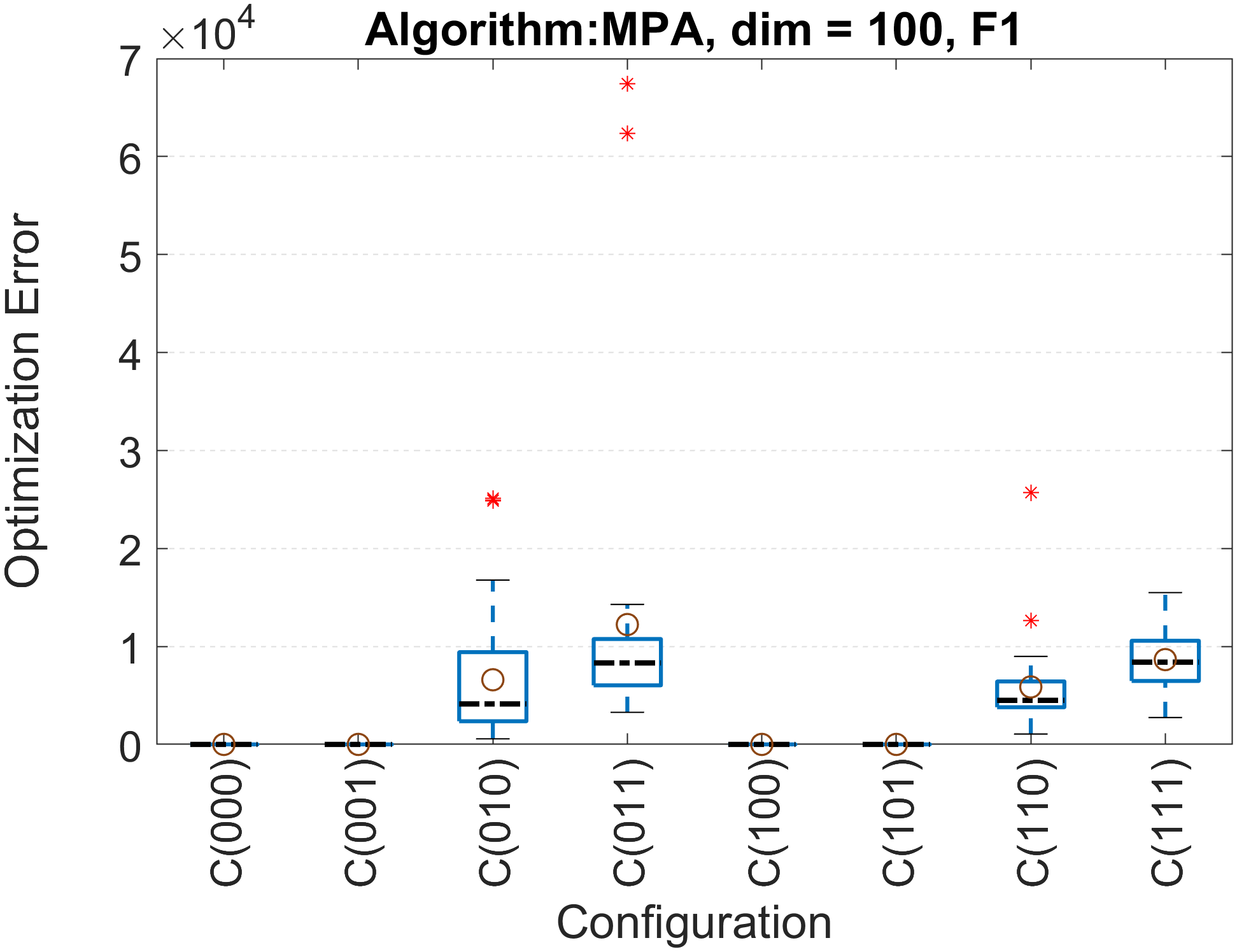}
\end{subfigure}

\end{figure}

\begin{figure}[ht]
\centering

\begin{subfigure}{0.49\textwidth}
    \centering
    \includegraphics[
        width=\linewidth,
        keepaspectratio
    ]{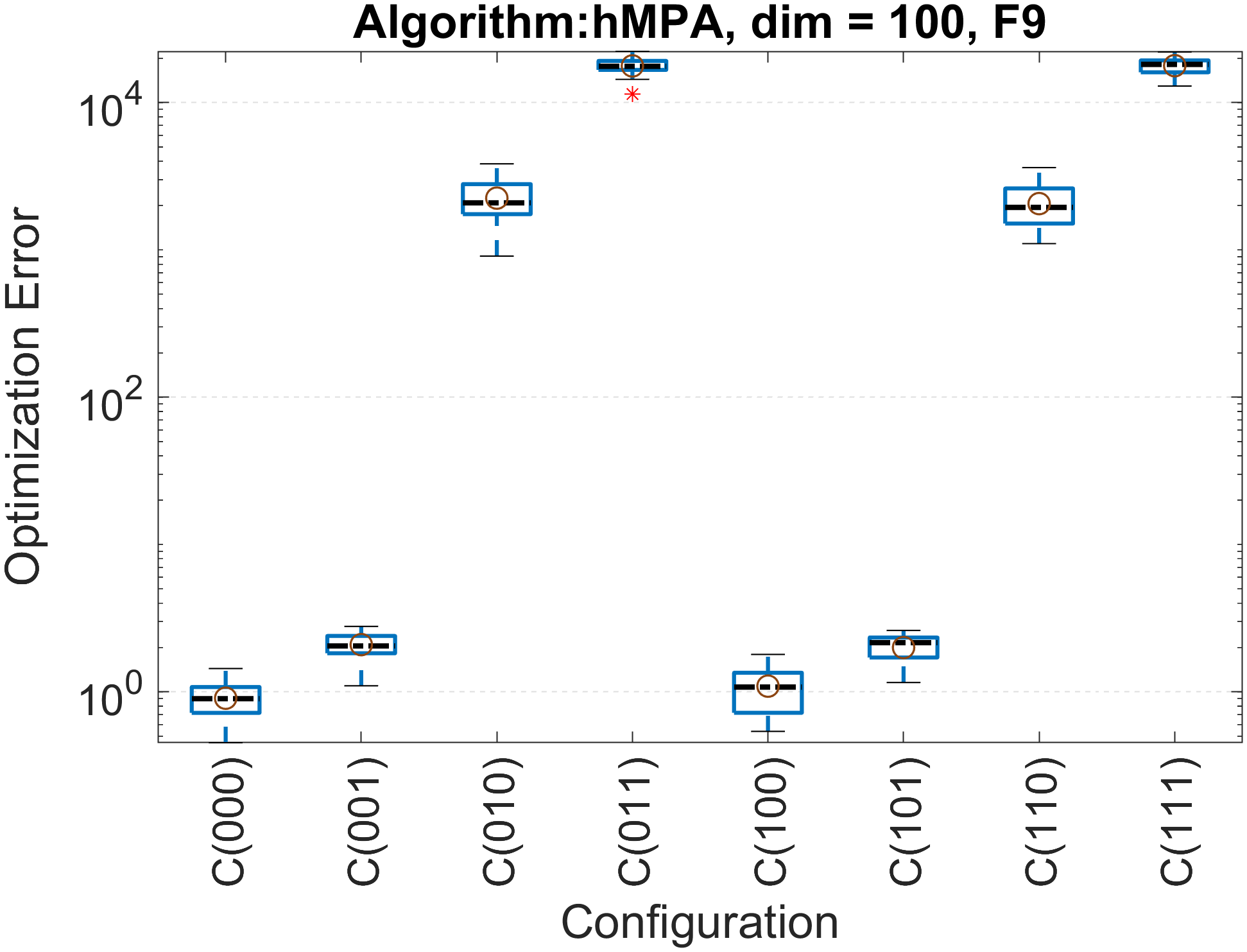}
\end{subfigure}
\hfill
\begin{subfigure}{0.49\textwidth}
    \centering
    \includegraphics[
        width=\linewidth,
        keepaspectratio
    ]{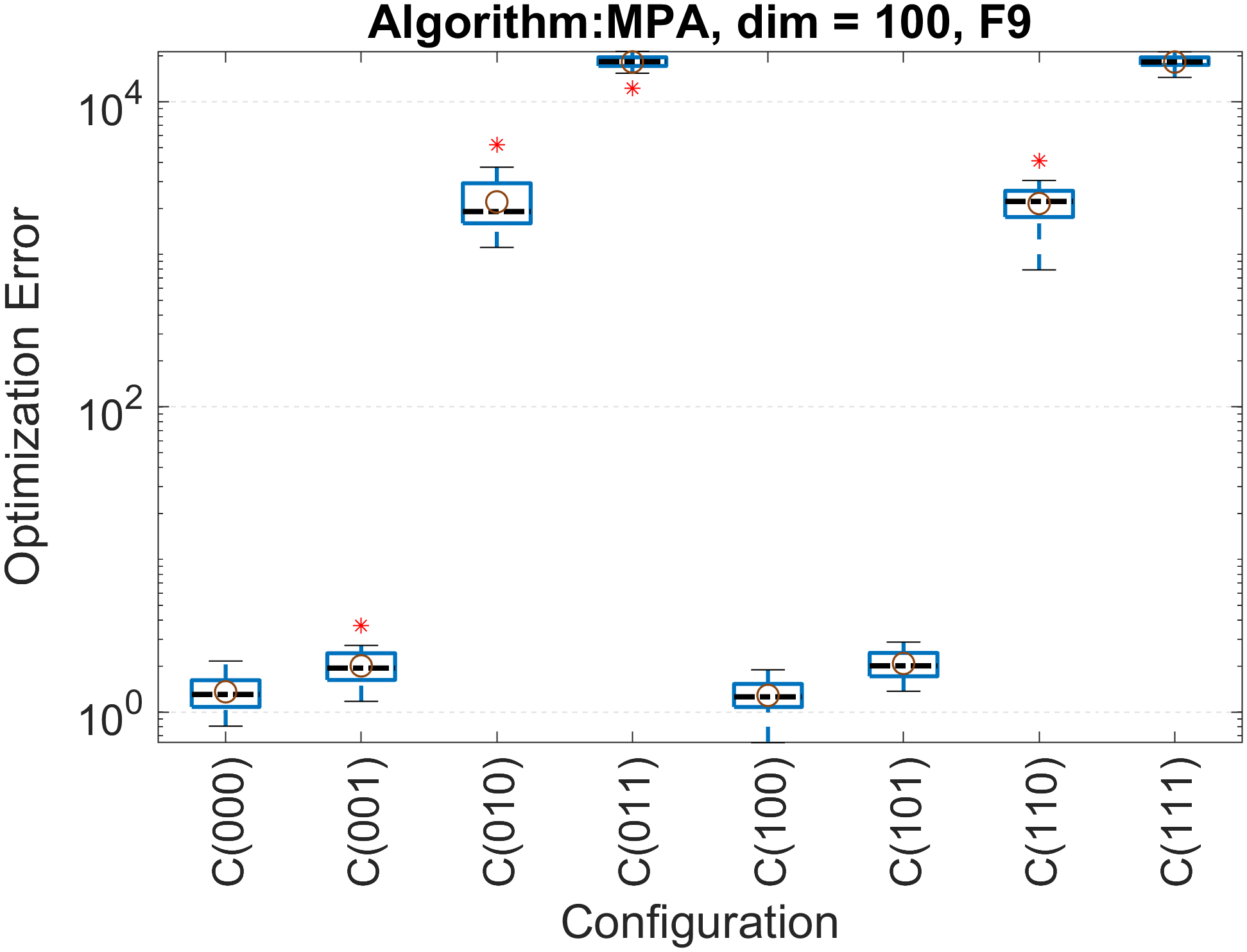}
\end{subfigure}

\end{figure}

\begin{figure}[ht]
\centering

\begin{subfigure}{0.49\textwidth}
    \centering
    \includegraphics[
        width=\linewidth,
        keepaspectratio
    ]{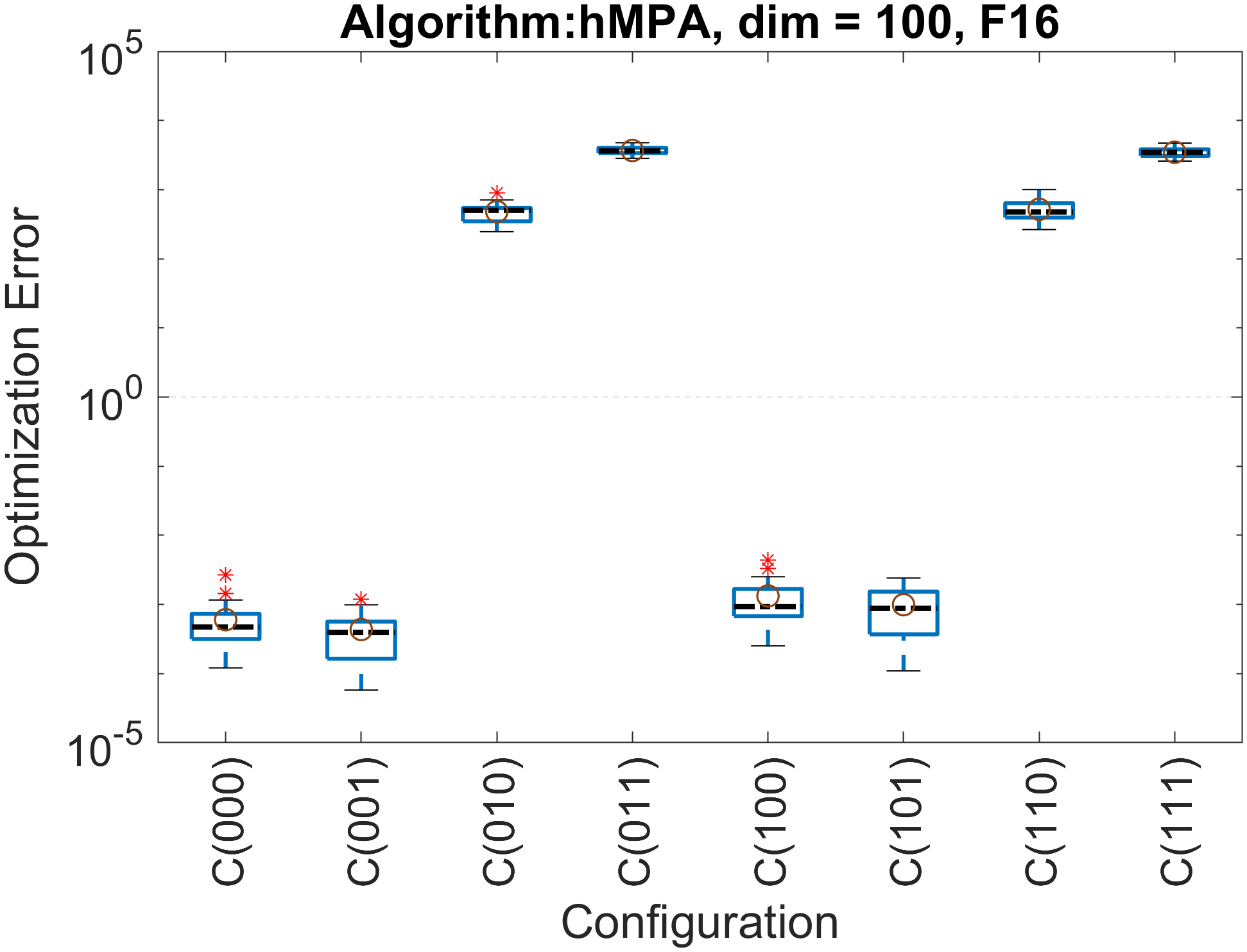}
\end{subfigure}
\hfill
\begin{subfigure}{0.49\textwidth}
    \centering
    \includegraphics[
        width=\linewidth,
        keepaspectratio
    ]{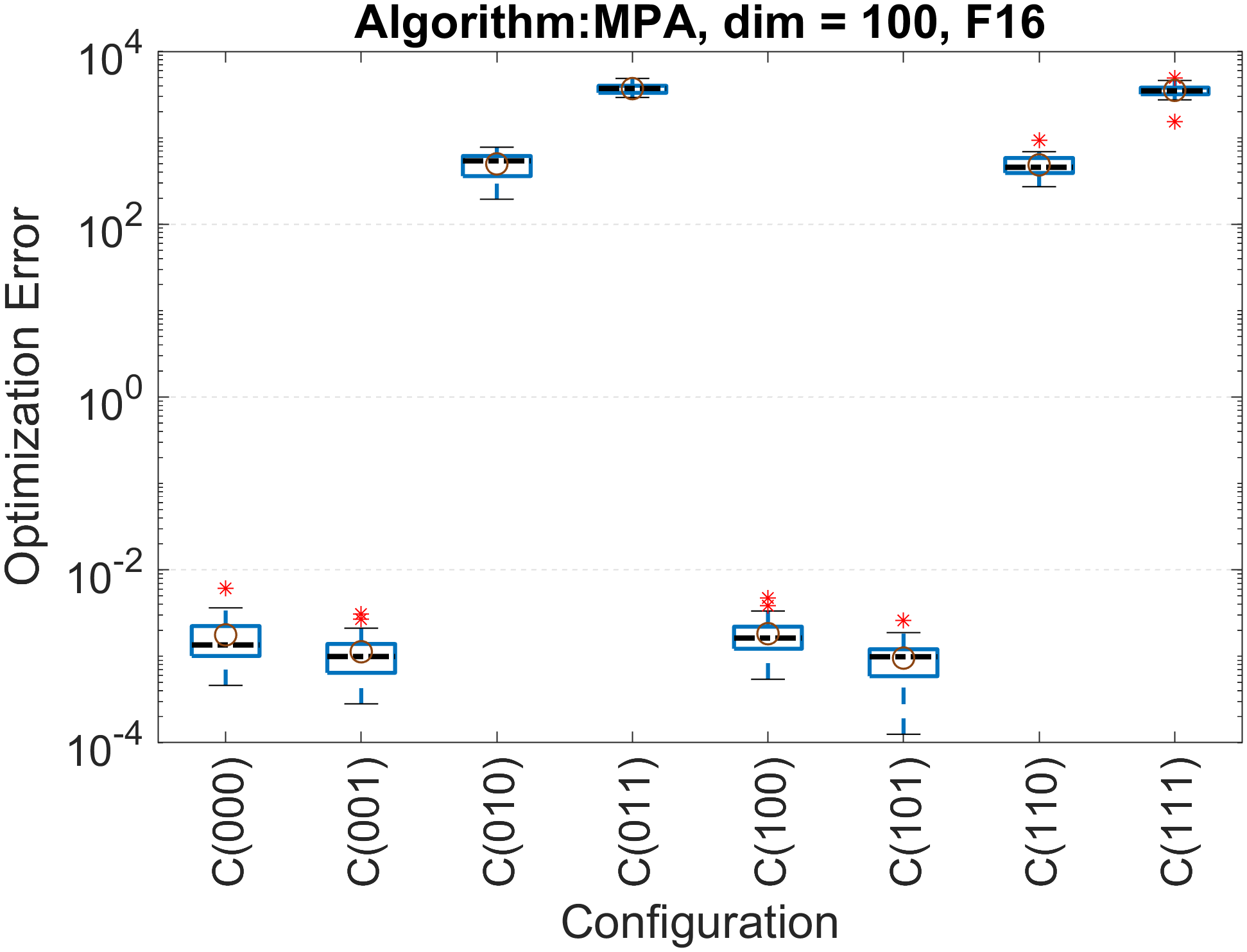}
\end{subfigure}

\end{figure}

\begin{figure}[ht]
\centering

\begin{subfigure}{0.49\textwidth}
    \centering
    \includegraphics[
        width=\linewidth,
        keepaspectratio
    ]{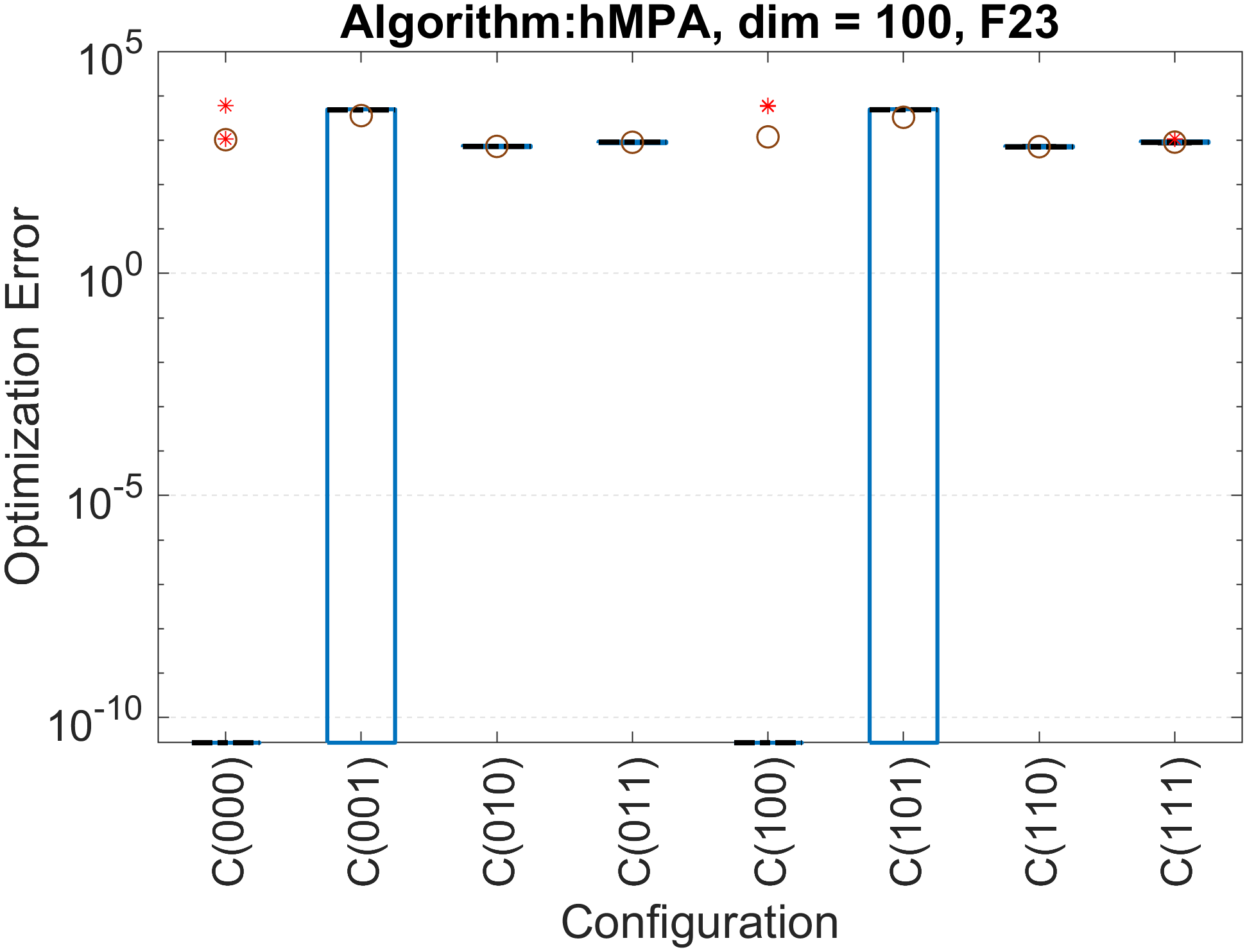}
\end{subfigure}
\hfill
\begin{subfigure}{0.49\textwidth}
    \centering
    \includegraphics[
        width=\linewidth,
        keepaspectratio
    ]{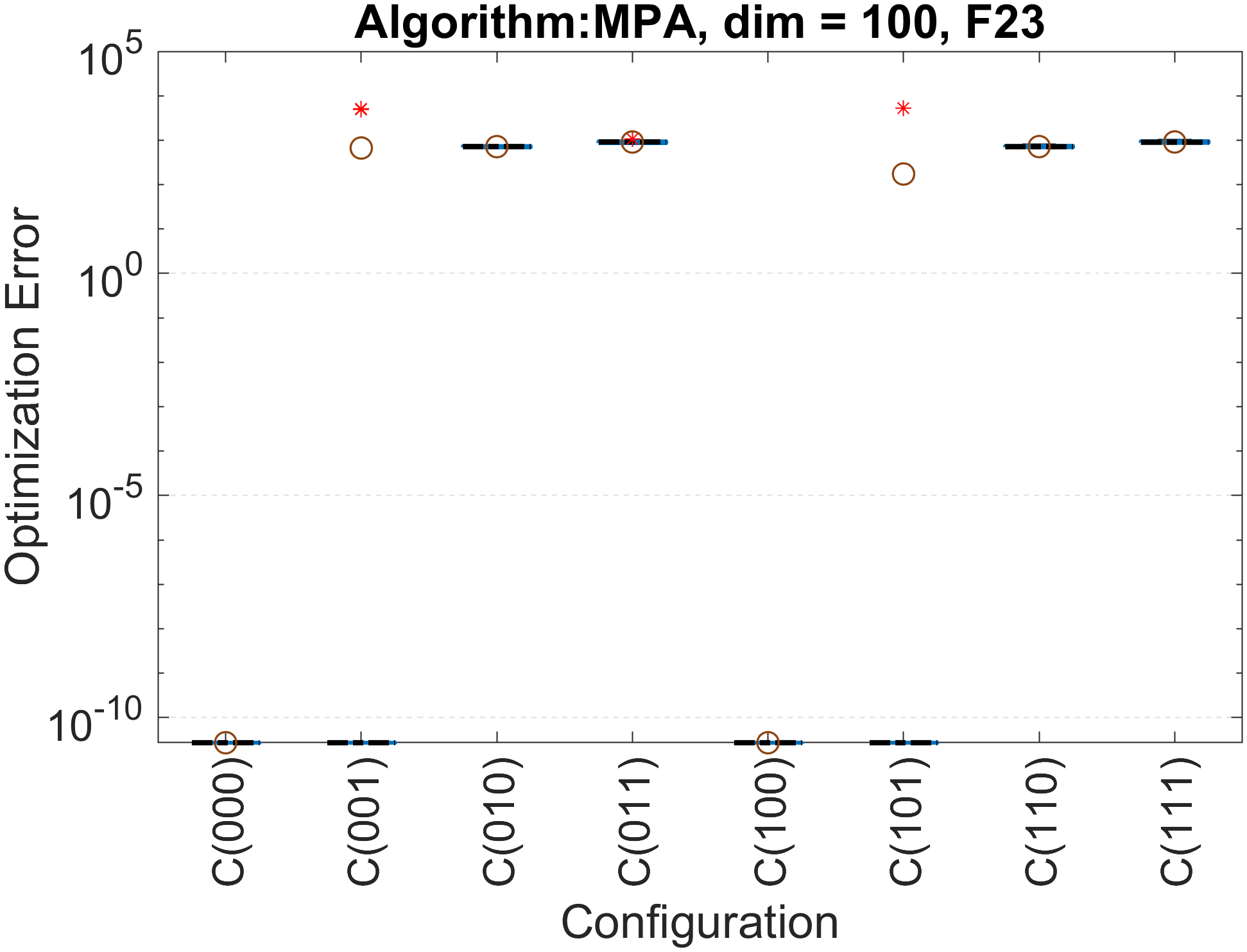}
\end{subfigure}

\vspace{0.5cm}

\includegraphics[
    width=0.5\textwidth,
    keepaspectratio
]{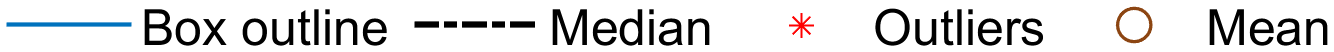}\\
\vspace{0.5cm}
Figure 2. Boxplots of final optimization errors for modified CEC2017 functions for $dim\in\{10, \ 30, \ 50, \ 100\}.$
\label{fig2}
\end{figure}
\clearpage
The largest spreads at the 3-largest dimensions (i.e. $dim\in\{30,50,100\}$) are exhibited by $C(010)$, $C(011)$, $C(110)$ and $C(111)$ (for function $F1$). At these dimensions, hMPA achieves its lowest $F9$ medians with $C(000)$. Conversely, MPA achieves its lowest medians with $C(000)$ at $dim = 30$ and with $C(100)$ at $dim\in\{50, \ 100\}$. The configuration $C(001)$ yields some of the lowest $F16$ medians for both algorithms. However, MPA achieves its lowest $F16$ median with $C(101)$ at $dim =100$. Across all dimensions, configurations $C(000)$ and $C(100)$ yield some of the lowest $F23$ medians for the hMPA and MPA  algorithms. For hMPA under $C(000)$, the $F23$ means exceed the medians. Upper outliers occur at $dim\in\{30, \ 50, \ 100\}$. Section~8 discusses these observations.
 
\vspace{-0.2cm}
\section{Discussion}
\label{8}
\vspace{-0.15cm}

Configuration sensitivity is already present in MPA, Faramarzi et al. in \cite{AF2020}, whereas the consequences of hybridization depend on the problem. By comparing the same functions, configurations, dimensions and evaluation budgets, it is possible to distinguish patterns shared by both methods from the benefits of adding the $h$-operator (Oszust et al., in \cite{MO2021}). Accuracy was assessed using errors relative to the optimum specific to each configuration. Eliminating the additive offset prevents the function level from being confused with solution quality. However, this does not eliminate the possible influence of bias on the optimization process itself.

Global Friedman tests confirm differences in configurations for both algorithms and evaluation schemes (DSC and eDSC) across all tested dimensions (see Table \ref{tab:friedman}). The consistent structure of mean ranks (Tables \ref{tab:dsc_ranks}-\ref{tab:edsc_ranks}) indicates the stability of the main relationship: the four highest-ranked positions are occupied by: $C(000)$, $C(001)$, $C(100)$ and $C(101)$. Differences relative to $C(000)$ for all configurations involving shifts remain significant after Bonferroni correction (Tables: \ref{tab:control_dsc}-\ref{tab:control_edsc}) for every dimension. Analysis of all triplets confirms that grouping them together does not imply an absence of differences within groups (Tables: \ref{tab:triples_dsc}-\ref{tab:triples_edsc}).
The different preferences observed for $F27$ and $F29$ restrict how widely this hierarchy can be applied. They do not replace the assessment under the full $C(111)$ setting.

Configuration rankings describe the relative order within each algorithm. The direct comparison between MPA and hMPA was based on \textbf{unstandardised differences in means and medians} i.e., $Eff = MPA-hMPA$ (e.g., MathWorks in \cite{MathW}). These measures retain the original error scale, enabling the direction and magnitude of change to be evaluated. A positive $Eff$ indicates an advantage for hMPA. At $dim = 100$ and $F9$ under $C(000)$, the mean difference is 0.452, with a $95\%$ Confidence Interval (CI) of [0.309, 0.594], while the median difference is 0.404 [0.269, 0.629]. Both measures also indicate an advantage for hMPA at $dim = 100$ on $F16$ under $C(001)$, but an advantage for MPA at $dim = 100$ on $F23$ under $C(001)$. These results demonstrate that adding $h$ can enhance accuracy under certain conditions but does not consistently offer a comparable benefit across all configurations.

Convergence complements the comparison of final errors by describing how they are reduced (see Figure~1), while boxplots (see Figure~2) show how the results are distributed across runs. At $dim = 50$ and under $C(001)$ on $F16$, the later main decrease achieved by hMPA leads to a lower final error than that achieved by MPA. Therefore, a final advantage does not necessarily imply faster improvement during the initial stages of computation. For $F1$ and $dim\in \{30, \  50, \ 100\}$, configurations involving shift combine greater dispersion with higher errors. In contrast, clustering the results on $F9$ and $F16$ under $C(011)$ and $C(111)$ does not ensure high accuracy. These results distinguish final solution quality from consistency across runs.

The overall balance of indications depends on dimensionality (see Table \ref{tab:effect_direction_summary}). At $dim = 10$, the indications more frequently favour MPA. In contrast, at $dim = 50$ and $dim = 100$, the indications more frequently favour hMPA for both measures. At $dim = 30$, the indications are equally divided for the mean, while those based on the median more frequently favour hMPA. Therefore, the balance does not change monotonically. The counts reflect the frequency of indications, not the magnitude of effects, and differences in error scales preclude direct aggregation.

\begin{table}[ht]
\centering
\caption{Counts of mean- and median-based $Eff$ comparisons classified by 95\% CI bounds. For each $dim$, 232 comparisons are included (29 functions $\times$ 8 configurations).}
\label{tab:effect_direction_summary}

\begin{tabular}{c ccc @{\qquad} ccc}
\midrule
 & \multicolumn{3}{c}{\textbf{Mean difference}}
 & \multicolumn{3}{c}{\textbf{Median difference}} \\
\cmidrule(lr){2-4} \cmidrule(lr){5-7}
$dim$ & $L>0$ & $L\leq 0\leq U$ & $U<0$
      & $L>0$ & $L\leq 0\leq U$ & $U<0$ \\
\midrule
10  & 21 & 185 & 26 & 16 & 173 & 43 \\
30  & 18 & 196 & 18 & 14 & 212 &  6 \\
50  & 27 & 198 &  7 & 27 & 202 &  3 \\
100 & 47 & 179 &  6 & 34 & 195 &  3 \\
\midrule
\end{tabular}
\end{table}
$L$ and $U$ denote the lower and upper bounds of the $95\%$ CI, respectively.
The presence of $0$ (zero) within an interval does not establish equivalence.

Most confidence intervals include $0$ for every dimension. At $dim = 100$, this applies to $77.2\%$ of mean comparisons and $84.1\%$ of median comparisons (favourable indications extend beyond functions $F1$, $F9$, $F16$ and $F23$, which are shown in the plots: e.g., Fig. 1 and Fig. 2). However, this result does not establish equivalence.

The intervals are individual, and the adjustment applied to DSC and eDSC comparisons does not account for Eff effects. Agreement in the overall balance of both measures does not imply agreement for every setting. For example, at $dim = 50$, only 18 of the 27 positive indications are shared by both measures. No settings exhibited intervals that excluded zero on opposite sides. A median interval of $[0, 0]$ does not rule out variability in the results.

The \texttt{sign} alone does not determine practical significance either. The positive mean effect on $F1$ under $C(000)$ at $dim = 10$ is approximately: 
$2.61 \times 10^{-154}$.

Function $F23$ reveals the limitations of an assessment based solely on typical performance. At $dim = 100$, configurations $C(000)$ and $C(100)$ have lower medians for hMPA than $C(010)$ and $C(110)$, but higher means. In a direct comparison of the algorithms, there are $0$ median differences for the first pair, whereas the mean differences are: $-1026.39$ and $-1188.10$, with confidence intervals entirely below $0$. Therefore, narrow boxes do not rule out significant errors in certain cases. $F23$ accounts for four of the six negative mean intervals at $dim = 100$. Therefore, the more favourable overall balance for hMPA does not preclude local deterioration relative to MPA.

The DSC and eDSC (see e.g., Section \ref{sec7}) comparison reveals different configuration preferences for hMPA on $F23$ at $dim = 100$. A more favourable spatial assessment of $C(000)$ and $C(100)$ does not necessarily result in a lower mean error than $C(010)$ and $C(110)$. Therefore, solution compactness does not guarantee a more favourable mean error. With 30 runs, the covariance rank does not exceed 29. The applied regularisation stabilises the hypervolume, but does not provide additional information about variability. As the budget increases proportionally with dimensionality ($FES_{max} = 10000\cdot dim$), the effect of dimensionality cannot be separated from the increase in budget, and asymptotic scaling cannot be inferred.

The construction of the $h$-operator (see Section \ref{Hybr}) provides a basis for interpretation: calibration uses function values and separate coordinate models, so preserving the ordering of individuals does not guarantee invariance to bias or rotation. While the observed differences are consistent with such sensitivity, they do not allow the contributions of individual mechanisms to be disentangled. Interpreting the effects of individual configuration components depends on the operations performed by the implementation. $F6$ is independent of $C3$ when the remaining conditions are held fixed. Disabling $C2$ for $F7$ and in the \texttt{Lunacek component} of $F13$ also removes conditional sign reversals and therefore does not isolate translation.

\vspace{-0.3cm}
\section{Conclusions}
\label{9}

This study examined how hMPA (Oszust et al. in \cite{MO2021}) and MPA (Faramarzi et al. in \cite{AF2020}) responded to controlled activation of bias, shift and rotation in a parameterised CEC~2017 benchmark. The 
8-configurations did not form a universal difficulty ordering. Their effects depended on the transformation configuration, test function, and problem dimension. Configuration sensitivity was observed in both MPA and hMPA. In contrast, the benefits of hybridisation were problem-dependent.
DSC and eDSC (e.g., Eftimov and Koro{\v{s}}ec in \cite{TE2019}) provided complementary evidence. DSC characterized distributions of final objective values, whereas eDSC characterized final-solution distributions in the decision space. The findings also demonstrate that convergence dynamics and final accuracy are distinct performance aspects. Run-to-run variability and solution-space structure offer further insights.

The methodological contribution extends beyond hMPA and MPA. The proposed parameterization independently controls bias, shift, and rotation while preserving the CEC~2017 (Awad et al. in \cite{NHA2016}) benchmark structure.

This test set has remained in use in recent CEC competitions. For instance, for \texttt{bound-constrained} single-objective problems in the CEC~2024 (Qiao et al. in \cite{KQ2023}) competition, 29 problems from the CEC~2017 competition
were used with $dim = 30$. A fixed budget of  $10000\cdot dim$,
and the same accuracy of the optimization and the speed were evaluated. In the CEC~2025 competition, Ban et al. in \cite{XB2024}, the same set of problems, dimensionality, evaluation budget, and configurations
regarding speed and accuracy were retained. Analogical transformation configurations and statistical process flows can therefore
be applied to other continuous optimization algorithms. Only the
algorithm-specific settings and implementation details need to be
adjusted.

The broader relevance of this study concerns the transferability of the diagnostic framework rather than the direct generalisation of numerical results.

The evidence is limited to the tested MPA and hMPA implementations and the adopted parameter settings, as well as 30 independent runs. It also covers the analysed CEC~2017 functions and the dimension-dependent evaluation budget. The study does not isolate the contribution of the predicted-candidate operator (see Section \ref{Hybr}). Nor does it formally estimate the interactions between: bias, shift and rotation.

Future research should strengthen statistical inference by using multiplicity-adjusted simultaneous confidence intervals. Controlled experiments should isolate the operator's contribution. Further studies should examine the sensitivity of eDSC regularisation and separate dimensionality from the evaluation budget. They should also validate refined hMPA variants on independent benchmarks.


\vspace{0.3cm}
\textbf{Data availability:} The results supporting the findings of this study are reported in the article. The underlying numerical datasets are available from the corresponding author upon reasonable request.

\textbf{Declaration of Competing Interest:} The authors declare that they have no known competing financial interests or personal relationships that could have appeared to influence the work reported in this paper.

\vspace{-1cm}
\renewcommand{\refname}{

\end{document}